\documentclass{article}

\usepackage[final]{nips_2018}

\usepackage[utf8]{inputenc} 
\usepackage[T1]{fontenc}    
\usepackage{hyperref}       
\usepackage{url}            
\usepackage{booktabs}       
\usepackage{amsfonts}       
\usepackage{nicefrac}       
\usepackage{microtype}      
\usepackage{wrapfig}
\usepackage{xr}
\usepackage{stfloats}

\usepackage{amssymb}
\usepackage{amsmath}
\usepackage{algorithm}
\usepackage{algorithmic}
\usepackage{multirow}
\usepackage{graphicx}
\usepackage{multicol}
\usepackage{array}
\usepackage[font=small]{caption}
\usepackage{color}

\hypersetup{
    colorlinks=true,
    citecolor=black,
    linkcolor=black,
    urlcolor=black
}

\title{Reversible Recurrent Neural Networks}

\author{
  Matthew MacKay, Paul Vicol, Jimmy Ba, Roger Grosse \\
  University of Toronto \\
  Vector Institute \\
  \texttt{\{mmackay, pvicol, jba, rgrosse\}@cs.toronto.edu}
}

\begin{document}

\maketitle

\begin{abstract}
Recurrent neural networks (RNNs) provide state-of-the-art performance in processing sequential data but are memory intensive to train, limiting the flexibility of RNN models which can be trained. Reversible RNNs---RNNs for which the hidden-to-hidden transition can be reversed---offer a path to reduce the memory requirements of training, as hidden states need not be stored and instead can be recomputed during backpropagation. We first show that perfectly reversible RNNs, which require no storage of the hidden activations, are fundamentally limited because they cannot forget information from their hidden state. We then provide a scheme for storing a small number of bits in order to allow perfect reversal with forgetting. Our method achieves comparable performance to traditional models while reducing the activation memory cost by a factor of 10--15. We extend our technique to attention-based sequence-to-sequence models, where it maintains performance while reducing activation memory cost by a factor of 5--10 in the encoder, and a factor of 10--15 in the decoder.
\end{abstract}

\section{Introduction}
\label{sec:introduction}

Recurrent neural networks (RNNs) have attained state-of-the-art performance on a variety of tasks, including speech recognition \cite{graves2013}, language modeling \cite{melis2017,merity2017}, and machine translation \cite{bahdanau2014, wu2016}. However, RNNs are memory intensive to train. The standard training algorithm is truncated backpropagation through time (TBPTT) \cite{werbos1990,rumelhart1986}. In this algorithm, the input sequence is divided into subsequences of smaller length, say $T$. Each of these subsequences is processed and the gradient is backpropagated. If $H$ is the size of our model's hidden state, the memory required for TBPTT is $O(TH)$.  

Decreasing the memory requirements of the TBPTT algorithm would allow us to increase the length $T$ of our truncated sequences, capturing dependencies over longer time scales. Alternatively, we could increase the size $H$ of our hidden state or use deeper input-to-hidden, hidden-to-hidden, or hidden-to-output transitions, granting our model greater expressivity. Increasing the depth of these transitions has been shown to increase performance in polyphonic music prediction, language modeling, and neural machine translation (NMT) \cite{pascanu2013,barone2017,zilly2016}.

Reversible recurrent network architectures present an enticing way to reduce the memory requirements of TBPTT. Reversible architectures enable the reconstruction of the hidden state at the current timestep given the next hidden state and the current input, which would enable us to perform TBPTT without storing the hidden states at each timestep. In exchange, we pay an increased computational cost to reconstruct the hidden states during backpropagation.

We first present reversible analogues of the widely used Gated Recurrent Unit (GRU) \cite{cho2014} and Long Short-Term Memory (LSTM) \cite{hochreiter1997} architectures. We then show that any perfectly reversible RNN requiring no storage of hidden activations will fail on a simple one-step prediction task. This task is trivial to solve even for vanilla RNNs, but perfectly reversible models fail since they need to memorize the input sequence in order to solve the task.
In light of this finding, we extend the memory-efficient reversal method of \citet{maclaurin2015}, storing a handful of bits per unit in order to allow perfect reversal for architectures which forget information.

We evaluate the performance of these models on language modeling and neural machine translation benchmarks. Depending on the task, dataset, and chosen architecture, reversible models (without attention) achieve 10--15-fold memory savings over traditional models.
Reversible models achieve approximately equivalent performance to traditional LSTM and GRU models on word-level language modeling on the Penn TreeBank dataset \cite{marcus1993} and lag 2--5 perplexity points behind traditional models on the WikiText-2 dataset \cite{merity2016}.

Achieving comparable memory savings with attention-based recurrent sequence-to-sequence models is difficult, since the encoder hidden states must be kept simultaneously in memory in order to perform attention. We address this challenge by performing attention over a small subset of the hidden state, concatenated with the word embedding. With this technique, our reversible models succeed on neural machine translation tasks, outperforming baseline GRU and LSTM models on the Multi30K dataset~\cite{elliott2016multi30k} and achieving competitive performance on the IWSLT 2016~\cite{iwslt2016} benchmark. Applying our technique reduces memory cost by a factor of 10--15 in the decoder, and a factor of 5--10 in the encoder.\footnote{Code will be made available at \url{https://github.com/matthewjmackay/reversible-rnn}}

\vspace{-2mm}
\section{Background}
\vspace{-2mm}
We begin by describing techniques to construct reversible neural network architectures, which we then adapt to RNNs. Reversible networks were first motivated by the need for flexible probability distributions with tractable likelihoods \cite{papamakarios2017,dinh2016,kingma2016}. Each of these architectures defines a mapping between probability distributions, one of which has a simple, known density. Because this mapping is reversible with an easily computable Jacobian determinant, maximum likelihood training is efficient.

A recent paper, closely related to our work, showed that reversible network architectures can be adapted to image classification tasks \cite{gomez2017}. Their architecture, called the Reversible Residual Network or \textit{RevNet}, is composed of a series of reversible blocks. Each block takes an input $x$ and produces an output $y$ of the same dimensionality. The input $x$ is separated into two groups: $x = [x_1; x_2]$, and outputs are produced according to the following coupling rule: 
\begin{equation}
y_1 = x_1 + F(x_2) \qquad y_2 = x_2 + G(y_1)
\end{equation}%
where $F$ and $G$ are residual functions analogous to those in standard residual networks \cite{he2016}. The output $y$ is formed by concatenating $y_1$ and $y_2$, $y = [y_1 ; y_2]$. Each layer's activations can be reconstructed from the next layer's activations as follows:
\begin{equation}
x_2 = y_2 - G(y_1) \qquad x_1 = y_1 - F(x_2)
\end{equation}%
Because of this property, activations from the forward pass need not be stored for use in the backwards pass. Instead, starting from the last layer, activations of previous layers are reconstructed during backpropagation\footnote{The activations prior to a pooling step must still be saved, since this involves projection to a lower dimensional space, and hence loss of information.}. Because reversible backprop requires an additional computation of the residual functions to reconstruct activations, it requires $33\%$ more arithmetic operations than ordinary backprop and is about $50\%$ more expensive in practice. Full details of how to efficiently combine reversibility with backpropagation may be found in \citet{gomez2017}. 

\vspace{-2mm}
\section{Reversible Recurrent Architectures}
\vspace{-2mm}

The techniques used to construct RevNets can be combined with traditional RNN models to produce reversible RNNs. In this section, we propose reversible analogues of the GRU and the LSTM.
\vspace{-2mm}
\subsection{Reversible GRU}
\vspace{-2mm}

\label{subsec:revgru}

 We start by recalling the GRU equations used to compute the next hidden state $h^{(t+1)}$ given the current hidden state $h^{(t)}$ and the current input $x^{(t)}$ (omitting biases): 
\begin{equation}
\begin{aligned}
{}[z^{(t)}; r^{(t)}] = \sigma(W [x^{(t)}; h^{(t-1)}]) &\qquad
g^{(t)} = \tanh(U [x^{(t)}; r^{(t)} \odot h^{(t-1)}]) \\
h^{(t)} = z^{(t)} \odot &h^{(t-1)} + (1 - z^{(t)}) \odot g^{(t)}
\end{aligned}
\end{equation}
Here, $\odot$ denotes elementwise multiplication.
To make this update reversible, we separate the hidden state $h$ into two groups, $h = [h_1; h_2]$. These groups are updated using the following rules:
\begin{multicols}{2}
\noindent
\begin{equation}
\begin{aligned}
\relax[z_1^{(t)}&; r_1^{(t)}] = \sigma(W_1 [x^{(t)}; h_2^{(t-1)}]) \label{eqn:revgru-update1}\\
g_1^{(t)} &= \tanh(U_1 [x^{(t)}; r_1^{(t)} \odot h_2^{(t-1)}]) \\
h_1^{(t)} &= z_1^{(t)} \odot h_1^{(t-1)} + (1 - z_1^{(t)}) \odot g_1^{(t)} 
\end{aligned}
\end{equation}
\noindent
\begin{equation}
\begin{aligned}
\relax[z_2^{(t)}&; r_2^{(t)}] = \sigma(W_2 [x^{(t)}; h_1^{(t)}]) \label{eqn:revgru-update2} \\
g_2^{(t)} &= \tanh(U_2 [x^{(t)}; r_2^{(t)} \odot h_1^{(t)}]) \\
h_2^{(t)} &= z_2^{(t)} \odot h_2^{(t-1)} + (1 - z_2^{(t)}) \odot g_2^{(t)} 
\end{aligned}
\end{equation}
\end{multicols}
\vspace{-2mm}


Note that $h_1^{(t)}$ and not $h_1^{(t-1)}$ is used to compute the update for $h_2^{(t)}$. We term this model the Reversible Gated Recurrent Unit, or \textit{RevGRU}. Note that $z_i^{(t)} \neq 0$ for $i = 1, 2$ as it is the output of a sigmoid, which maps to the open interval $(0,1)$. This means the RevGRU updates are reversible in exact arithmetic: given $h^{(t)} = [h_1^{(t)}; h_2^{(t)}]$, we can use $h_1^{(t)}$ and $x^{(t)}$ to find $z_2^{(t)}$, $r_2^{(t)}$, and $g_2^{(t)}$ by redoing part of our forwards computation. Then we can find $h_2^{(t-1)}$ using:
\begin{equation}
h_2^{(t-1)} = [h_2^{(t)} - (1 - z_2^{(t)}) \odot g_2^{(t)}] \odot 1/z_2^{(t)}
\end{equation} %
$h_1^{(t-1)}$ is reconstructed similarly.
We address numerical issues which arise in practice in Section \ref{sec:fin-prec-rev}.

\vspace{-2mm}
\subsection{Reversible LSTM}
\label{subsec:revlstm}
\vspace{-2mm}

We next construct a reversible LSTM. The LSTM separates the hidden state into an output state $h$ and a cell state $c$. The update equations are:
\vspace{-0.75cm}
\begin{multicols}{2}
\noindent
\begin{equation}
[f^{(t)}, i^{(t)}, o^{(t)}] = \sigma(W [x^{(t)}, h^{(t-1)}])
\end{equation}
\noindent
\begin{equation}
g^{(t)} = \tanh(U[x^{(t)}, h^{(t-1)}])
\end{equation}
\end{multicols}
\vspace{-1.2cm}
\begin{multicols}{2}
\noindent
\begin{equation}
c^{(t)} = f^{(t)} \odot c^{(t-1)} + i^{(t)} \odot g^{(t)} \qquad
\end{equation}
\noindent
\begin{equation}
h^{(t)} = o^{(t)} \odot \tanh(c^{(t)})
\end{equation}
\end{multicols}
\vspace{-0.5cm}

We cannot straightforwardly apply our reversible techniques, as the update for $h^{(t)}$ is not a non-zero linear transformation of $h^{(t-1)}$. Despite this, reversibility can be achieved using the equations:

\vspace{-0.7cm}
\begin{multicols}{2}
\noindent
\begin{equation}
\hspace{-0.0899cm}[f_1^{(t)}, i_1^{(t)}, o_1^{(t)}, p_1^{(t)}] = \sigma(W_1 [x^{(t)}, h_2^{(t-1)}]) \label{eqn:revlstm-zfo1}
\end{equation}
\noindent
\begin{equation}
g_1^{(t)} = \tanh(U_1 [x^{(t)}, h_2^{(t-1)}])
\end{equation}
\end{multicols}
\vspace{-1.5cm}
\begin{multicols}{2}
\noindent
\begin{equation}
c_1^{(t)} = f_1^{(t)} \odot c_1^{(t-1)} + i_1^{(t)} \odot g_1^{(t)} \label{eqn:revlstm-c1}
\end{equation}
\noindent
\begin{equation}
h_1^{(t)} = p_1^{(t)} \odot h_1^{(t-1)} + o_1^{(t)} \odot \tanh(c_1^{(t)}) \label{eqn:revlstm-h1}
\end{equation}
\end{multicols}
\vspace{-0.3cm}

We calculate the updates for $c_2, h_2$ in an identical fashion to the above equations, using $c_1^{(t)}$ and $h_1^{(t)}$. We call this model the Reversible LSTM, or \textit{RevLSTM}.

\vspace{-2mm}
\subsection{Reversibility in Finite Precision Arithmetic}
\label{sec:fin-prec-rev}
\vspace{-2mm}

We have defined RNNs which are reversible in exact arithmetic. In practice, the hidden states cannot be perfectly reconstructed due to finite numerical precision. Consider the RevGRU equations \ref{eqn:revgru-update1} and \ref{eqn:revgru-update2}.
If the hidden state $h$ is stored in fixed point, multiplication of $h$ by $z$ (whose entries are less than 1) destroys information, preventing perfect reconstruction. Multiplying a hidden unit by $1/2$, for example, corresponds to discarding its least-significant bit, whose value cannot be recovered in the reverse computation.
These errors from information loss accumulate exponentially over timesteps, causing the initial hidden state obtained by reversal to be far from the true initial state.
The same issue also affects the reconstruction of the RevLSTM hidden states.
Hence, we find that forgetting is the main roadblock to constructing perfectly reversible recurrent architectures. 

There are two possible avenues to address this limitation. The first is to remove the forgetting step. For the RevGRU, this means we compute $z_i^{(t)}$, $r_i^{(t)}$, and $g_i^{(t)}$ as before, and update $h_i^{(t)}$ using:
\begin{equation}
\label{eqn:exact-revgru}
h_i^{(t)} = h_i^{(t-1)} + (1 - z_i^{(t)}) \odot g_i^{(t)}
\end{equation}
We term this model the No-Forgetting RevGRU or \textit{NF-RevGRU}. Because the updates of the NF-RevGRU do not discard information, we need only store one hidden state in memory at a given time during training. Similar steps can be taken to define a NF-RevLSTM.

The second avenue is to accept some memory usage and store the information forgotten from the hidden state in the forward pass. We can then achieve perfect reconstruction by restoring this information to our hidden state in the reverse computation. We discuss how to do so efficiently in Section \ref{sec:mem-eff-reversibility}.


\section{Impossibility of No Forgetting}
\vspace{-2.3mm}
\begin{figure*}[t]
\begin{center}
\centerline{\includegraphics[width=\textwidth]{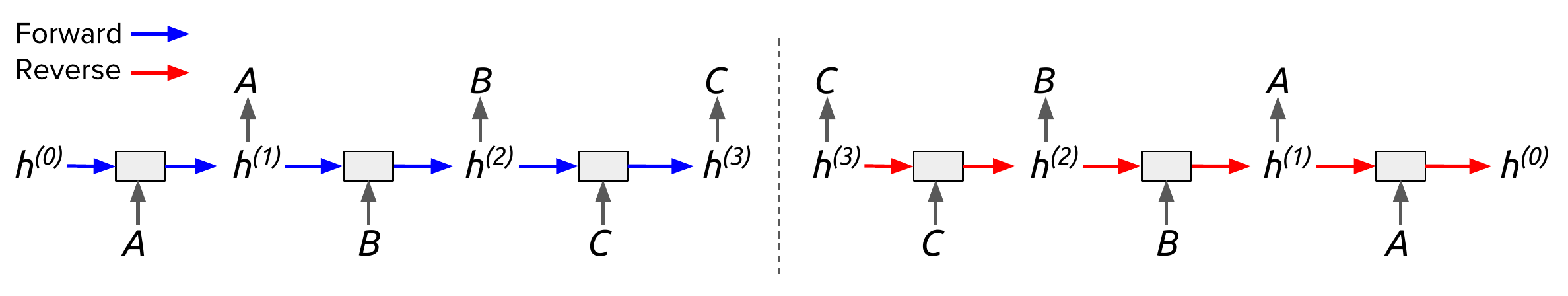}}
\caption{Unrolling the reverse computation of an exactly reversible model on the repeat task yields a sequence-to-sequence computation. {\bf Left:} The repeat task itself, where the model repeats each input token. {\bf Right:} Unrolling the reversal. The model effectively uses the final hidden state to reconstruct all input tokens, implying that the entire input sequence must be stored in the final hidden state.}
\label{fig:toy-task}
\end{center}
\vskip -0.4in
\end{figure*}

We have shown reversible RNNs in finite precision can be constructed by ensuring that no information is discarded. We were unable to find such an architecture that achieved acceptable performance on tasks such as language modeling\footnote{We discuss our failed attempts in Appendix \ref{app:no-forget-failures}.}. This is consistent with prior work which found forgetting to be crucial to LSTM performance \cite{gers1999, greff2017}. In this section, we argue that this results from a fundamental limitation of no-forgetting reversible models: 
if none of the hidden state can be forgotten, then the hidden state at any given timestep must contain enough information to reconstruct all previous hidden states. Thus, any information stored in the hidden state at one timestep must remain present at all future timesteps to ensure exact reconstruction, overwhelming the storage capacity of the model.


We make this intuition concrete by considering an elementary sequence learning task, the \textit{repeat} task. In this task, an RNN is given a sequence of discrete tokens and must simply repeat each token at the subsequent timestep. This task is trivially solvable by ordinary RNN models with only a handful of hidden units, since it doesn't require modeling long-distance dependencies. But consider how an exactly reversible model would perform the repeat task. Unrolling the reverse computation, as shown in Figure \ref{fig:toy-task}, reveals a sequence-to-sequence computation in which the encoder and decoder weights are tied. The encoder takes in the tokens and produces a final hidden state. The decoder uses this final hidden state to produce the input sequence in reverse sequential order. 


Notice the relationship to another sequence learning task, the \textit{memorization} task, used as part of a curriculum learning strategy by \citet{zaremba2014}. After an RNN observes an entire sequence of input tokens, it is required to output the input sequence in reverse order. As shown in Figure \ref{fig:toy-task}, the memorization task for an ordinary RNN reduces to the repeat task for an NF-RevRNN. Hence, if the memorization task requires a hidden representation size that grows with the sequence length, then so does the repeat task for NF-RevRNNs.

We confirmed experimentally that NF-RevGRU and NF-RevLSM networks with limited capacity were unable to solve the repeat task\footnote{We include full results and details in Appendix \ref{app:toy-task}. The argument presented applies to idealized RNNs able to implement any hidden-to-hidden transition and whose hidden units can store 32 bits each. We chose to use the LSTM and the NF-RevGRU as approximations to these idealized models since they performed best at their respective tasks.}.
Interestingly, the NF-RevGRU was able to memorize input sequences using considerably fewer hidden units than the ordinary GRU or LSTM, suggesting it may be a useful architecture for tasks requiring memorization. 
Consistent with the results on the repeat task, the NF-RevGRU and NF-RevLSTM were unable to match the performance of even vanilla RNNs on word-level language modeling on the Penn TreeBank dataset \cite{marcus1993}.

\vspace{-4mm}
\section{Reversibility with Forgetting}
\label{sec:mem-eff-reversibility}
\vspace{-2mm}

The impossibility of zero forgetting leads us to explore the second possibility to achieve reversibility: storing information lost from the hidden state during the forward computation, then restoring it in the reverse computation. Initially, we investigated \textit{discrete forgetting}, in which only an integral number of bits are allowed to be forgotten. This leads to a simple implementation: if $n$ bits are forgotten in the forwards pass, we can store these $n$ bits on a stack, to be popped off and restored to the hidden state during reconstruction. However, restricting our model to forget only an integral number of bits led to a substantial drop in performance compared to baseline models\footnote{Algorithmic details and experimental results for discrete forgetting are given in Appendix \ref{app:discrete-forgetting}.}. For the remainder of this paper, we turn to \textit{fractional forgetting}, in which a fractional number of bits are allowed to be forgotten.



To allow forgetting of a fractional number of bits, we use a technique introduced by \citet{maclaurin2015} to store lost information. To avoid cumbersome notation, we do away with super- and subscripts and consider a single hidden unit $h$ and its forget value $z$. We represent $h$ and $z$ as fixed-point numbers (integers with an implied radix point). For clarity, we write $h = 2^{-R_H}h^*$ and $z = 2^{-R_Z}z^*$. Hence, $h^*$ is the number stored on the computer and multiplication by $2^{-R_H}$ supplies the implied radix point. In general, $R_H$ and $R_Z$ are distinct. Our goal is to multiply $h$ by $z$, storing as few bits as necessary to make this operation reversible. 


The full process of reversible multiplication is shown in detail in Algorithm \ref{alg:reversible-mult}. The algorithm maintains an integer information buffer which stores $h^* \mod 2^{R_Z}$ at each timestep, so integer division of $h^*$ by $2^{R_Z}$ is reversible. However, this requires enlarging the buffer by $R_Z$ bits at each timestep. \citet{maclaurin2015} reduced this storage requirement by shifting information from the buffer back onto the hidden state. Reversibility is preserved if the shifted information is small enough so that it does not affect the reverse operation (integer division of $h^*$ by $z^*$). We include a full review of the algorithm of \citet{maclaurin2015} in Appendix \ref{app:maclaurin-review}.

However, this trick introduces a new complication not discussed by \citet{maclaurin2015}: the information shifted from the buffer could introduce significant noise into the hidden state. Shifting information requires adding a positive value less than $z^*$ to $h^*$. Because $z^* \in (0, 2^{R_Z})$ ($z$ is the output of a sigmoid function  and $z = 2^{-R_Z}z^*$), $h=2^{-R_H}h^*$ may be altered by as much $(2^{R_Z} - 1)/2^{R_H}$. If $R_Z \geq R_H$, this can alter the hidden state $h$ by $1$ or more\footnote{We illustrate this phenomenon with a concrete example in Appendix \ref{app:buffer-noise}.}. This is substantial, as in practice we observe $|h| \leq 16$. Indeed, we observed severe performance drops for $R_H$ and $R_Z$ close to equal.

The solution is to limit the amount of information moved from the buffer to the hidden state by setting $R_Z$ smaller than $R_H$. We found $R_H = 23$ and $R_Z = 10$ to work well. The  amount of noise added onto the hidden state is bounded by $2^{R_Z - R_H}$, so with these values, the hidden state is altered by at most $2^{-13}$. While the precision of our forgetting value $z$ is limited to $10$ bits, previous work has found that neural networks can be trained with precision as low as 10--15 bits and reach the same performance as high precision networks \cite{gupta2015,courbariaux2014}. We find our situation to be similar.

\begin{algorithm}[t]
  \small
  \caption{Exactly reversible multiplication (\citet{maclaurin2015})}
  \label{alg:reversible-mult}
\begin{algorithmic}[1]
  \STATE {\bfseries Input:} Buffer integer $B$, hidden state $h = 2^{-R_H} h^*$, forget value $z = 2^{-R_Z} z^*$ with $0 < z^* < 2^{R_Z}$
  \STATE $B \gets B \times 2^{R_Z}$ \COMMENT{make room for new information on buffer}                \label{step:f1}
  \STATE $B \gets B + (h^* \! \mod 2^{R_Z})$ \COMMENT{store lost information in buffer}   \label{step:f2}
  \STATE $h^* \gets h^* \div 2^{R_Z}$ \COMMENT{divide by denominator of $z$}                   \label{step:f3}
  \STATE $h^* \gets h^* \times z^*$ \COMMENT{multiply by numerator of $z$}                 \label{step:b1}
  \STATE $h^* \gets  h^* +  (B \! \mod z^*)$ \COMMENT{add information to hidden state}\label{step:b2}
  \STATE $B \gets B \div z^*$ \COMMENT{shorten information buffer}              \label{step:b3}
  \STATE \textbf{return} updated buffer $B$, updated value $h = 2^{-R_H} h^*$
\end{algorithmic}
\end{algorithm}

\vspace{-2mm}
\paragraph{Memory Savings}
To analyze the savings that are theoretically possible using the procedure above, consider an idealized memory buffer which maintains dynamically resizing storage integers $B^i_h$ for each hidden unit $h$ in groups $i = 1, 2$ of the RevGRU model. Using the above procedure, at each timestep the number of bits stored in each $B^i_h$ grows by:
\begin{equation} 
\label{eqn:cost-per-step}
R_Z - \log_2(z^*_{i,h}) = \log_2\left(2^{R_Z} / z_{i,h}^*\right) =\log_2\left(1/z_{i,h}\right)
\end{equation}
If the entries of $z_{i,h}$ are not close to zero, this compares favorably with the na{\"\i}ve cost of $32$ bits per timestep. The total storage cost of TBPTT for a RevGRU model with hidden state size $H$ on a sequence of length $T$ will be \footnote{For the RevLSTM, we would sum over $p_i^{(t)}$ and $f_i^{(t)}$ terms.}:
\begin{equation}
\label{eqn:optimal-cost}
-\left[\sum_{t = T}^T \sum_{h = 1}^H \log_2(z_{1, h}^{(t)}) + \log_2(z_{2, h}^{(t)})\right]
\end{equation}
Thus, in the idealized case, the number of bits stored equals the number of bits forgotten.

\vspace{-2mm}
\subsection{GPU Considerations}
\vspace{-2mm}
For our method to be used as part of a practical training procedure, we must run it on a parallel architecture such as a GPU. This introduces additional considerations which require modifications to Algorithm \ref{alg:reversible-mult}: (1) we implement it with ordinary finite-bit integers, hence dealing with overflow, and (2) for GPU efficiency, we ensure uniform memory access patterns across all hidden units.
\vspace{-2mm}
\paragraph{Overflow.} Consider the storage required for a single hidden unit. Algorithm \ref{alg:reversible-mult} assumes unboundedly large integers, and hence would need to be implemented using dynamically resizing integer types, as was done by \citet{maclaurin2015}. But such data structures would require non-uniform memory access patterns, limiting their efficiency on GPU architectures. Therefore, we modify the algorithm to use ordinary finite integers. In particular, instead of a single integer, the buffer is represented with a sequence of 64-bit integers $(B_0, \dots, B_D)$. Whenever the last integer in our buffer is about to overflow upon multiplication by $2^{R_Z}$, as required by step \ref{step:b1} of Algorithm \ref{alg:reversible-mult}, we append a new integer $B_{D+1}$ to the sequence. Overflow will occur if $B_D > 2^{64 - R_Z}$.

After appending a new integer $B_{D+1}$, we apply Algorithm \ref{alg:reversible-mult} unmodified, using $B_{D+1}$ in place of $B$. It is possible that up to $R_Z - 1$ bits of $B_D$ will not be used, incurring an additional penalty on storage cost. We experimented with several ways of alleviating this penalty but found that none improved significantly over the storage cost of the initial method.
\vspace{-2mm}
\paragraph{Vectorization.} Vectorization imposes an additional penalty on storage. For efficient computation, we cannot maintain different size lists as buffers for each hidden unit in a minibatch. Rather, we must store the buffer as a three-dimensional tensor, with dimensions corresponding to the minibatch size, the hidden state size, and the length of the buffer list. This means each list of integers being used as a buffer for a given hidden unit must be the same size. Whenever a buffer being used for any hidden unit in the minibatch overflows, an extra integer must be added to the buffer list for every hidden unit in the minibatch. Otherwise, the steps outlined above can still be followed. 

We give the complete, revised algorithm in Appendix \ref{app:vectorized-rev-mult}. The compromises to address overflow and vectorization entail additional overhead. We measure the size of this overhead in Section \ref{sec:experiments}.

\vspace{-2mm}
\subsection{Memory Savings with Attention}
\label{subsec:attention-mem-savings}
\vspace{-2mm}

\begin{figure*}[t]
\centering
\includegraphics[width=0.8\textwidth]{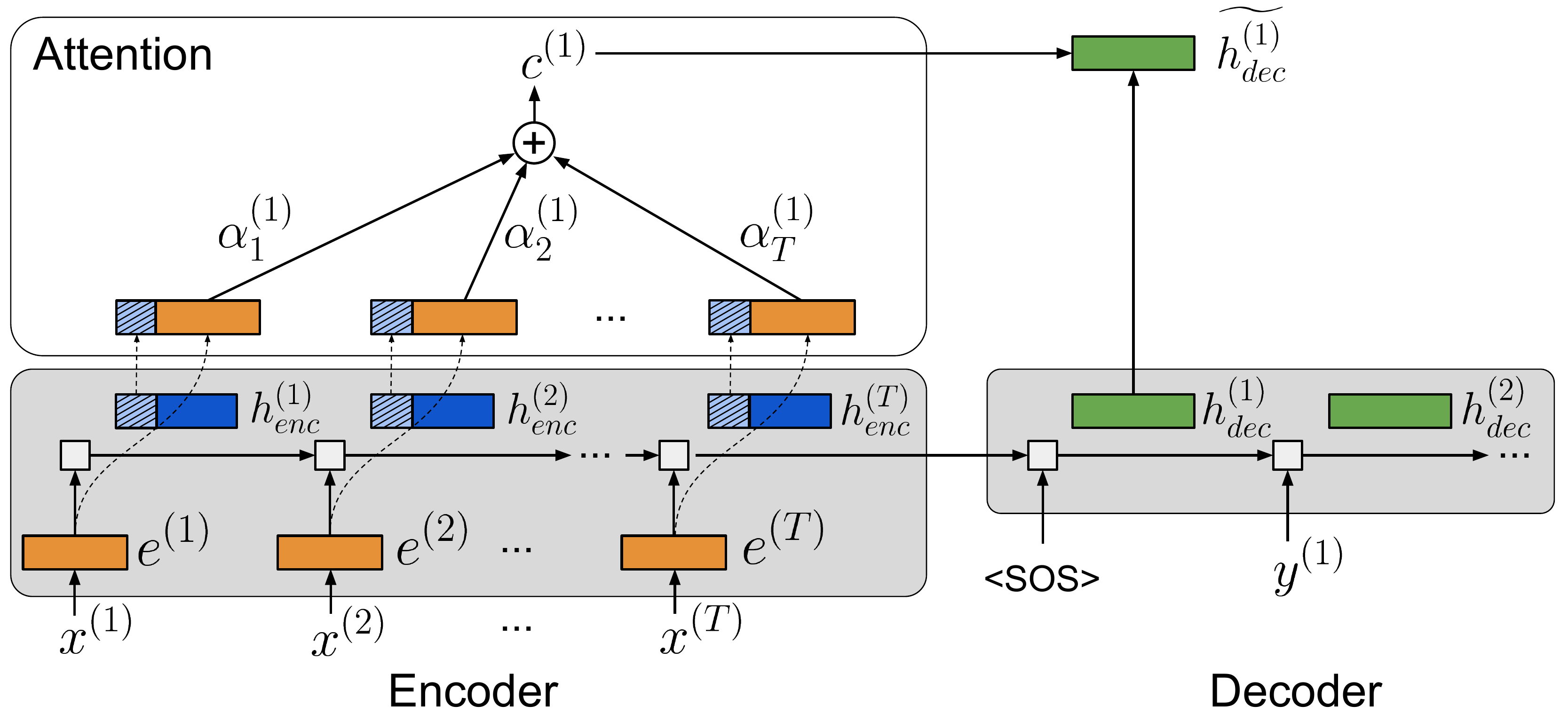}
\caption{\textbf{Attention mechanism for NMT.} The word embeddings, encoder hidden states, and decoder hidden states are color-coded orange, blue, and green, respectively; the striped regions of the encoder hidden states represent the \textit{slices} that are stored in memory for attention. The final vectors used to compute the context vector are concatenations of the word embeddings and encoder hidden state slices.
}
\label{fig:attention-diagram}
\vspace{-2mm}
\end{figure*}

Most modern architectures for neural machine translation make use of attention mechanisms~\cite{bahdanau2014,wu2016}; in this section, we describe the modifications that must be made to obtain memory savings when using attention.
We denote the source tokens by $x^{(1)}, x^{(2)}, \dots, x^{(T)}$, and the corresponding word embeddings by $e^{(1)}, e^{(2)}, \dots, e^{(T)}$.
We also use the following notation to denote \textit{vector slices}: given a vector $v \in \mathbb{R}^D$, we let $v[:k] \in \mathbb{R}^k$ denote the vector consisting of the first $k$ dimensions of $v$.
Standard attention-based models for NMT perform attention over the encoder hidden states; this is problematic from the standpoint of memory savings, because we must retain the hidden states in memory to use them when computing attention.
To remedy this, we explore several alternatives to storing the full hidden state in memory. In particular, we consider performing attention over: 1) the \textit{embeddings} $e^{(t)}$, which capture the semantics of individual words; 2) \textit{slices} of the encoder hidden states, $h^{(t)}_{enc}[:k]$ (where we consider $k = 20$ or $100$); and 3) the concatenation of embeddings and hidden state slices, $[e^{(t)} ; h^{(t)}_{enc}[:k]]$. Since the embeddings are computed directly from the input tokens, they don't need to be stored.
When we slice the hidden state, only the slices that are attended to must be stored. We apply our memory-saving buffer technique to the remaining $D-k$ dimensions.

In our NMT models, we make use of the global attention mechanism introduced by Luong et al.~\cite{luong2015effective}, where each decoder hidden state $h_{dec}^{(t)}$ is modified by incorporating context from the source annotations: a context vector $c^{(t)}$ is computed as a weighted sum of source annotations (with weights $\alpha^{(t)}_j$); $h_{dec}^{(t)}$ and $c^{(t)}$ are used to produce an attentional decoder hidden state $\widetilde{h_{dec}^{(t)}}$.
Figure~\ref{fig:attention-diagram} illustrates this attention mechanism, where attention is performed over the concatenated embeddings and hidden state slices.
Additional details on attention are provided in Appendix~\ref{app:attention}.

\vspace{-2mm}
\subsection{Additional Considerations}
\label{subsec:additional-mem-savings}
\vspace{-2mm}

\paragraph{Restricting forgetting.} In order to guarantee memory savings, we may restrict the entries of $z_i^{(t)}$ to lie in $(a, 1)$ rather than $(0, 1)$, for some $a > 0$. Setting $a = 0.5$, for example, forces our model to forget at most one bit from each hidden unit per timestep. This restriction may be accomplished by applying the linear transformation $x \mapsto (1-a)x+a$ to $z_i^{(t)}$ after its initial computation\footnote{For the RevLSTM, we would apply this transformation to $p_i^{(t)}$ and $f_i^{(t)}$.}.
\vspace{-2mm}
\paragraph{Limitations.} The main flaw of our method is the increased computational cost. We must reconstruct hidden states during the backwards pass and manipulate the buffer at each timestep. We find that each step of reversible backprop takes about 2-3 times as much computation as regular backprop. We believe this overhead could be reduced through careful engineering. We did not observe a slowdown in convergence in terms of number of iterations, so we only pay an increased per-iteration cost.

\vspace{-3mm}
\section{Experiments}
\label{sec:experiments}
\vspace{-2mm}

We evaluated the performance of reversible models on two standard RNN tasks: language modeling and machine translation.
We wished to determine how much memory we could save using the techniques we have developed, how these savings compare with those possible using an idealized buffer, and whether these memory savings come at a cost in performance.
We also evaluated our proposed attention mechanism on machine translation tasks.

\vspace{-2mm}
\subsection{Language Modeling Experiments}
\label{sec:lm-experiments}
\vspace{-3mm}

We evaluated our one- and two-layer reversible models on word-level language modeling on the Penn Treebank \cite{marcus1993} and WikiText-2 \cite{merity2016} corpora. In the interest of a fair comparison, we kept architectural and regularization hyperparameters the same between all models and datasets. We regularized the hidden-to-hidden, hidden-to-output, and input-to-hidden connections, as well as the embedding matrix, using various forms of dropout\footnote{We discuss why dropout does not require additional storage in Appendix \ref{app:dropout}.}. We used the hyperparameters from \citet{merity2017}. Details are provided in Appendix \ref{app:lm-experiments}. We include training/validation curves for all models in Appendix \ref{app:train-valid-curves}.

\vspace{-2mm}
\subsubsection{Penn TreeBank Experiments}
\label{subsubsec:ptb-experiments}
\vspace{-2mm}

\begin{table*}[t]
\caption{\small Validation perplexities (memory savings) on Penn TreeBank word-level language modeling. Results shown when forgetting is restricted to $2$, $3$, and $5$ bits per hidden unit per timestep and when there is no restriction.} 
\vspace{1mm}
    \footnotesize
    \centering
    \begin{tabular}{c|cccc||c|c}
    \toprule
    \textbf{Reversible Model} & \textbf{2 bit} & \textbf{3 bits} & \textbf{5 bits} & \textbf{No limit} & \textbf{Usual Model} & \textbf{No limit} \\ \midrule
  1 layer RevGRU  & 82.2 (13.8) & 81.1 (10.8) & 81.1 (7.4) & 81.5 (6.4) & 1 layer GRU & 82.2  \\
  2 layer RevGRU  & 83.8 (14.8) & 83.8 (12.0) & 82.2 (9.4) & 82.3 (4.9) & 2 layer GRU & 81.5 \\
  \midrule
  1 layer RevLSTM & 79.8 (13.8) & 79.4 (10.1) & 78.4 (7.4) & 78.2 (4.9) & 1 layer LSTM & 78.0 \\
  2 layer RevLSTM  & 74.7 (14.0) & 72.8 (10.0) & 72.9 (7.3) & 72.9 (4.9) & 2 layer LSTM & 73.0 \\
  \bottomrule
  \end{tabular}
\label{table:ptb-val-ppl}
\end{table*}

We conducted experiments on Penn TreeBank to understand the performance of our reversible models, how much restrictions on forgetting affect performance, and what memory savings are achievable. 

\vspace{-2mm}

\paragraph{Performance.} With no restriction on the amount forgotten, one- and two-layer RevGRU and RevLSTM models obtained roughly equivalent validation performance\footnote{Test perplexities exhibit similar patterns but are 3--5 perplexity points lower.} compared to their non-reversible counterparts, as shown in Table \ref{table:ptb-val-ppl}. To determine how little could be forgotten without affecting performance, we also experimented with restricting forgetting to at most $2$, $3$, or $5$ bits per hidden unit per timestep using the method of Section \ref{subsec:additional-mem-savings}. Restricting the amount of forgetting to $2, 3$, or $5$ bits from each hidden unit did not significantly impact performance.

Performance suffered once forgetting was restricted to at most $1$ bit. This caused a 4--5 increase in perplexity for the RevGRU. It also made the RevLSTM unstable for this task since its hidden state, unlike the RevGRU's, can grow unboundedly if not enough is forgotten. Hence, we do not include these results.

\vspace{-2mm}

\paragraph{Memory savings.} We tracked the size of the information buffer throughout training and used this to compare the memory required when using reversibility vs.~storing all activations. As shown in Appendix \ref{app:memorysavings}, the buffer size remains roughly constant throughout training. Therefore, we show the average ratio of memory requirements during training in Table \ref{table:ptb-val-ppl}. Overall, we can achieve a 10--15-fold reduction in memory when forgetting at most 2--3 bits, while maintaining comparable performance to standard models. Using Equation \ref{eqn:optimal-cost}, we also compared the actual memory savings to the idealized memory savings possible with a perfect buffer. In general, we use about twice the amount of memory as theoretically possible. Plots of memory savings for all models, both idealized and actual, are given in Appendix \ref{app:memorysavings}.



\begin{table*}[t]
    \caption{\small Validation perplexities on WikiText-2 word-level language modeling. Results shown when forgetting is restricted to $2$, $3$, and $5$ bits per hidden unit per timestep and when there is no restriction.} 
    \centering
    \small
    \begin{tabular}{c|cccc||c|c}
    \toprule
    \textbf{Reversible Model} & \textbf{2 bits} & \textbf{3 bits} & \textbf{5 bits} & \textbf{No limit} & \textbf{Usual model} & \textbf{No limit} \\ \midrule
    1 layer RevGRU & 97.7 & 97.2 & 96.3 & 97.1 & 1 layer GRU & 97.8 \\
    2 layer RevGRU & 95.2 & 94.7 & 95.3 & 95.0 & 2 layer GRU & 93.6\\ 
    \midrule
    1 layer RevLSTM     & 94.8 & 94.5 & 94.5 & 94.1 & 1 layer LSTM & 89.3 \\
    2 layer RevLSTM   & 90.7 & 87.7 & 87.0 & 86.0 & 2 layer LSTM & 82.2 \\ 
    \bottomrule
    \end{tabular}
    \vspace{-1mm}
    \label{table:wt2-val-ppl}
\end{table*}

\vspace{-2mm}
\subsubsection{WikiText-2 Experiments}
\vspace{-2mm}

We conducted experiments on the WikiText-2 dataset (WT2) to see how reversible models fare on a larger, more challenging dataset. We investigated various restrictions, as well as no restriction, on forgetting and contrasted with baseline models as shown in Table \ref{table:wt2-val-ppl}. The RevGRU model closely matched the performance of the baseline GRU model, even with forgetting restricted to $2$ bits. The RevLSTM lagged behind the baseline LSTM by about $5$ perplexity points for one- and two-layer models.

\vspace{-2mm}
\subsection{Neural Machine Translation Experiments}
\label{sec:nmt-experiments}
\vspace{-2mm}

We further evaluated our models on English-to-German neural machine translation (NMT).
We used a unidirectional encoder-decoder model and our novel attention mechanism described in Section \ref{subsec:attention-mem-savings}. We experimented on two datasets: Multi30K~\cite{elliott2016multi30k}, a  dataset of $\sim$30,000 sentence pairs derived from Flickr image captions, and IWSLT 2016~\cite{iwslt2016}, a larger dataset of $\sim$180,000 pairs.
Experimental details are provided in Appendix \ref{app:nmt-experiments}; training and validation curves are shown in Appendix~\ref{app:train-valid-curves-multi30k} (Multi30K) and \ref{app:train-valid-curves-iwslt} (IWSLT); plots of memory savings during training are shown in Appendix~\ref{app:memory-savings-nmt}.

For Multi30K, we used single-layer RNNs with 300-dimensional hidden states and 300-dimensional word embeddings for both the encoder and decoder.
Our baseline GRU and LSTM models achieved test BLEU scores of 32.60 and 37.06, respectively.
The test BLEU scores and encoder memory savings achieved by our reversible models are shown in Table~\ref{table:multi30kresults}, for several variants of attention and restrictions on forgetting.
For attention, we use Emb to denote word embeddings, $x$H for a $x$-dimensional slice of the hidden state (300H denotes the whole hidden state), and Emb+$x$H to denote the concatenation of the two.
Overall, while Emb attention achieved the best memory savings, Emb+20H achieved the best balance between performance and memory savings.
The RevGRU with Emb+20H attention and forgetting at most 2 bits achieved a test BLEU score of 34.41, outperforming the standard GRU, while reducing activation memory requirements by $7.1\times$ and $14.8\times$ in the encoder and decoder, respectively.
The RevLSTM with Emb+20H attention and forgetting at most 3 bits achieved a test BLEU score of 37.23, outperforming the standard LSTM, while reducing activation memory requirements by $8.9\times$ and $11.1\times$ in the encoder and decoder respectively.

For IWSLT 2016, we used 2-layer RNNs with 600-dimensional hidden states and 600-dimensional word embeddings for the encoder and decoder.
We evaluated reversible models in which the decoder used Emb+60H attention.
The baseline GRU and LSTM models achieved test BLEU scores of 16.07 and 22.35, respectively. The RevGRU achieved a test BLEU score of 20.70, outperforming the GRU, while saving $7.15\times$ and $12.92\times$ in the encoder and decoder, respectively. The RevLSTM achieved a score of 22.34, competitive with the LSTM, while saving $8.32\times$ and $6.57\times$ memory in the encoder and decoder, respectively. Both reversible models were restricted to forget at most $5$ bits.

\begin{table}[]
\centering
\footnotesize
\caption{Performance on the Multi30K dataset with different restrictions on forgetting. \textbf{P} denotes the test BLEU scores; \textbf{M} denotes the average memory savings of the encoder during training. 
}
\vspace{1mm}
\begin{tabular}{@{}llcccccccccc@{}}
\toprule
\multicolumn{1}{c}{\textbf{Model}}           & \multicolumn{1}{c}{\textbf{Attention}} & \multicolumn{2}{c}{\textbf{1 bit}} & \multicolumn{2}{c}{\textbf{2 bit}} & \multicolumn{2}{c}{\textbf{3 bit}} & \multicolumn{2}{c}{\textbf{5 bit}} & \multicolumn{2}{c}{\textbf{No Limit}} \\ \midrule
    &           & \textbf{P} & \textbf{M} & \textbf{P}       & \textbf{M}      & \textbf{P}      & \textbf{M}      & \textbf{P}       & \textbf{M}      & \textbf{P}        & \textbf{M}        \\ \midrule
\multicolumn{1}{c}{\multirow{5}{*}{RevLSTM}} & 20H       & 29.18  & 11.8  & 30.63  &  9.5  & 30.47  &  8.5  & 30.02  &  7.3  & 29.13  &  6.1    \\
\multicolumn{1}{c}{}                         & 100H      & 27.90  &  4.9  & 35.43  &  4.3  & 36.03  &  4.0  & 35.75  &  3.7  & 34.96  &  3.5    \\
\multicolumn{1}{c}{}                         & 300H      & 26.44  &  1.0  & 36.10  &  1.0  & 37.05  &  1.0  & 37.30  &  1.0  & 36.80  &  1.0    \\
\multicolumn{1}{c}{}                         & Emb       & 31.92  & 20.0  & 31.98  & 15.1  & 31.60  & 13.9  & 31.42  & 10.7  & 31.45  & 10.1    \\
\multicolumn{1}{c}{}                         & Emb+20H   & 36.80  & 12.1  & 36.78  &  9.9  & 37.23  &  8.9  & 36.45  &  8.1  & 37.30  &  7.4    \\ \midrule
\multicolumn{1}{c}{\multirow{5}{*}{RevGRU}}  & 20H      & 26.52  &  7.2   & 26.86  &  7.2  & 28.26  &  6.8  & 27.71  &  6.5  & 27.86  &  5.7    \\
\multicolumn{1}{c}{}                         & 100H     & 33.28  &  2.6   & 32.53  &  2.6  & 31.44  &  2.5  & 31.60  &  2.4  & 31.66  &  2.3    \\
\multicolumn{1}{c}{}                         & 300H     & 34.86  &  1.0   & 33.49  &  1.0  & 33.01  &  1.0  & 33.03  &  1.0  & 33.08  &  1.0    \\
\multicolumn{1}{c}{}                         & Emb      & 28.51  & 13.2   & 28.76  & 13.2  & 28.86  & 12.9  & 27.93  & 12.8  & 28.59  & 12.9    \\
\multicolumn{1}{c}{}                         & Emb+20H  & 34.00  &  7.2   & 34.41  &  7.1  & 34.39  &  6.4  & 34.04  &  5.9  & 34.94  &  5.7    \\
\bottomrule
\end{tabular}
\vspace{2mm}
\label{table:multi30kresults}
\vspace{-8mm}
\end{table}

\vspace{-3mm}
\section{Related Work}
\vspace{-3mm}

Several approaches have been taken to reduce the memory requirements of RNNs. Frameworks that use static computational graphs \cite{abadi2016, al2016} aim to allocate memory efficiently in the training algorithms themselves. Checkpointing \cite{martens2012training, chen2016training,gruslys2016} is a frequently used method. In this strategy, certain activations are stored as checkpoints throughout training and the remaining activations are recomputed as needed in the backwards pass. Checkpointing has previously been used to train recurrent neural networks on sequences of length $T$ by storing the activations every $\lceil \sqrt{T} \rceil$ layers \cite{martens2012training}. \citet{gruslys2016} further developed this strategy by using dynamic programming to determine which activations to store in order to minimize computation for a given storage budget.

Decoupled neural interfaces \cite{decoupled2017, czarnecki2017understanding} use auxilliary neural networks trained to produce the gradient of a layer's weight matrix given the layer's activations as input, then use these predictions to train, rather than the true gradient. This strategy depends on the quality of the gradient approximation produced by the auxilliary network. Hidden activations must still be stored as in the usual backpropagation algorithm to train the auxilliary networks, unlike our method.

Unitary recurrent neural networks \cite{arjovsky2016unitary,wisdom2016full,jing2016tunable} refine vanilla RNNs by parametrizing their transition matrix to be unitary. These networks are reversible in exact arithmetic \cite{arjovsky2016unitary}: the conjugate transpose of the transition matrix is its inverse, so the hidden-to-hidden transition is reversible. In practice, this method would run into numerical precision issues as floating point errors accumulate over timesteps. Our method, through storage of lost information, avoids these issues.

\vspace{-3mm}
\section{Conclusion}
\vspace{-3mm}
We have introduced reversible recurrent neural networks as a method to reduce the memory requirements of truncated backpropagation through time.
We demonstrated the flaws of exactly reversible RNNs, and developed methods to efficiently store information lost during the hidden-to-hidden transition, allowing us to reverse the transition during backpropagation.
Reversible models can achieve roughly equivalent performance to standard models while reducing the memory requirements by a factor of 5--15 during training. 
We believe reversible models offer a compelling path towards constructing more flexible and expressive recurrent neural networks.


\subsubsection*{Acknowledgments}
We thank Kyunghyun Cho for experimental advice and discussion. We also thank Aidan Gomez, Mengye Ren, Gennady Pekhimenko, and David Duvenaud for helpful discussion.
MM is supported by an NSERC CGS-M award, and PV is supported by an NSERC PGS-D award.

\bibliographystyle{unsrtnat}
\bibliography{bibliography}

\clearpage
\section*{Appendix}
\appendix

Here, we provide additional details about our models and results.
This appendix is structured as follows:

\begin{itemize}
    \item We discuss no-forgetting failures in Sec.~\ref{app:no-forget-failures}.
    \item We present results for our toy memorization experiment in Sec.~\ref{app:toy-task}.
    \item We provide details on reversible multiplication in Sec.~\ref{app:rev-multiplication}.
    \item We discuss discrete forgetting in Sec.~\ref{app:discrete-forgetting}.
    \item We discuss reversibility with dropout in Sec.~\ref{app:dropout}.
    \item We provide details about the attention mechanism we use in Sec.~\ref{app:attention}.
    \item We provide details on our language modeling (LM) and neural machine translation (NMT) experiments in Sec.~\ref{app:exp-details}.
    \item We plot the memory savings during training for many configurations of our RevGRU and RevLSTM models on LM and NMT in Sec.~\ref{app:memorysavings}.
    \item We provide training and validation curves for each model on the Penn TreeBank and WikiText2 language modeling task, and on the Multi30K and IWSLT-2016 NMT tasks in Sec.~\ref{app:train-valid-curves}.
\end{itemize}

\section{No-Forgetting Failures}\label{app:no-forget-failures}
We tried training NF-RevGRU models on the Penn TreeBank dataset. Without regularization, the training loss (not perplexity) of NF models blows up and remains above 100.
This is because the norm of the hidden state grows very quickly. We tried many techniques to remedy this, including: 1) penalizing the hidden state norm; 2) using different optimizers; 3) using layer normalization; and 4) using better initialization.
The best-performing model we found reached 110 train perplexity on PTB without any regularization; in contrast, even heavily regularized baseline models can reach 50 train perplexity.

\section{Toy Task Experiment}
\label{app:toy-task}
We trained an LSTM on the memorization task and an NF-RevGRU on the repeat task on sequences of length $20,35,$ and $50$.
To vary the complexity of the tasks, we experimented with hidden state sizes of $8, 16$ and $32$.
We trained on randomly generated synthetic sequences consisting of $8$ possible input tokens.
To evaluate performance, we generated an evaluation batch of $10,000$ randomly generated sequences and report the average number of tokens correctly predicted over all sequences in this batch.
To ensure exact reversibility of the NF-RevGRU, we used a fixed point representation of the hidden state, while activations were computed in floating point.

Each token was input to the model as a one-hot vector.
For the remember task, we appended another category to these one-hot vectors indicating whether the end of the input sequence has occurred.
This category was set to $0$ before the input sequence terminated and was $1$ afterwards. Models were trained by a standard cross-entropy loss objective.

We used the Adam optimizer~\cite{kingma2014} with learning rate $0.001$.
We found that a large batch size of $20,000$ was needed to achieve the best performance.
We noticed that performance continued to improve, albeit slowly, over long periods of time, so we trained our models for $1$ million batches. 
We report the maximum number of tokens predicted correctly over the course of training, as there are slight fluctuations in evaluation performance during training.

We found a surprisingly large difference in performance between the two tasks, as shown in Table~\ref{table:repeat-vs-remember}. In particular, the NF-RevGRU was able to correctly predict more tokens than expected, indicating that it was able to store a surprising amount of information in its hidden state. We suspect that the NF-RevGRU learns how to compress information more easily than an LSTM. The function NF-RevGRU must learn for the repeat task is inherently local, in contrast to the function the LSTM must learn for the remember task, which has long term dependencies.

\begin{table}[H]
\caption{Number of correct predictions made by an exactly reversible model, which cannot forget, on the repeat task and a traditional model, which can forget, on the memorization task. We expect these models to achieve equivalent performance given the same hidden state size and sequence length. With random guessing, both models would be expected to correctly predict Sequence Length/8 tokens. We also include the number of bits stored per hidden unit, after subtracting out chance accuracy.}
\centering
\begin{small}
\begin{tabular}{cccccc}
\toprule
\multirow{2}{*}{\textbf{Hidden Units}} & \multirow{2}{*}{\textbf{Sequence Length}} & \multicolumn{2}{c}{\textbf{Repeat (NF-RevGRU)}} & \multicolumn{2}{c}{\textbf{Memorization (LSTM)}} \\
& & Tokens predicted & Bits/units & Tokens predicted & Bits/unit \\
\midrule
\midrule
\multirow{3}{*}{8} & 20 & 7.9 & 2.0 & 7.4 & 1.8\\
& 35 & 13.1 & 3.3 & 9.7 & 2.0 \\
& 50 & 18.6 & 4.6 & 13.0 & 2.5\\
\midrule
\multirow{3}{*}{16} & 20 & 19.9 & 3.3 & 13.7 & 2.1 \\
& 35 & 25.4 & 3.9 & 14.3 & 1.9 \\
& 50 & 27.3 & 3.9 & 17.2 & 2.1 \\
\midrule
\multirow{3}{*}{32} & 20 & 20.0 & 2.6 & 20.0 & 2.6 \\
& 35 & 35.0 & 5.6 & 20.6 & 2.4 \\
& 50 & 47.9 & 6.2 & 21.5 & 2.3 \\
\bottomrule
\end{tabular}
\vspace{1mm}
\label{table:repeat-vs-remember}
\end{small}
\end{table}

\section{Reversible Multiplication}\label{app:rev-multiplication}

\subsection{Review of Algorithm of \citet{maclaurin2015}}\label{app:maclaurin-review}

\begin{algorithm}[t]
  \small
  \caption{Exactly reversible multiplication (\citet{maclaurin2015})}
\begin{algorithmic}[1]
  \STATE {\bfseries Input:} Buffer integer $B$, hidden state $h = 2^{-R_H} h^*$, forget value $z = 2^{-R_Z} z^*$ with $0 < z^* < 2^{R_Z}$
  \STATE $B \gets B \times 2^{R_Z}$ \COMMENT{make room for new information on buffer}                
  \STATE $B \gets B + (h^* \! \bmod 2^{R_Z})$ \COMMENT{store lost information in buffer}   
  \STATE $h^* \gets h^* \div 2^{R_Z}$ \COMMENT{divide by denominator of $z$}                   
  \STATE $h^* \gets h^* \times z^*$ \COMMENT{multiply by numerator of $z$}                 
  \STATE $h^* \gets  h^* +  (B \! \bmod z^*)$ \COMMENT{add information to hidden state}
  \STATE $B \gets B \div z^*$ \COMMENT{shorten information buffer}              
  \STATE \textbf{return} updated buffer $B$, updated value $h = 2^{-R_H} h^*$
\end{algorithmic}
\end{algorithm}

We restate the algorithm of \citet{maclaurin2015} above for convenience. Recall the goal is to multiply $h = 2^{-R_H}h^*$ by $z = 2^{-R_Z} z^*$, storing as few bits as necessary to make this operation reversible. This multiplication is accomplished by first dividing $h^*$ by $2^{R_Z}$ then multiplying by $z^*$. 

First, observe that integer division of $h^*$ by $2^{R_Z}$ can be made reversible through knowledge of $h^* \bmod 2^{R_Z}$:
\begin{equation}
    h^* = (h^* \div 2^{R_Z}) \times 2^{R_Z} + (h^* \bmod 2^{R_Z})
\end{equation}
Thus, the remainders at each timestep must be stored in order to ensure reversibility. The remainders could be stored as separate integers, but this would entail $32$ bits of storage at each timestep. Instead, the remainders are stored in a single integer information buffer $B$, which is assumed to dynamically resize upon overflow. At each timestep, the buffer's size must be enlarged by $R_Z$ bits to make room:
\begin{equation}
    B \gets B \times 2^{R_Z}
\end{equation}
Then a new remainder can be added to the buffer:
\begin{equation}
B \gets B + (h^* \! \bmod 2^{R_Z})
\end{equation}
The storage cost has been reduced from $32$ bits to $R_Z$ bits per timestep, but even further savings can be realized. Upon multiplying $h^*$ by $z^*$, there is an opportunity to add an integer $e \in \{0, 1, \dots, z^* -1 \}$ to $h^*$ without affecting the reverse process (integer division by $z^*$):
\begin{equation}
    h^* = (h^* \times z + e) \div z
\end{equation}
\citet{maclaurin2015} took advantage of this and moved information from the buffer $B$ to $h^*$ by adding $B \bmod z^*$ to $h^*$. This allows division $B$ by $z^*$ since this division can be reversed by knowledge of the modulus $B \bmod z^*$, which can be recovered from $h^*$ in the reverse process:
\begin{align}
    h^* &\gets h^* + (B \bmod z^*) \\
    B &\gets B \div z
\end{align}
We give the complete reversal algorithm as Algorithm \ref{alg:reverse-process}.

\begin{algorithm}[t]
  \small
  \caption{Reverse process of \citet{maclaurin2015}'s Algorithm}
  \label{alg:reverse-process}
\begin{algorithmic}[1]
  \STATE {\bfseries Input:} Updated buffer integer $B$, updated hidden state $h = 2^{-R_H} h^*$, forget value $z = 2^{-R_Z} z^*$ with $0 < z^* < 2^{R_Z}$
  \STATE $B \gets B \times z$                
  \STATE $B \gets B + (h^* \! \bmod z)$   
  \STATE $h^* \gets h^* \div z$                  
  \STATE $h^* \gets h^* \times 2^{R_Z}$             
  \STATE $h^* \gets  h^* +  (B \! \bmod 2^{R_Z})$ 
  \STATE $B \gets B \div 2^{R_Z}$             
  \STATE \textbf{return} Original buffer $B$, original hidden state $h = 2^{-R_H} h^*$
\end{algorithmic}
\end{algorithm}

\subsection{Noise in Buffer Computations}
\label{app:buffer-noise}

Suppose we have $R_H = R_Z = 4$, $h^* = 16$, $z^* = 17$ and $B = 1$. We hope to compute the new value for $h$ of $h = \frac{h^*}{2^{R_H}} \times \frac{z^*}{2^{R_Z}} = \frac{17}{16} = 1.0625$. Executing Algorithm \ref{alg:reversible-mult} we have:
\begin{align*}
B &\gets B \times 2^{R_Z} = 16 \\
B &\gets B + (h^* \bmod 2^{R_Z}) = 16 \\
h^* &\gets h^* \div 2^{R_Z} = 1 \\
h^* &\gets h^* \times z^* = 17 \\
h^* &\gets h^* + (B \bmod 17) = 33 \\
B &\gets B \div z^* = 0
\end{align*}
At the conclusion of the algorithm, we have that $h = \frac{h^*}{2^{R_Z}} = \frac{33}{16} = 2.0625$. The addition of information from the buffer onto the hidden state has altered it from its intended value. 

\subsection{Vectorized reversible multiplication}
\label{app:vectorized-rev-mult}

We let $N$ denote the current minibatch size. Algorithm~\ref{alg:overflow-reversible-mult} shows the vectorized reversible multiplication.

 \begin{algorithm}[H]
   \small
   \caption{Exactly reversible multiplication with overflow}
   \label{alg:overflow-reversible-mult}
\begin{algorithmic}[1]
    \STATE {\bfseries Input:} Hidden state $h = 2^{-R_H} h^*$ with dimensions $(N, H)$; forget value $z = 2^{-R_Z} z^*$ with $0< z^* < 2^{R_Z}$ and dimensions $(N, H)$; current buffer $B$, an integer tensor with dimensions $(N,H)$; past buffers $B_{past}$, an integer tensor with dimensions $(N,H,D)$
    \IF{any entry of $B$ is $\geq 2^{64 - R_Z}$}
        \STATE $B_{past} \gets [B_{past}, B]$ \COMMENT{Append $B$ to end of $B_{past}$}
        \STATE $B \gets$ tensor of zeroes with dimensions $(N, H)$ \COMMENT{Initialize new buffer}
    \ENDIF
        \STATE Execute Algorithm \ref{alg:reversible-mult} unchanged
    \STATE \textbf{return} updated buffer $B$, updated past buffers $B_{past}$, updated value $h$
\end{algorithmic}
\end{algorithm}
\vspace{-4mm}

\section{Discrete Forgetting}
\label{app:discrete-forgetting}

\subsection{Description}
Here, we consider forgetting a discrete number of bits at each timestep. This is much easier to implement than fractional forgetting, and it is interesting to explore whether fractional forgetting is necessary or if discrete forgetting will suffice.

Recall that the RevGRU updates proposed in Equations \ref{eqn:revgru-update1} and \ref{eqn:revgru-update2}. If all entries of $z_i$ are non-positive powers of $2$, then multiplication by $z_i$ corresponds exactly to a right-shift of the bits of $h_i$\footnote{When $h_i$ is negative, we must perform an additional step of appending ones to the bit representation of $h_i$ due to using two's complement representation.}. The shifted off bits can be stored in a stack, to be popped off and restored in the reverse computation. We enforce this condition by changing the equation computing $z_i$. We first choose the largest negative power of $2$ that $z_i$ could possibly represent, say $F$. $z_1^{(t)}$ is computed using\footnote{Note that the $\text{Softmax}$ is computed over rows, so the first dimension of the matrix $Q$ must be $FH$.}:

\vspace{-0.3cm}
\begin{equation}
\begin{aligned}
s_1^{(t)}[i,j] = \text{ReLU}(Q [x^{(t)}, h_2^{(t-1)}])&[Hi + j] \text{ for } 1\leq i \leq H, 1 \leq j \leq F \\
o_1^{(t)} = \text{Softmax}(\text{SampleOneHot}(s_1^{(t)})) & \qquad
z_1^{(t)} = [1, 0.5, 0.25, \dots, 2^{-F}] \cdot o_1^{(t)}
\end{aligned}
\end{equation}
The equations to calculate $z_2^{(t)}$ are analogous. We use similar equations to compute $f_i^{(t)}, p_i^{(t)}$ for the RevLSTM. To train these models, we must use techniques to estimate gradients of functions of discrete random variables. We used both the Straight-Through Categorical estimator \cite{bengio2013estimating} and the Straight-Through Gumbel-Softmax estimator \cite{jang2016,maddison2016}. In both these estimators, the forward pass is discretized but gradients during backpropagation are computed as if a continuous sample were used.


The memory savings this represents over traditional models depends on the maximum number of bits $F$ allowed to be forgotten. Instead of storing 32 bits for hidden unit per timestep, we must instead only store at most $F$ bits. We do so by using a list of integers $B = (B_1, B_2, \dots, B_D)$ as an information buffer. To store $n$ bits in $B$, we shift the bits of each $B_i$ left by $n$, then add the $n$ bits to be stored onto $B_1$. We move the bits shifted off of $B_i$ onto $B_{i+1}$ for $i \in \{1, \dots, D-1\}$. If stored bits are shifted off of $B_D$, we must append another integer to $B$. In practice, we store $F$ bits for each hidden unit regardless of its corresponding forget value. This stores some extraneous bits but is much easier to implement when vectorizing over the hidden unit dimension and the batch dimension on the GPU, as is required for computational efficiency.

\subsection{Experiments}

For discrete forgetting, we found the Straight-Through Gumbel-Softmax gradient estimator to consistently achieve results 2--3 perplexity better than the Straight-Through categorical estimator. Hence, all discrete forgetting models whose results are reported were trained using the Straight-Through Gumbel-Softmax estimator.

\begin{table*}[t]
\vspace{1mm}
    \small
    \centering
    \begin{tabular}{l|ccccc|ccccc}
    \hline
    \multirow{2}{*}{\textbf{Model}} & \multicolumn{5}{c|}{\textbf{One layer}} & \multicolumn{5}{c}{\textbf{Two layers}} \\ \cline{2-11} 
  & \textbf{1 bit} & \textbf{2 bits} & \textbf{3 bits} & \textbf{5 bits} & \textbf{No limit} & \textbf{1 bit} & \textbf{2 bits} & \textbf{3 bits} & \textbf{5 bits} & \textbf{No limit} \\ \hline
  GRU       & -    & -    & -    & -    & 82.2 & -    & -    & -    & -    & 81.5 \\
  DF-RevGRU & 93.6 & 94.1 & 93.9 & 94.7 & -    & 93.5 & 92.0 & 93.1 & 94.3 & - \\
  FF-RevGRU & 86.0 & 82.2 & 81.1 & 81.1 & 81.5 & 87.0 & 83.8 & 83.8 & 82.2 & 82.3 \\
  \hline
  LSTM       & -    & -    & -    & -    & 78.0 & -    & -    & -    & -    & 73.0  \\
  DF-RevLSTM & 85.4 & 85.1 & 86.1 & 86.8 & -    & 78.1 & 78.3 & 79.1 & 78.6 & - \\
  FF-RevLSTM & -    & 79.8 & 79.4 & 78.4 & 78.2 & -    & 74.7 & 72.8 & 72.9 & 72.9 \\
  \hline
  \end{tabular}
\caption{\small Validation perplexities on Penn TreeBank word-level language modeling. Test perplexities exhibit a similar pattern but are 3--5 perplexity points lower. DF denotes discrete forgetting and FF denotes fractional forgetting. We show perplexities when forgetting is restricted to $1$, $2$, $3$, and $5$ bits per hidden unit and when there is no limit placed on the amount forgotten.}
\label{table:df-ptb-val-ppl}
\end{table*}

\paragraph{Discrete vs. Fractional Forgetting.} We show complete results on Penn TreeBank validation perplexity in Table \ref{table:df-ptb-val-ppl}. Overall, models which use discrete forgetting performed 4-10 perplexity points worse on the validation set than their fractional forgetting counterparts. It could be the case that the stochasticity of the samples used in discrete forgetting models already imposes a regularizing effect, causing discrete models to be too heavily regularized. To check, we also ran experiments using lower dropout rates and found that discrete forgetting models still lagged behind their fractional counterparts. We conclude that information must be discarded from the hidden state in fine, not coarse, quantities.

\section{Discussion of Dropout}\label{app:dropout}

First, consider dropping out elements of the input.
If the same elements are dropped out at each step, we simply store the single mask used, then apply it to the input at each step of our forwards and reverse computation.

Applying dropout to the hidden state does not entail information loss (and hence additional storage), since we can interpret dropout as masking out elements of the input/hidden-to-hidden matrices. If the same dropout masks are used at each timestep, as is commonly done in RNNs, we store the single weight mask used, then use the dropped-out matrix in the forward and reverse passes.
If the same rows of these matrices are dropped out (as in variational dropout), we need only store a mask the same size as the hidden state.

If we wish to sample different dropout masks at each timestep, which is not commonly done in RNNs, we would either need to store the mask used at each timestep, which is memory intensive, or devise a way to recover the sampled mask in the reverse computation (e.g., using a reversible sampler, or using a deterministic function to set the random seed at each step).

\section{Attention Details}
\label{app:attention}

In our NMT experiments, we use the \textit{global attention mechanism} introduced by Luong et al.~\cite{luong2015effective}.
We consider attention performed over a set of \textit{source-side annotations} $\{ s^{(1)}, \dots, s^{(T)} \}$, which can be either: 1) the encoder hidden states, $s^{(t)} = h^{(t)}_{enc}$; 2) the source embeddings, $s^{(t)} = e^{(t)}$; or 3) a concatenation of the embeddings and $k$-dimensional slices of the hidden states, $s^{(t)} = [e^{(t)}; h^{(t)}_{enc}[:k]]$.
When using global attention, the model first computes the decoder hidden states $\{ h^{(1)}_{dec}, \dots, h^{(M)}_{dec} \}$ as in the standard encoder-decoder paradigm, and then it \textit{modifies} each $h^{(t)}_{dec}$ by incorporating context from the source annotations.
A context vector $c^{(t)}$ is computed as a weighted sum of the source annotations:
\begin{equation}
c^{(t)} = \sum_{j=1}^T \alpha^{(t)}_j s^{(j)}
\end{equation}
where the weights $\alpha^{(t)}_j$ are computed by scoring the similarity between the ``current'' decoder hidden state $h^{(t)}_{dec}$ and each of the encoder annotations:
\begin{equation}
\alpha^{(t)}_j = \frac{\exp(\text{score}(h^{(t)}_{dec}, s^{(j)}))}{\sum_{k=1}^T \exp(\text{score}(h^{(t)}_{dec}, s^{(k)}))}
\end{equation}

As the $\text{score}$ function, we use the ``general'' formulation proposed by Luong et al.:
\begin{equation}
\text{score}(h^{(t)}_{dec}, s^{(j)}) = (h^{(t)}_{dec})^\top W_a s^{(j)}
\end{equation}

Then, the original decoder hidden state $h^{(t)}_{dec}$ is modified via the context $c^{(t)}$, to produce an \textit{attentional} hidden state $\widetilde{h^{(t)}_{dec}}$:
\begin{equation}
\widetilde{h^{(t)}_{dec}} = \tanh(W_c [ c^{(t)} ; h^{(t)}_{dec} ])
\end{equation}
Finally, the attentional hidden state $\widetilde{h^{(t)}_{dec}}$ is passed into the softmax layer to produce the output distribution:
\begin{equation}
p(y^{(t)} \mid y^{(1)}, \dots, y^{(t-1)}, \mathbf{x}) = \text{softmax}\left(W_s \widetilde{h^{(t)}_{dec}}\right)
\end{equation}

\section{Experiment Details}
\label{app:exp-details}

All experiments were implemented using PyTorch \cite{paszke2017automatic}. Neural machine translation experiments were implemented using OpenNMT \cite{2017opennmt}.

\subsection{Language Modeling Experiments}
\label{app:lm-experiments}

We largely followed \citet{merity2017} in setting hyperparameters. All one-layer models used $650$ hidden units and all two-layer models used $1150$ hidden units in their first layer and $650$ in their second. We kept our embedding size constant at $650$ through all experiments. 

Notice that with a fixed hidden state size, a reversible architecture will have fewer parameters than a standard architecture. If the total number of hidden units is $H$, the number of hidden-to-hidden parameters is $2 \times (H/2)^2 = H^2/2$ in a reversible model, compared to $H^2$ for its non-reversible counterpart. For the RevLSTM, there are extra hidden-to-hidden parameters due to the $p$ gate needed for reversibility. Each model also has additional parameters associated with the input-to-hidden connections and embedding matrix.

We show the total number of parameters in each model, including embeddings, in Table \ref{table:ptb-params}.

We used DropConnect \cite{wan2013} with probability $0.5$ to regularize all hidden-to-hidden matrices. We applied variational dropout \cite{gal2016theoretically} on the inputs and outputs of the RNNs. The inputs to the first layer were dropped out with probability $0.3$. The outputs of each layer were dropped out with probability $0.4$. As in \citet{gal2016theoretically}, we used embedding dropout with probability $0.1$.  We also applied weight decay with scalar factor $1.2 \times 10^{-6}$.

We used a learning rate of $20$ for all models, clipping the norm of the gradients to be smaller than $0.1$. We decayed the learning rate by a factor of $4$ once the nonmonotonic criterion introduced by \citet{merity2017} was triggered and used the same non-monotone interval of $5$ epochs. For discrete forgetting models, we found that a learning rate decay factor of $2$ worked better. Training was stopped once the learning rate is below $10^{-2}$.

\begin{table}
\caption{Total number of parameters in each model used for LM.}
\vspace{-2mm}
\centering
\setlength\tabcolsep{4pt}
\vspace{5mm}
\centering
\small
\begin{tabular}{lc}
\toprule
\small
\textbf{Model} & \textbf{Total number of parameters} \\ \midrule
1 layer GRU            & 9.0M              \\
1 layer RevGRU         & 8.4M              \\
\hline
1 layer LSTM           & 9.9M              \\
1 layer RevLSTM        & 9.7M              \\ 
\hline
2 layer GRU           & 16.2M             \\
2 layer RevGRU           & 13.6M              \\
\hline 
2 layer LSTM           & 19.5M              \\
2 layer RevLSTM           & 18.4M              \\\bottomrule

\end{tabular}
\vspace{1mm}
\label{table:ptb-params}
\end{table}

Like \citet{merity2017}, we used variable length backpropagation sequences. The base sequence length was set to $70$ with probability $0.95$ and set to $35$ otherwise. The actual sequence length used was then computed by adding random noise from $\mathcal{N}(0, 5)$ to the base sequence length. We rescaled the learning rate linearly based on the length of the truncated sequences, so for a given minibatch of length $T$, the learning rate used was $20 \times \frac{T}{70}$.

\subsection{Neural Machine Translation Experiments}
\label{app:nmt-experiments}
\vspace{-2mm}

\paragraph{Multi30K Experiments.}
The Multi30K dataset \cite{elliott2016multi30k} contains English-German sentence pairs derived from captions of Flickr images, and consists of 29,000 training, 1,015 validation, and 1,000 test sentence pairs. The average length of the source (English) sequences is 13 tokens, and the average length of the target (German) sequences is 12.4 tokens.

We applied variational dropout with probability 0.4 to inputs and outputs.
We trained on mini-batches of size 64 using SGD.
The learning rate was initialized to 0.2 for GRU and RevGRU, 0.5 for RevLSTM, and 1 for the standard LSTM---these values were chosen to optimize the performance of each model.
The learning rate was decayed by a factor of 2 each epoch when the validation loss failed to improve from the previous epoch.
Training halted when the learning rate dropped below 0.001.
Table~\ref{table:multi30k-val-bleu} shows the validation BLEU scores of each RevGRU and RevLSTM variant.

\begin{table}[H]
\centering
\footnotesize
\caption{BLEU scores on the Multi30K validation set. For the attention type, Emb denotes word embeddings, $x$H denotes a $x$-dimensional slice of the hidden state (300H corresponds to the whole hidden state), and Emb+$x$H denotes the concatenation of the two.}
\vspace{1mm}
\label{table:multi30k-val-bleu}
\begin{tabular}{@{}clccccc@{}}
\toprule
\textbf{Model}           & \multicolumn{1}{c}{\textbf{Attention}} & \textbf{1 bit} & \textbf{2 bit} & \textbf{3 bit} & \textbf{5 bit} & \textbf{No Limit} \\ \midrule
\multirow{5}{*}{RevLSTM} & 20H     & 28.51   & 29.72   & 30.65   & 29.82   & 29.11   \\
                         & 100H    & 28.10   & 35.52   & 36.13   & 34.97   & 35.14   \\
                         & 300H    & 26.46   & 36.73   & 37.04   & 37.32   & 37.27   \\
                         & Emb     & 31.27   & 30.96   & 31.41   & 31.31   & 31.95   \\
                         & Emb+20H & 36.33   & 36.75   & 37.54   & 36.89   & 36.51   \\ \midrule
\multirow{5}{*}{RevGRU} & 20H      & 25.96   & 25.86   & 27.25   & 27.13   & 26.96   \\
                        & 100H     & 32.52   & 32.86   & 31.08   & 31.16   & 31.87   \\
                        & 300H     & 34.26   & 34.00   & 33.02   & 33.08   & 32.24   \\
                        & Emb      & 27.57   & 27.59   & 28.03   & 27.24   & 28.07   \\
                        & Emb+20H  & 33.67   & 34.94   & 34.36   & 34.87   & 35.12   \\ \bottomrule
\end{tabular}
\end{table}

\paragraph{IWSLT-2016 Experiments.}

For both the encoder and decoder we used unidirectional, two-layer RNNs with 600-dimensional hidden states and 600-dimensional word embeddings.
We applied variational dropout with probability 0.4 to the inputs and the output of each layer.
The learning rates were initialized to 0.2 for the GRU, RevGRU, and RevLSTM, and 1 for the LSTM.
We used the same learning rate decay and stopping criterion as for the Multi30K experiments.

The RevGRU with attention over the concatenation of embeddings and a 60-dimensional slice of the hidden state and 5 bit forgetting achieved a BLEU score of 23.65 on the IWSLT validation set; the RevLSTM with the same attention and forgetting configuration achieved a validation BLEU score of 26.17.
The baseline GRU achieved a validation BLEU score of 18.92, while the baseline LSTM achieved 26.31.

\section{Memory Savings}
\label{app:memorysavings}

\subsection{Language modeling}

\subsubsection*{1 layer RevGRU on Penn TreeBank}
\begin{figure}[H] 
  \caption*{Ratio of memory used by storing discarded information in a buffer and using reversibility vs. storing all activations na{\"\i}vely. \textbf{Left:} Actual savings obtained by our method.  \textbf{Right:} Idealized savings obtained by using a perfect buffer.}
  \begin{center}
    \includegraphics[width=0.5\textwidth]{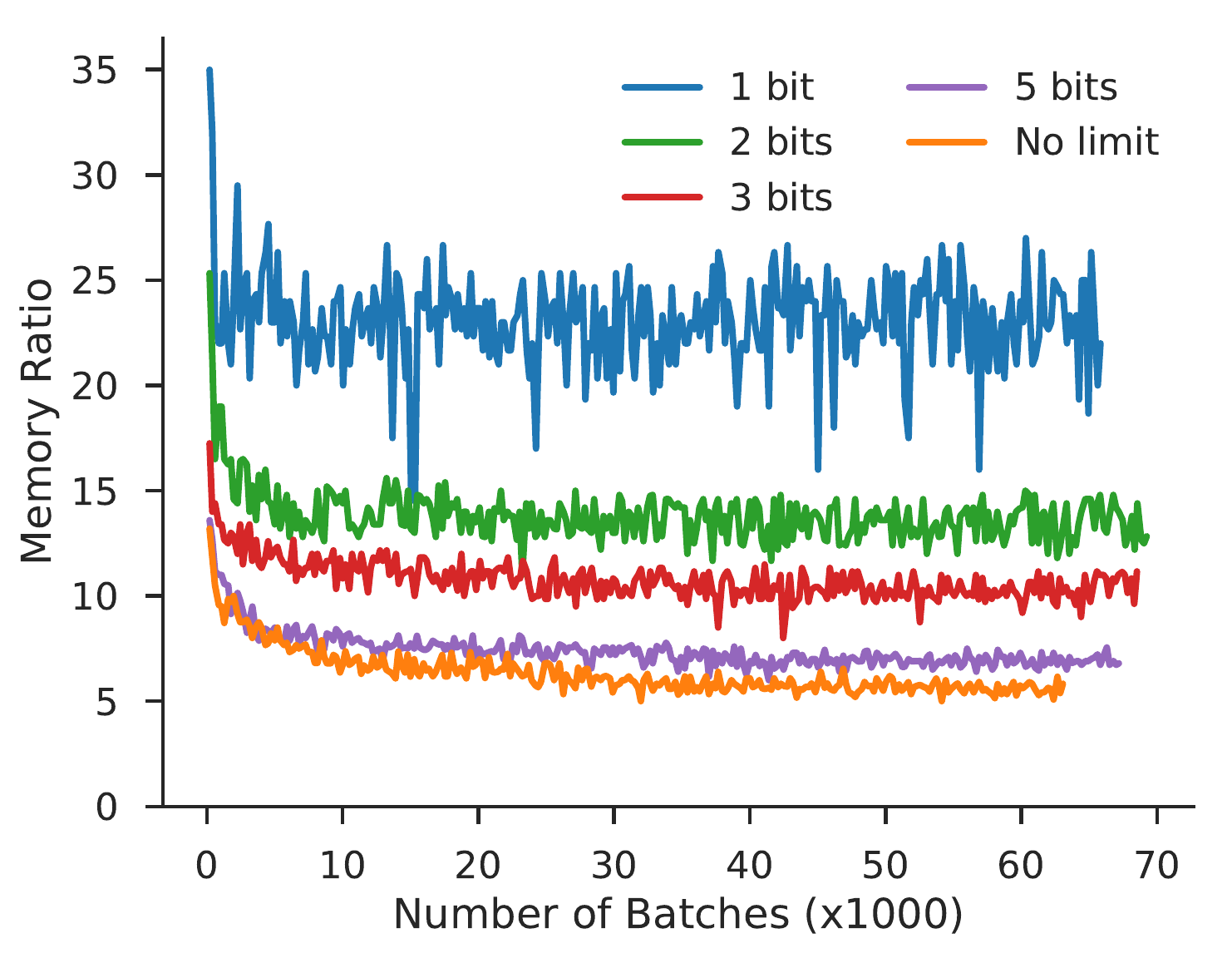}\includegraphics[width=0.5\textwidth]{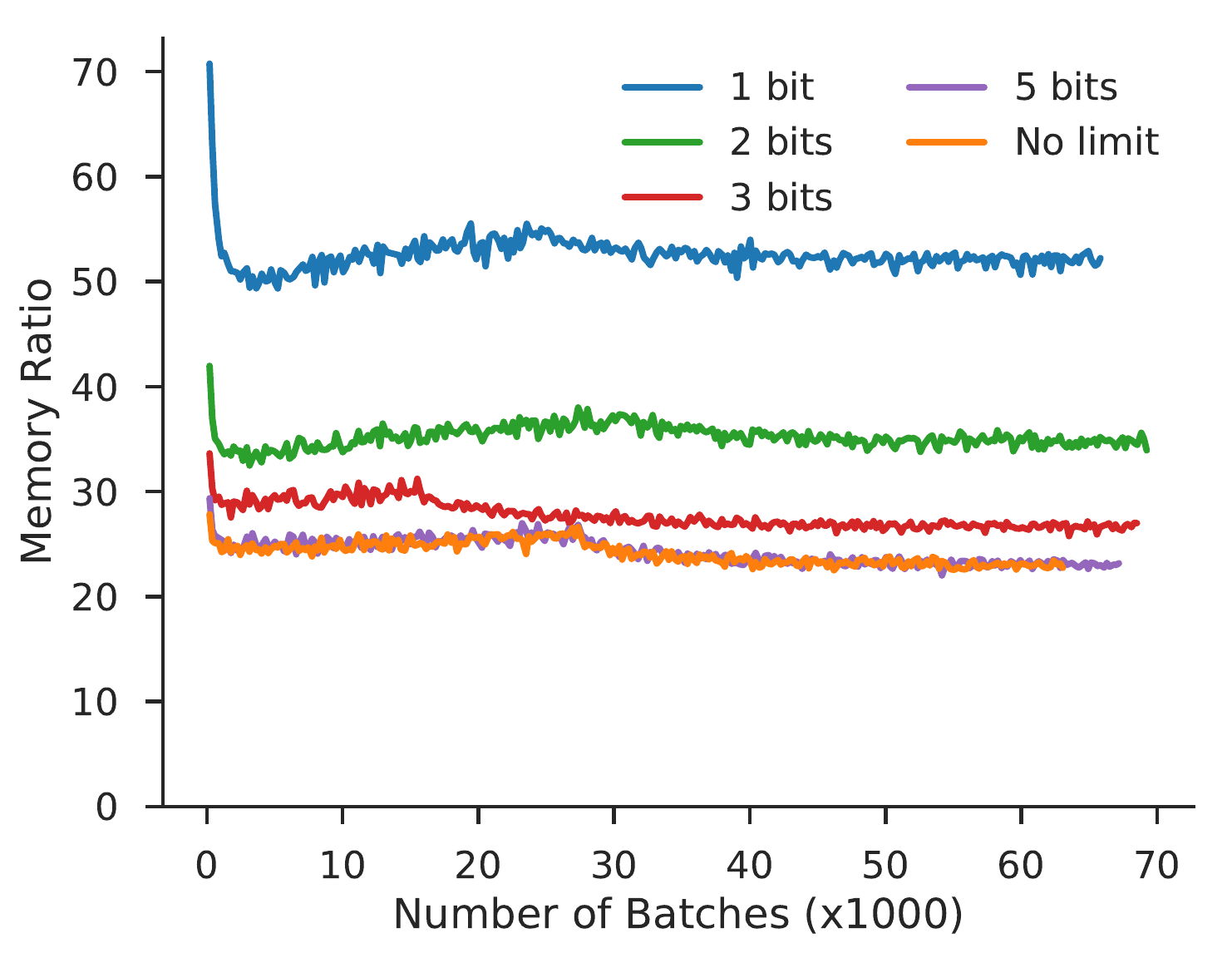}
  \end{center}
  \label{fig:single-dropout}
\end{figure}

\subsubsection*{2 layer RevGRU on Penn TreeBank}
\begin{figure}[H]
  \caption*{Ratio of memory used by storing discarded information in a buffer and using reversibility vs. storing all activations na{\"\i}vely. \textbf{Left:} Actual savings obtained by our method.  \textbf{Right:} Idealized savings obtained by using a perfect buffer.}
  \begin{center}
    \includegraphics[width=0.5\textwidth]{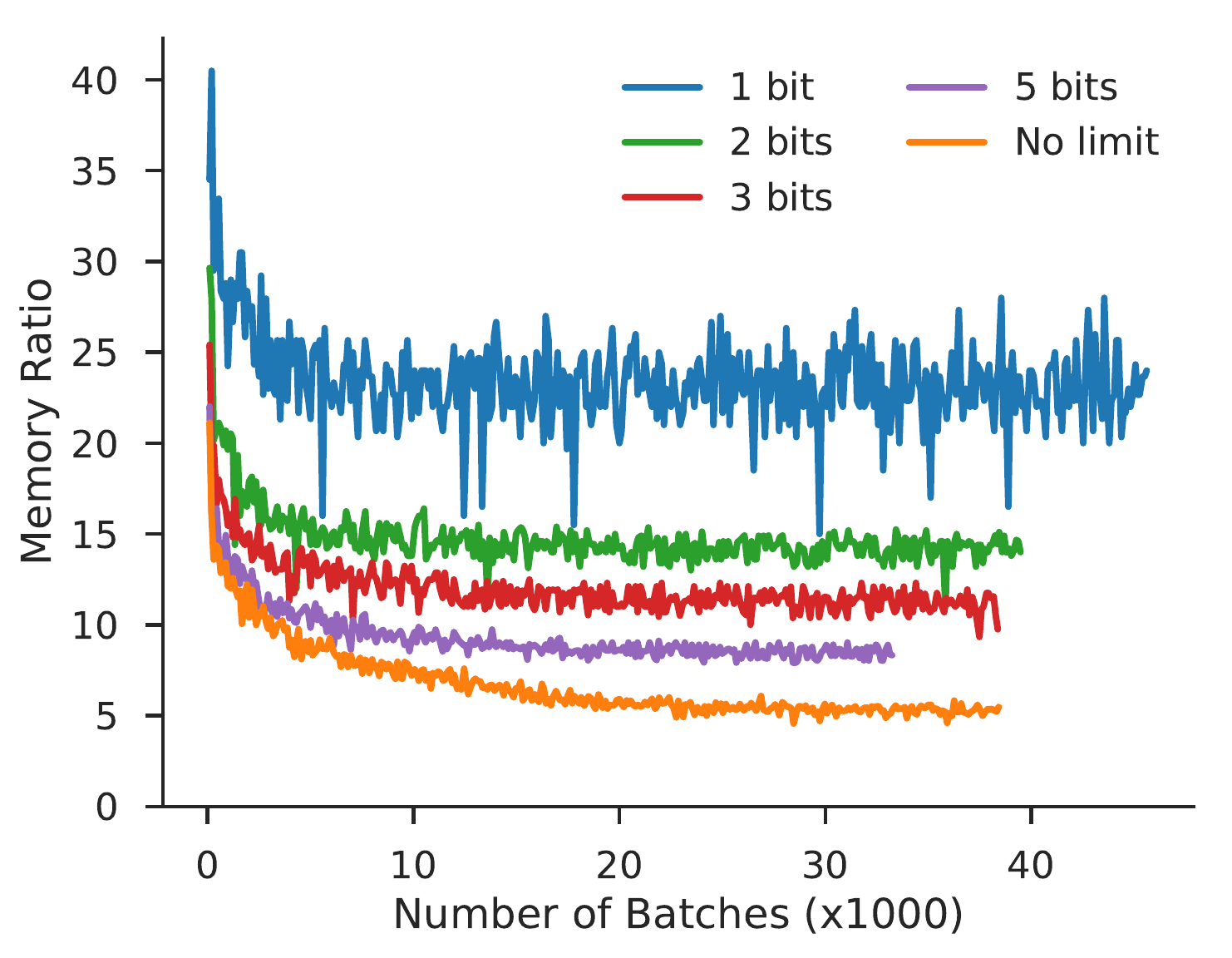}\includegraphics[width=0.5\textwidth]{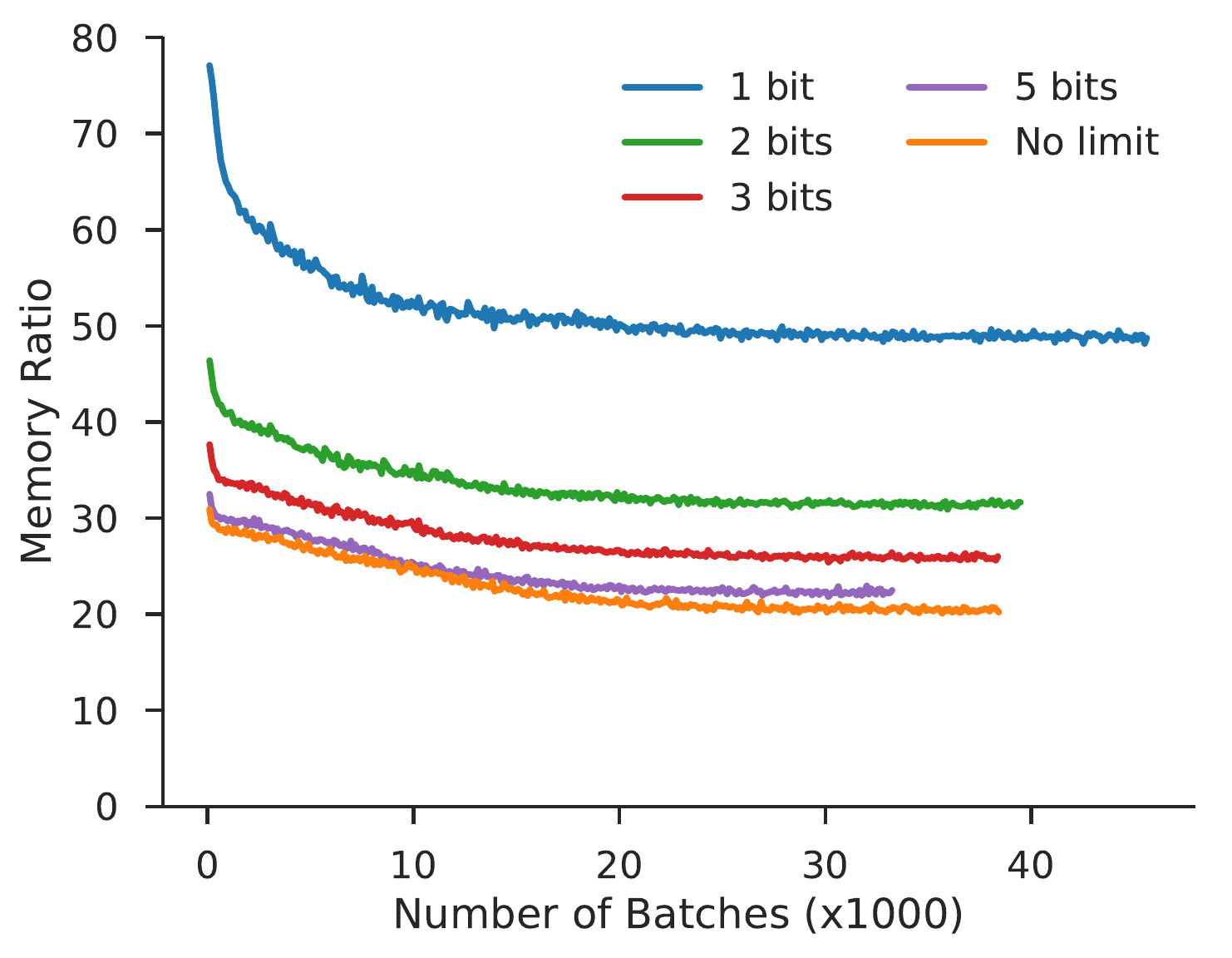}
  \end{center}
  \label{fig:single-dropout}
\end{figure}

\subsubsection*{1 layer RevLSTM on Penn TreeBank}
\begin{figure}[H]
  \caption*{Ratio of memory used by storing discarded information in a buffer and using reversibility vs. storing all activations na{\"\i}vely. \textbf{Left:} Actual savings obtained by our method.  \textbf{Right:} Idealized savings obtained by using a perfect buffer.}
  \begin{center}
    \includegraphics[width=0.5\textwidth]{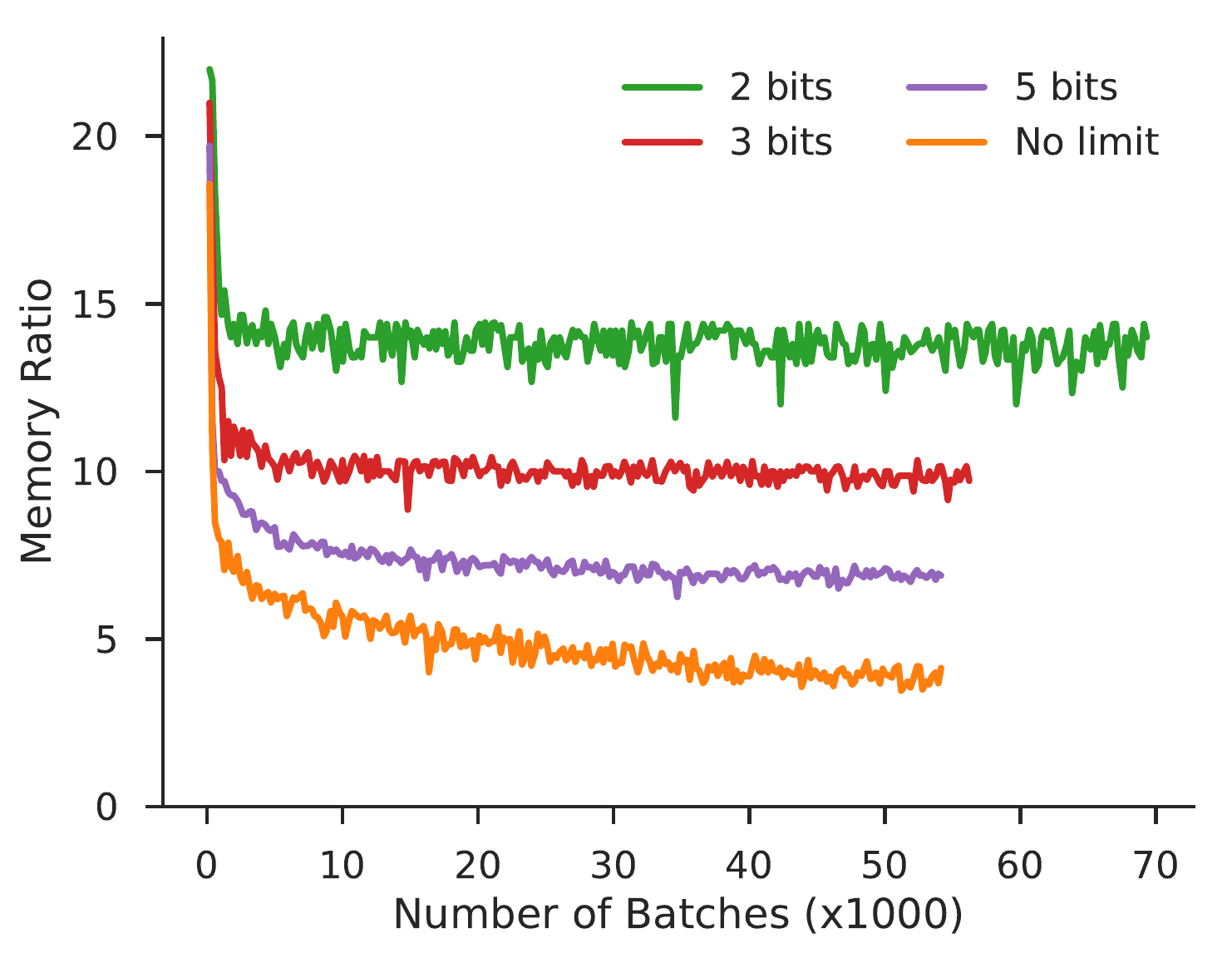}\includegraphics[width=0.5\textwidth]{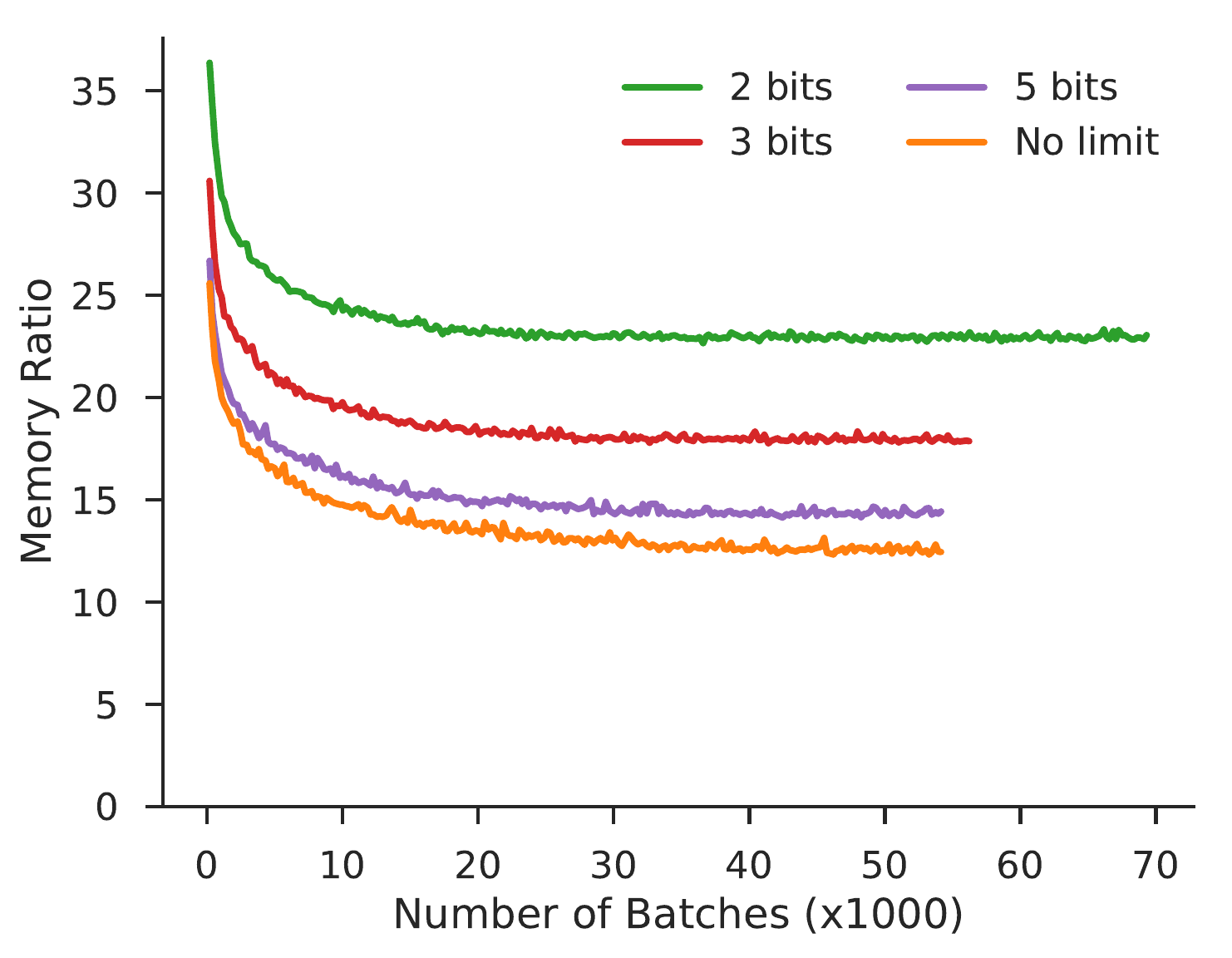}
  \end{center}
  \label{fig:single-dropout}
\end{figure}

\subsubsection*{2 layer RevLSTM on Penn TreeBank}
\begin{figure}[H]
  \caption*{Ratio of memory used by storing discarded information in a buffer and using reversibility vs. storing all activations na{\"\i}vely. \textbf{Left:} Actual savings obtained by our method.  \textbf{Right:} Idealized savings obtained by using a perfect buffer.}
  \begin{center}
    \includegraphics[width=0.5\textwidth]{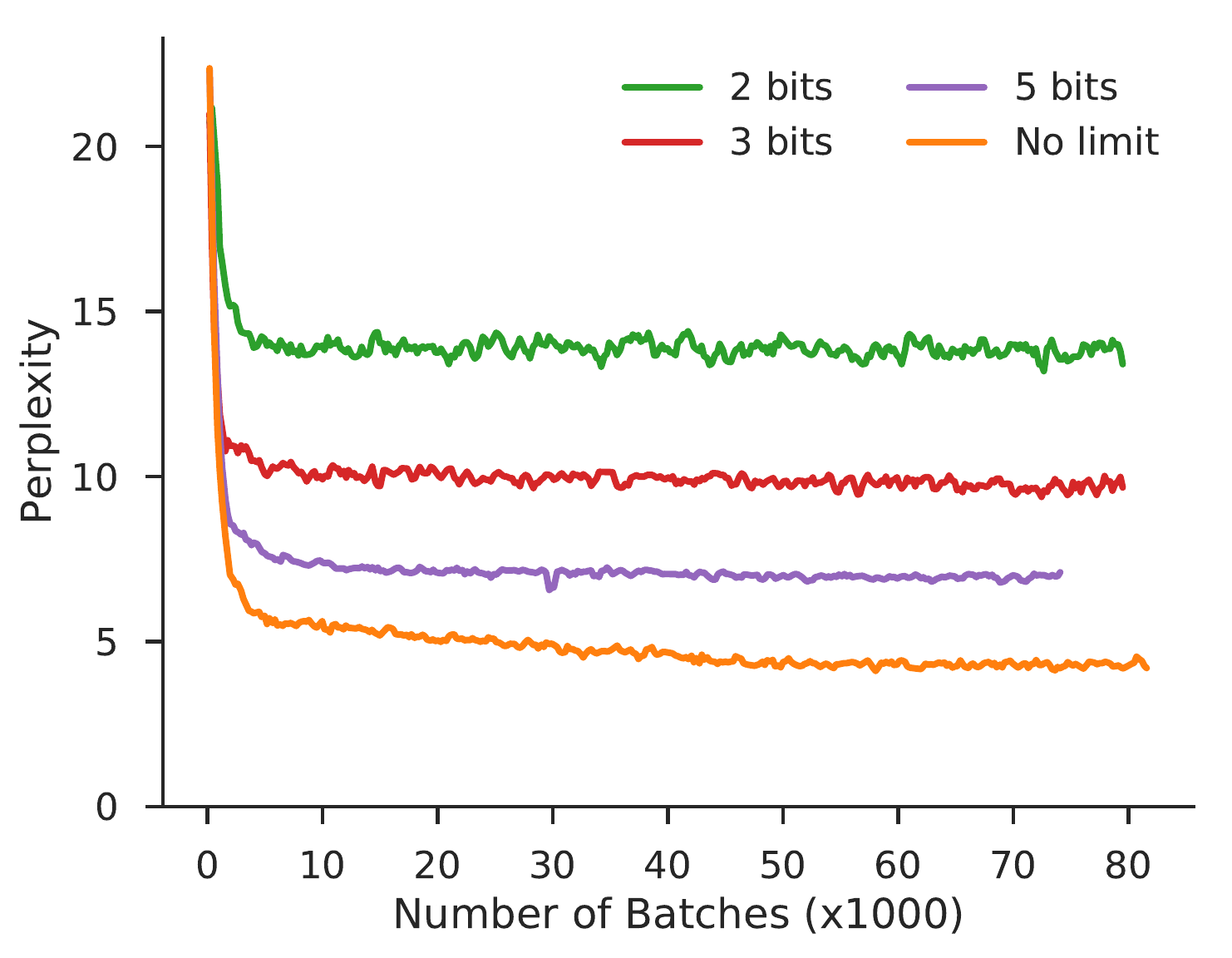}\includegraphics[width=0.5\textwidth]{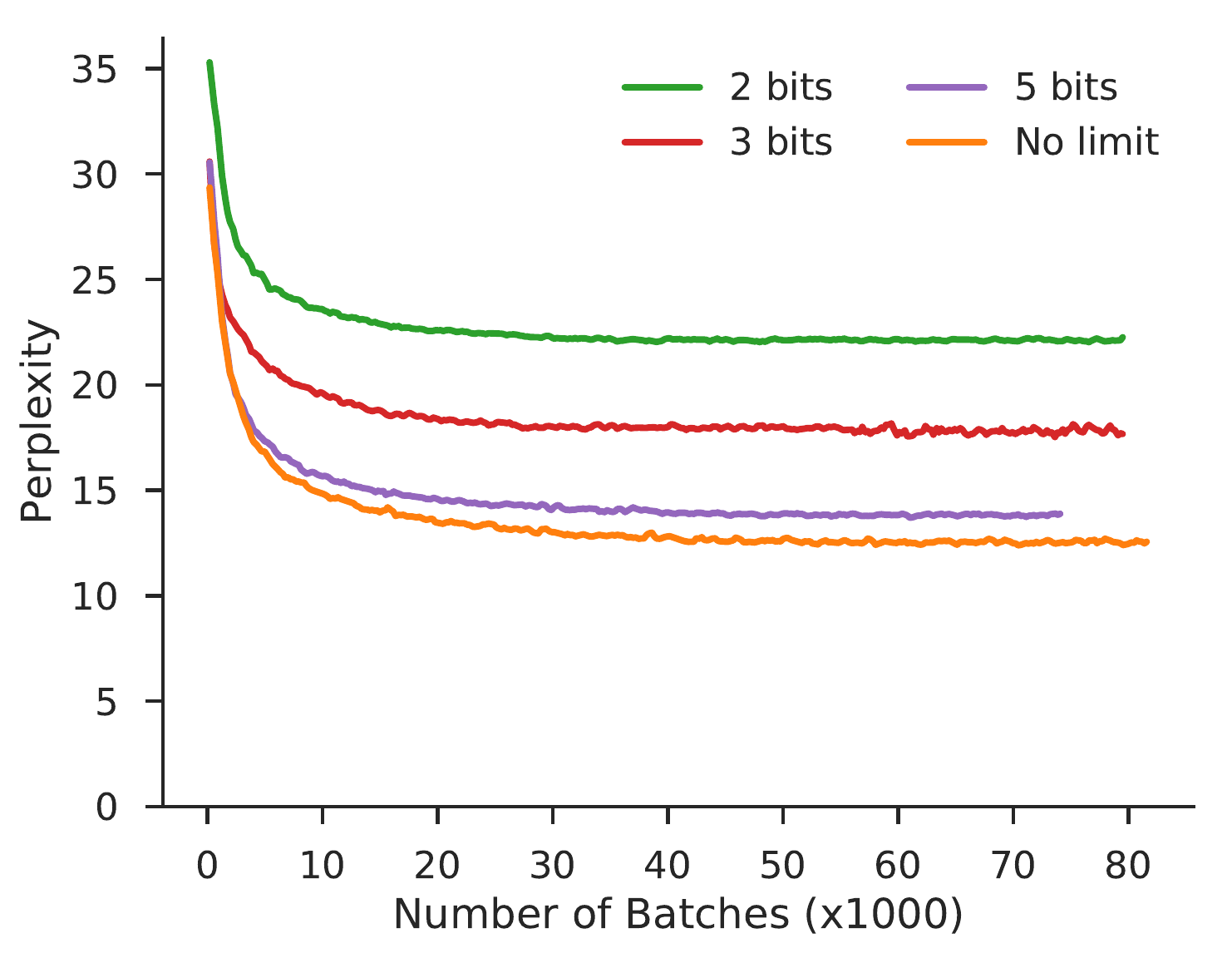}
  \end{center}
  \label{fig:single-dropout}
\end{figure}

\subsection{Neural Machine Translation}
\label{app:memory-savings-nmt}

In this section, we show the memory savings achieved by the encoder and decoder of our reversible NMT models. The memory savings refer to the ratio of the amount of memory needed to store discarded information in a buffer for reversibility, compared to storing all activations.
Table~\ref{table:decoder-memory-savings} shows the memory savings in the decoder for various RevGRU and RevLSTM models on Multi30K.

\begin{table}[H]
\centering
\footnotesize
\caption{Average memory savings in the decoder for NMT on the Multi30K dataset, during training. For the attention type, Emb denotes word embeddings, $x$H denotes a $x$-dimensional slice of the hidden state, and Emb+$x$H denotes the concatenation of the two.}
\vspace{1mm}
\label{table:decoder-memory-savings}
\begin{tabular}{@{}clccccc@{}}
\toprule
\textbf{Model}           & \multicolumn{1}{c}{\textbf{Attention}} & \textbf{1 bit} & \textbf{2 bit} & \textbf{3 bit} & \textbf{5 bit} & \textbf{No Limit} \\ \midrule
\multirow{5}{*}{RevLSTM} & 20H       & 24.0   & 13.6   & 10.7   & 7.9    & 6.6   \\
                         & 100H      & 24.1   & 13.9   & 10.1   & 8.0    & 5.5   \\
                         & 300H      & 24.7   & 13.4   & 10.7   & 8.3    & 6.5   \\
                         & Emb       & 24.1   & 13.5   & 10.5   & 8.0    & 6.7   \\
                         & Emb+20H   & 24.4   & 13.7   & 11.1   & 7.8    & 7.8   \\ \midrule
\multirow{5}{*}{RevGRU}  & 20H       & 24.1   & 13.5   & 11.1   &  8.8   &  7.9  \\
                         & 100H      & 26.0   & 14.1   & 12.2   &  9.5   &  8.2  \\
                         & 300H      & 26.1   & 14.8   & 13.0   & 10.0   &  9.8  \\
                         & Emb       & 25.9   & 14.1   & 12.5   &  9.8   &  8.3  \\
                         & Emb+20H   & 25.5   & 14.8   & 12.9   & 11.2   &  8.9  \\ \bottomrule
\end{tabular}
\end{table}

In sections~\ref{app:revgru-mem-plots}, \ref{app:revlstm-mem-plots}, and \ref{app:iwslt-mem-plots}, we plot the memory savings during training for RevGRU and RevLSTM models on Multi30K and IWSLT-2016, using various levels of forgetting. In each plot, we show the actual memory savings achieved by our method, as well as the idealized savings obtained by using a perfect buffer.

\subsubsection{RevGRU on Multi30K}
\label{app:revgru-mem-plots}

\begin{figure}[H]
    \centering
    \includegraphics[width=0.32\linewidth]{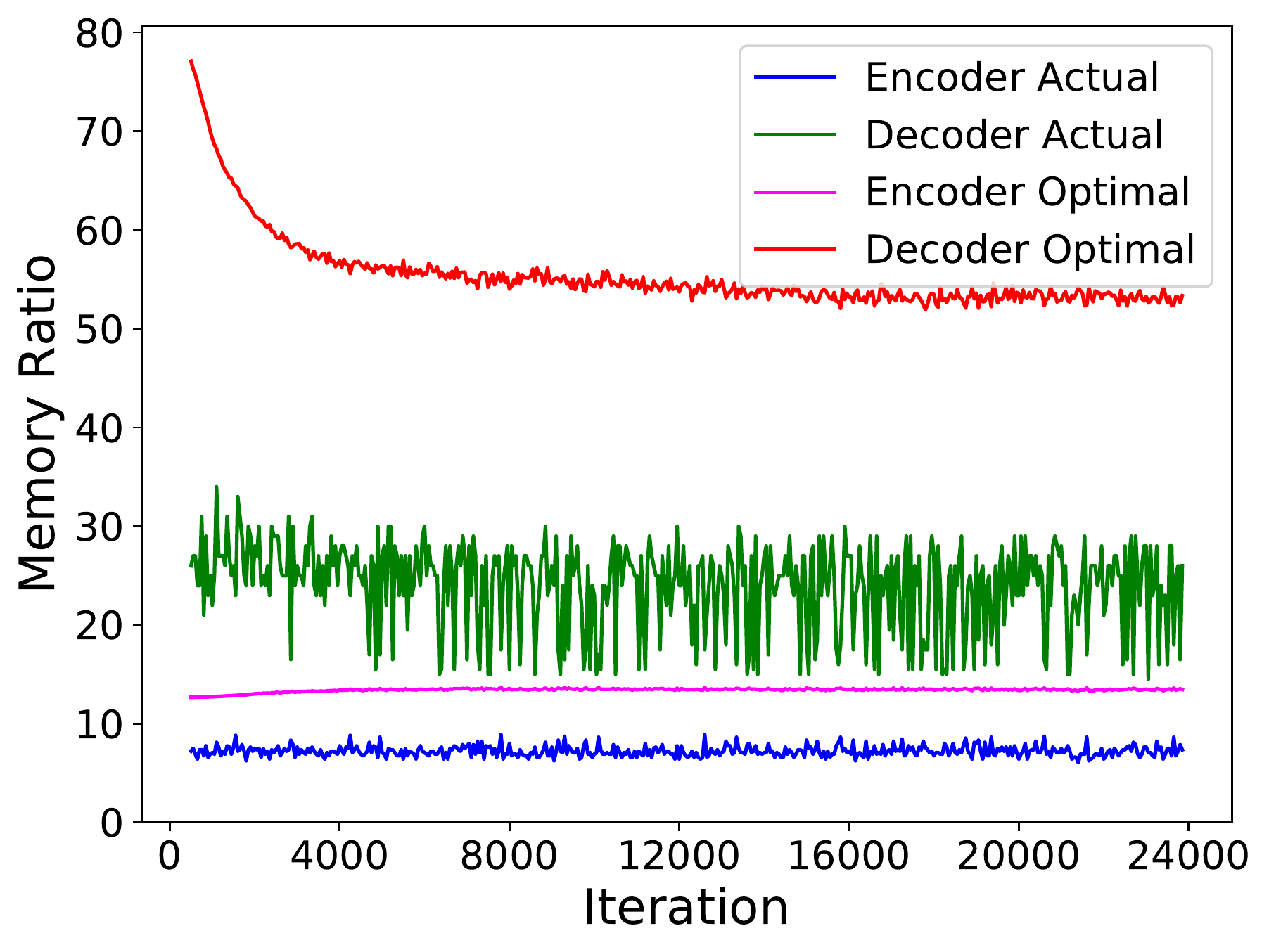}
    \includegraphics[width=0.32\linewidth]{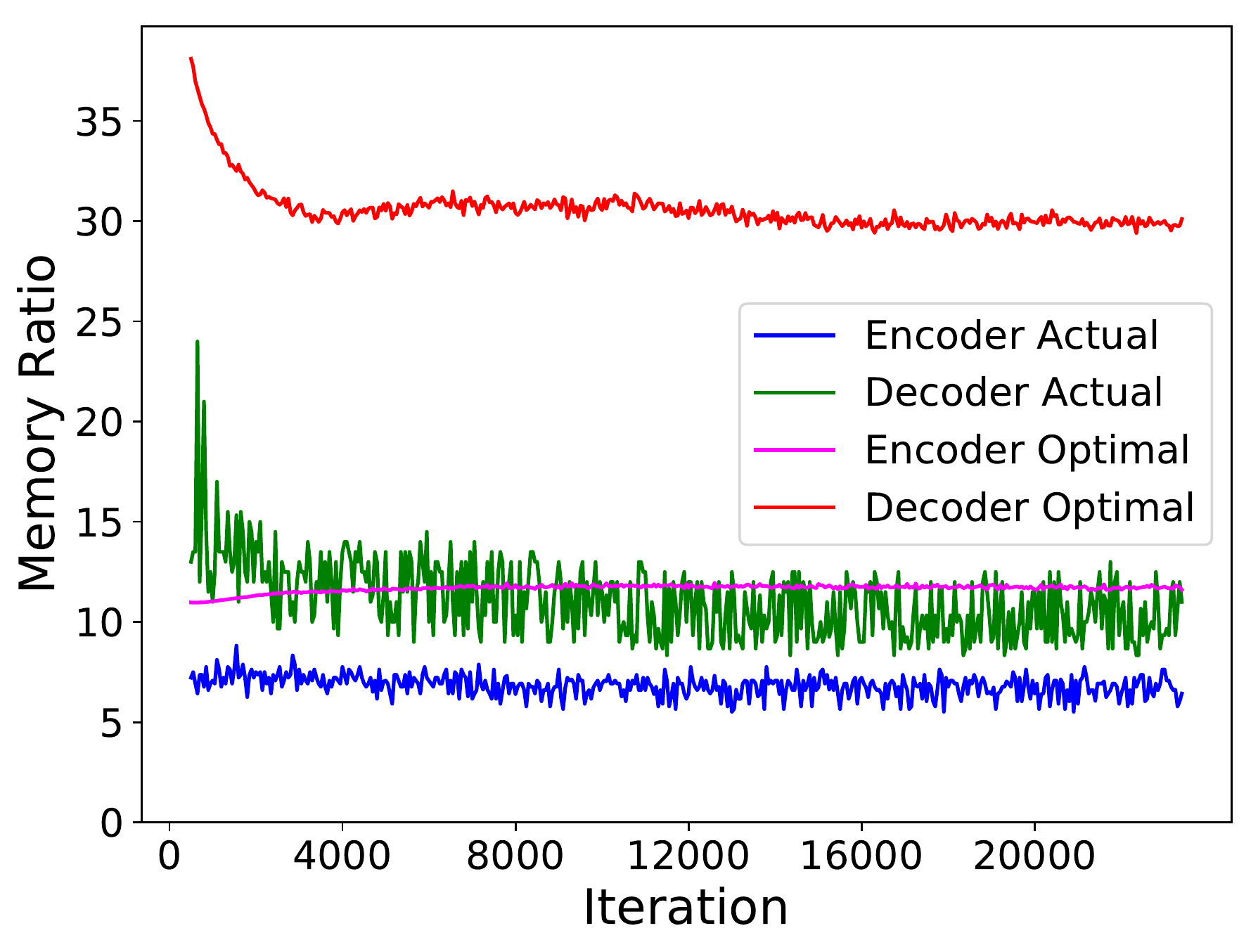}
    \includegraphics[width=0.32\linewidth]{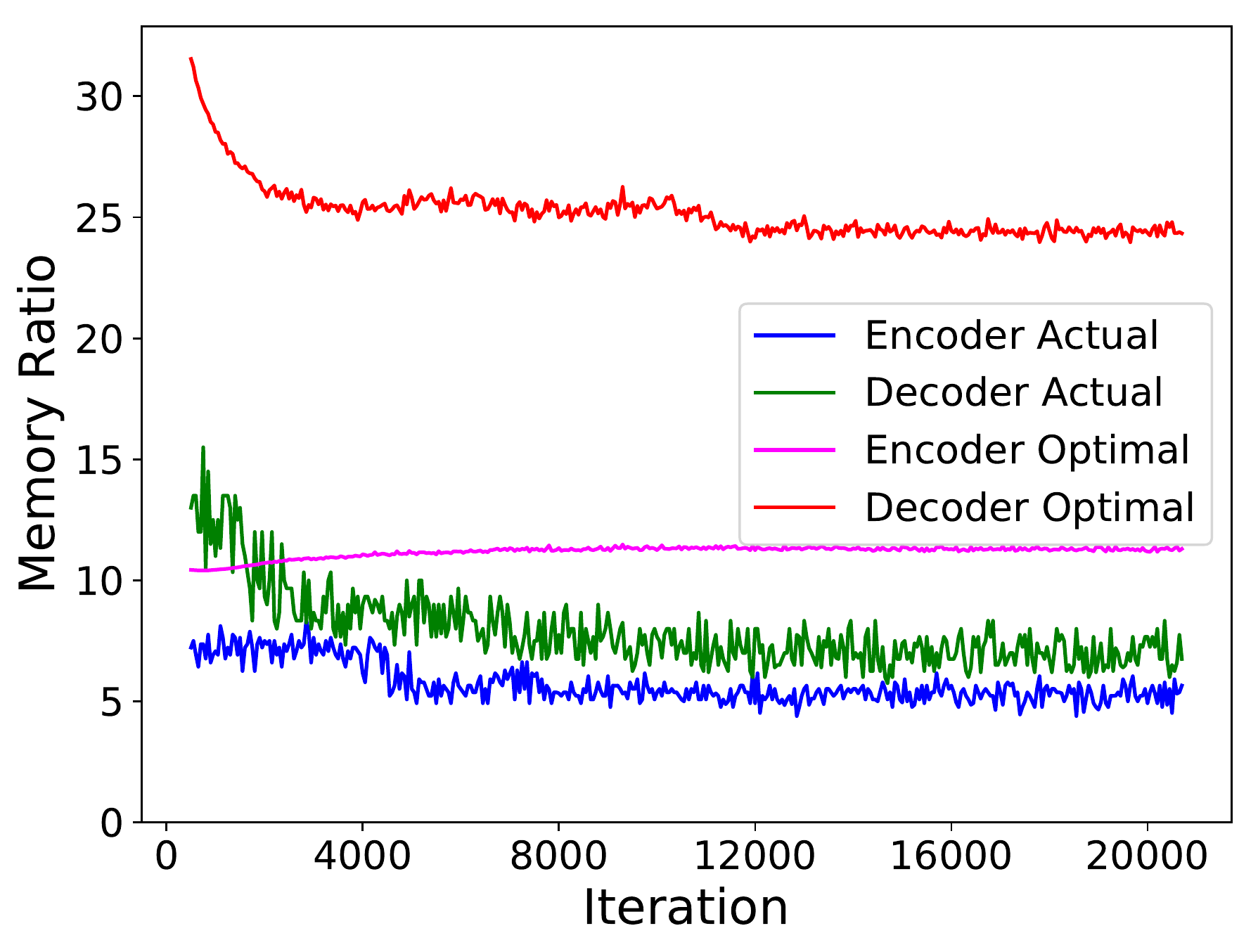}
    \label{fig:revgrumemorysavings}
    \caption{\textbf{RevGRU 20H.} From left to right: 1 bit, 3 bits, and no limit on forgetting.}
\end{figure}

\begin{figure}[H]
    \centering
    \includegraphics[width=0.32\linewidth]{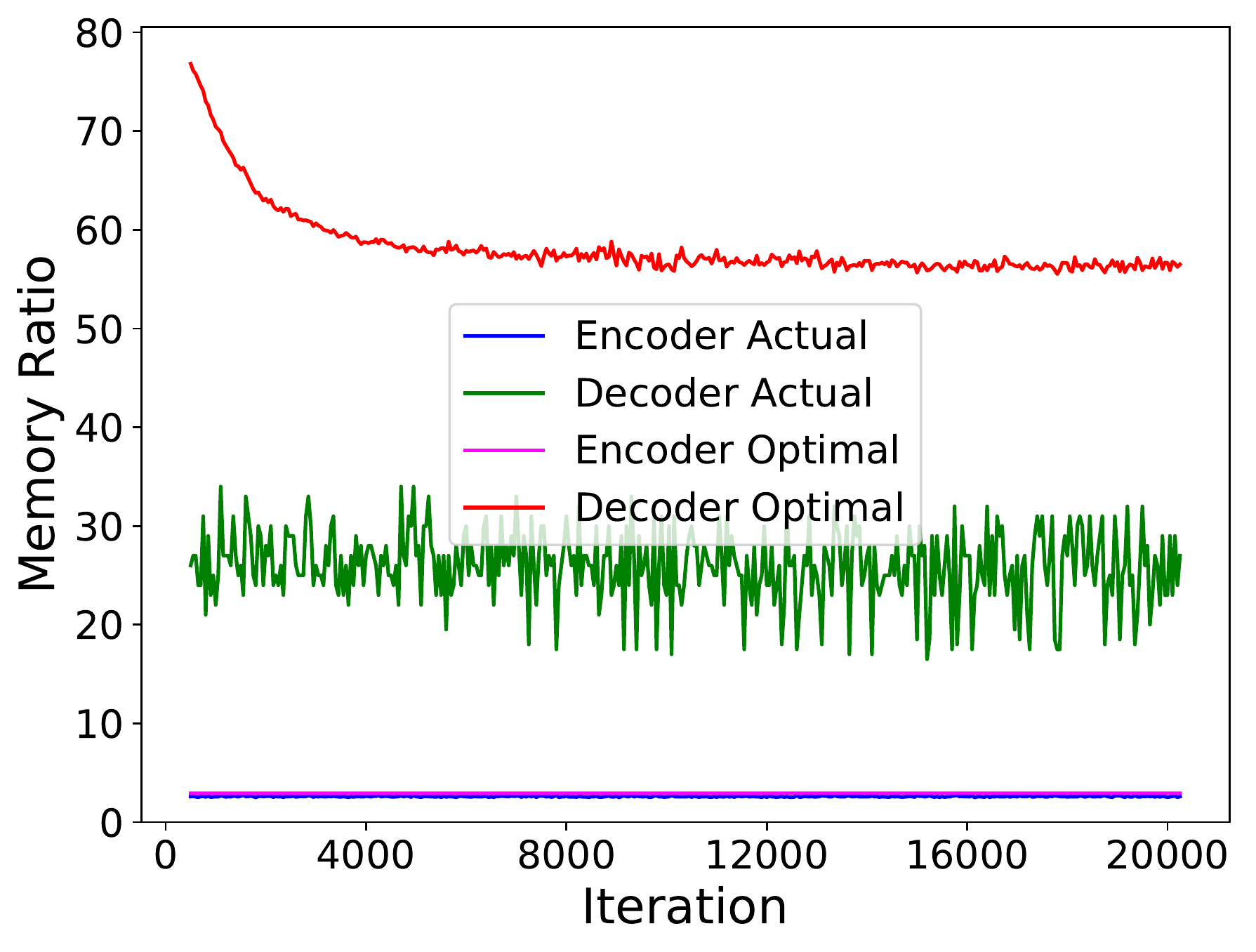}
    \includegraphics[width=0.32\linewidth]{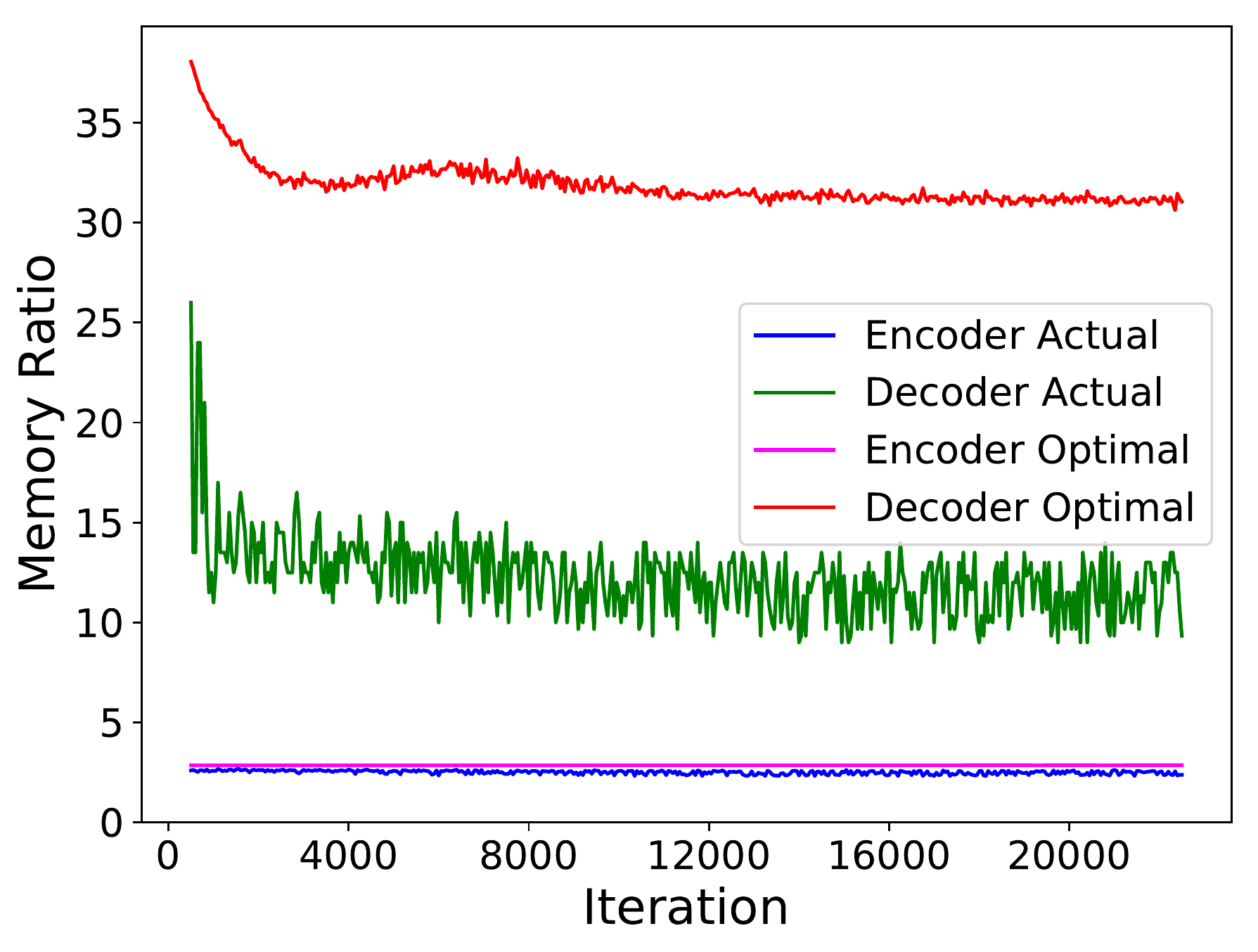}
    \includegraphics[width=0.32\linewidth]{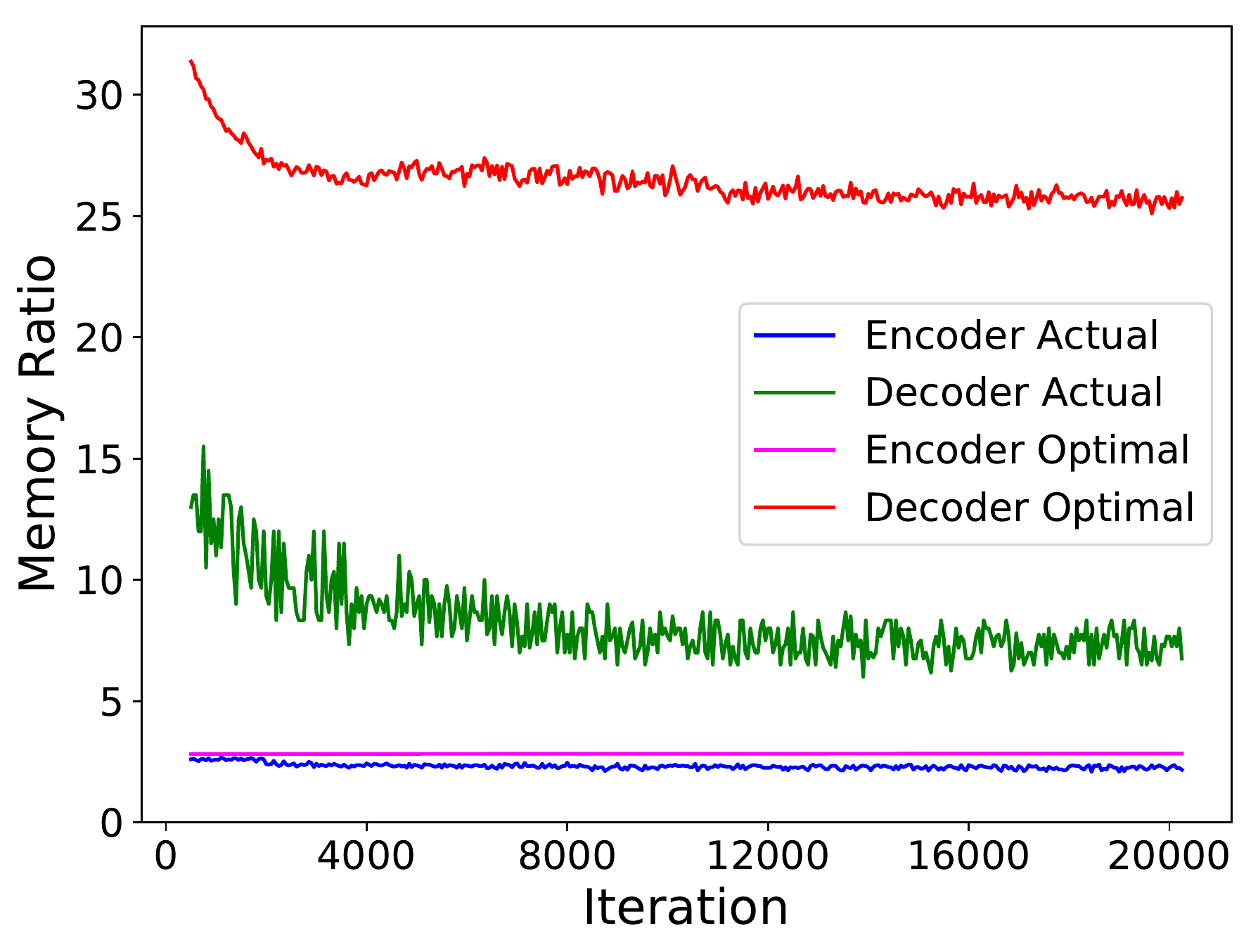}
    \label{fig:revgrumemorysavings}
    \caption{\textbf{RevGRU 100H.}  From left to right, 1 bit, 3 bits, and no limit on forgetting.}
\end{figure}

\begin{figure}[H]
    \centering
    \includegraphics[width=0.32\linewidth]{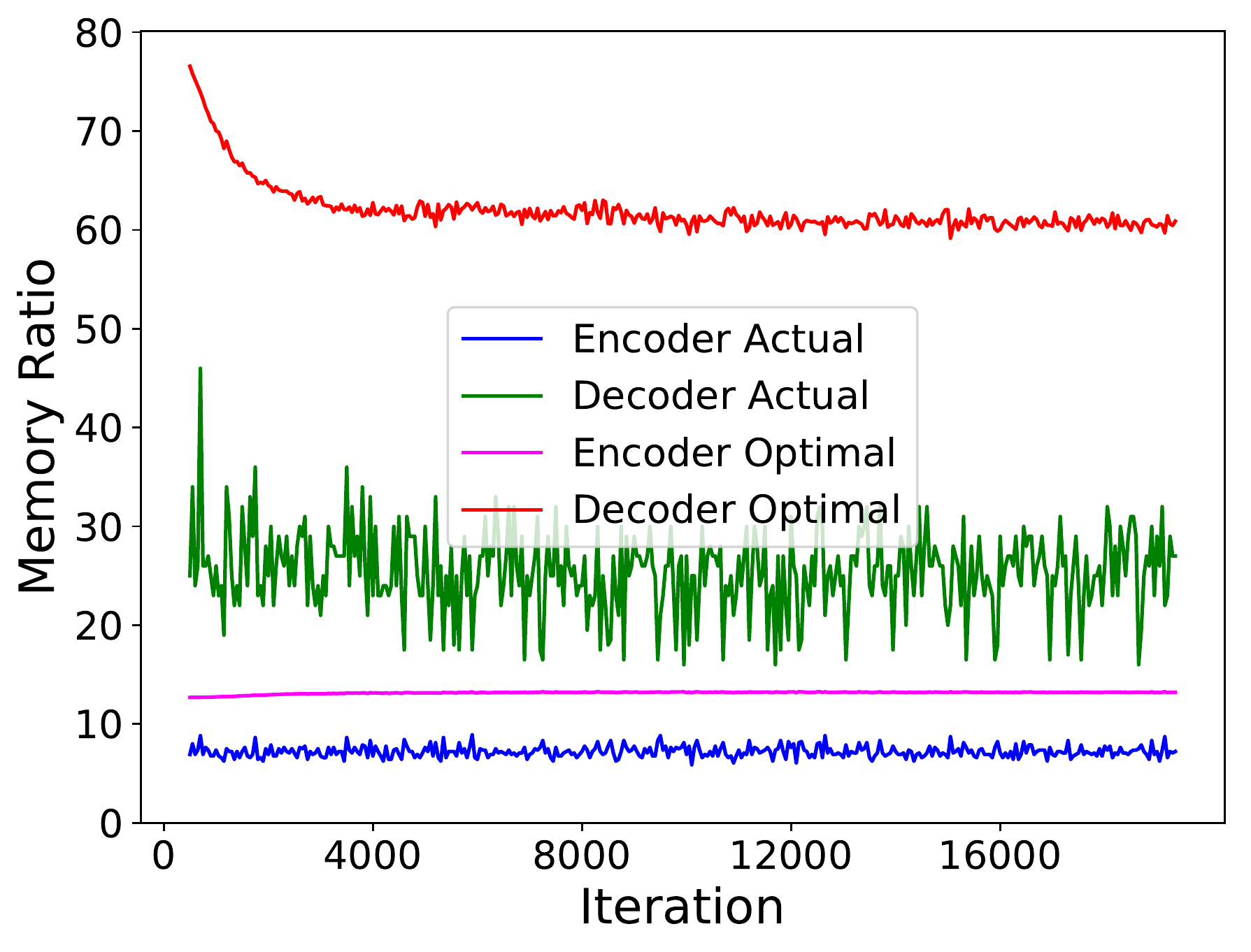}
    \includegraphics[width=0.32\linewidth]{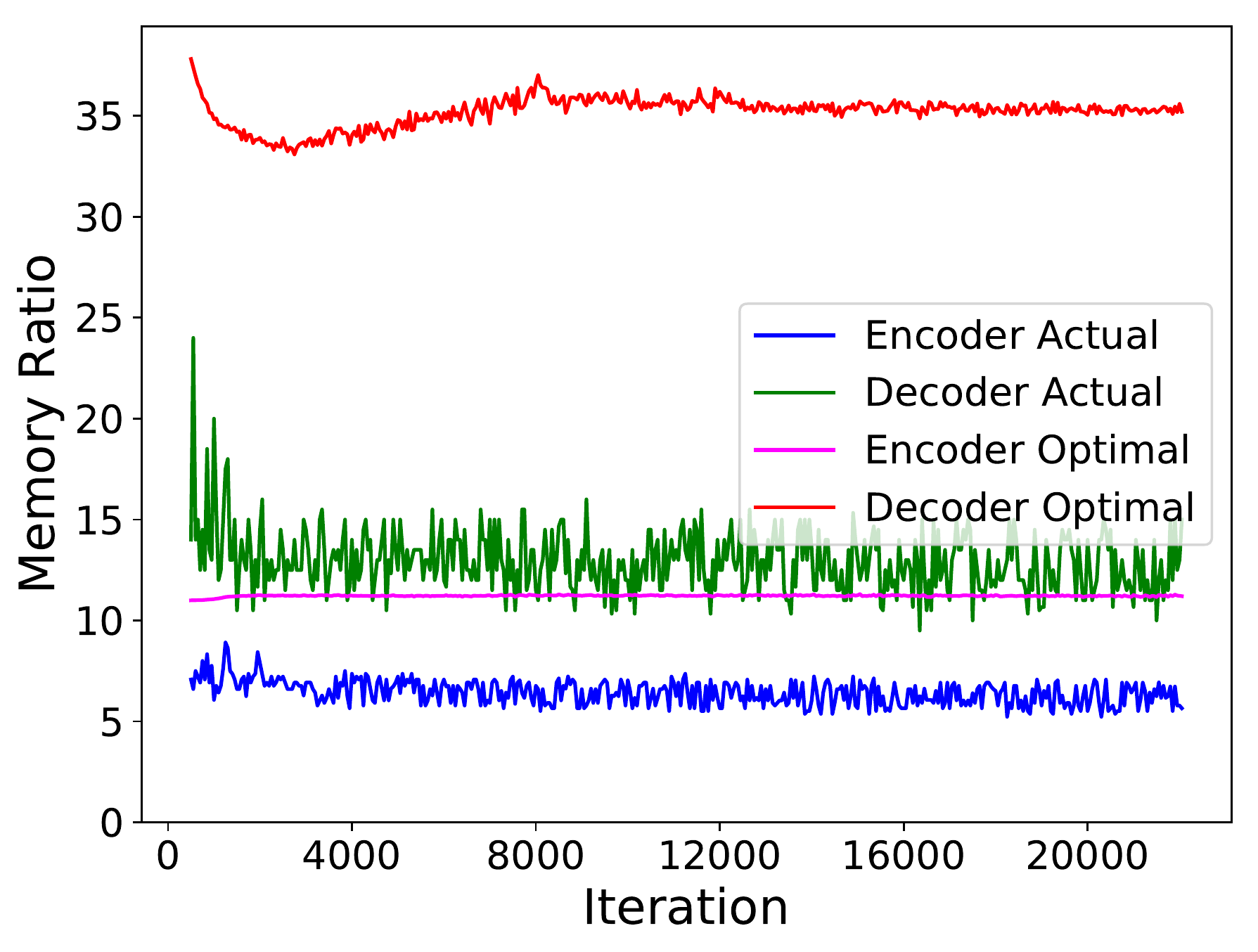}
    \includegraphics[width=0.32\linewidth]{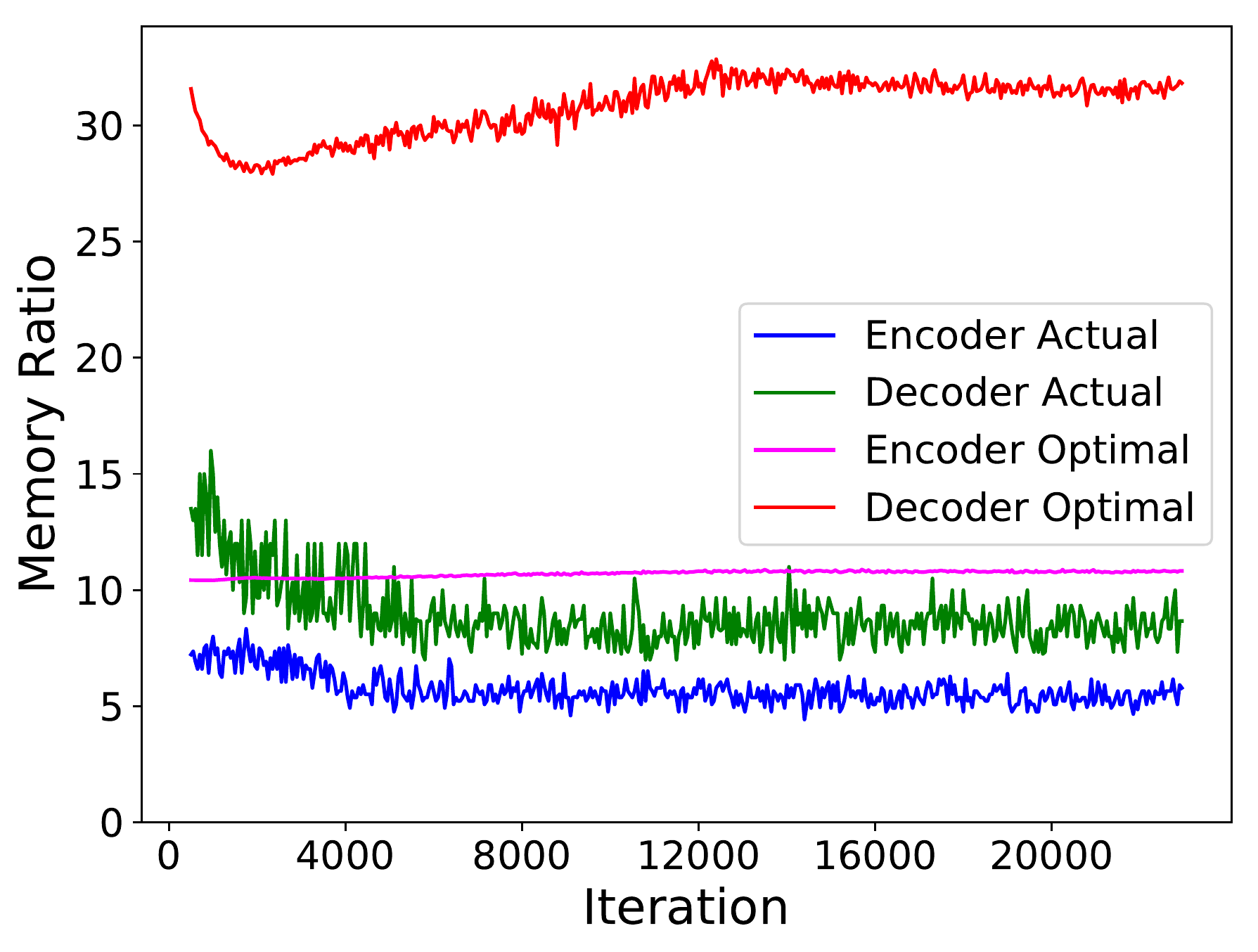}
    \label{fig:revgrumemorysavings}
    \caption{\textbf{RevGRU Emb+20H.} From left to right: 1 bit, 3 bits, and no limit on forgetting.}
\end{figure}

\subsubsection{RevLSTM on Multi30K}
\label{app:revlstm-mem-plots}

\begin{figure}[H]
    \centering
    \includegraphics[width=0.32\linewidth]{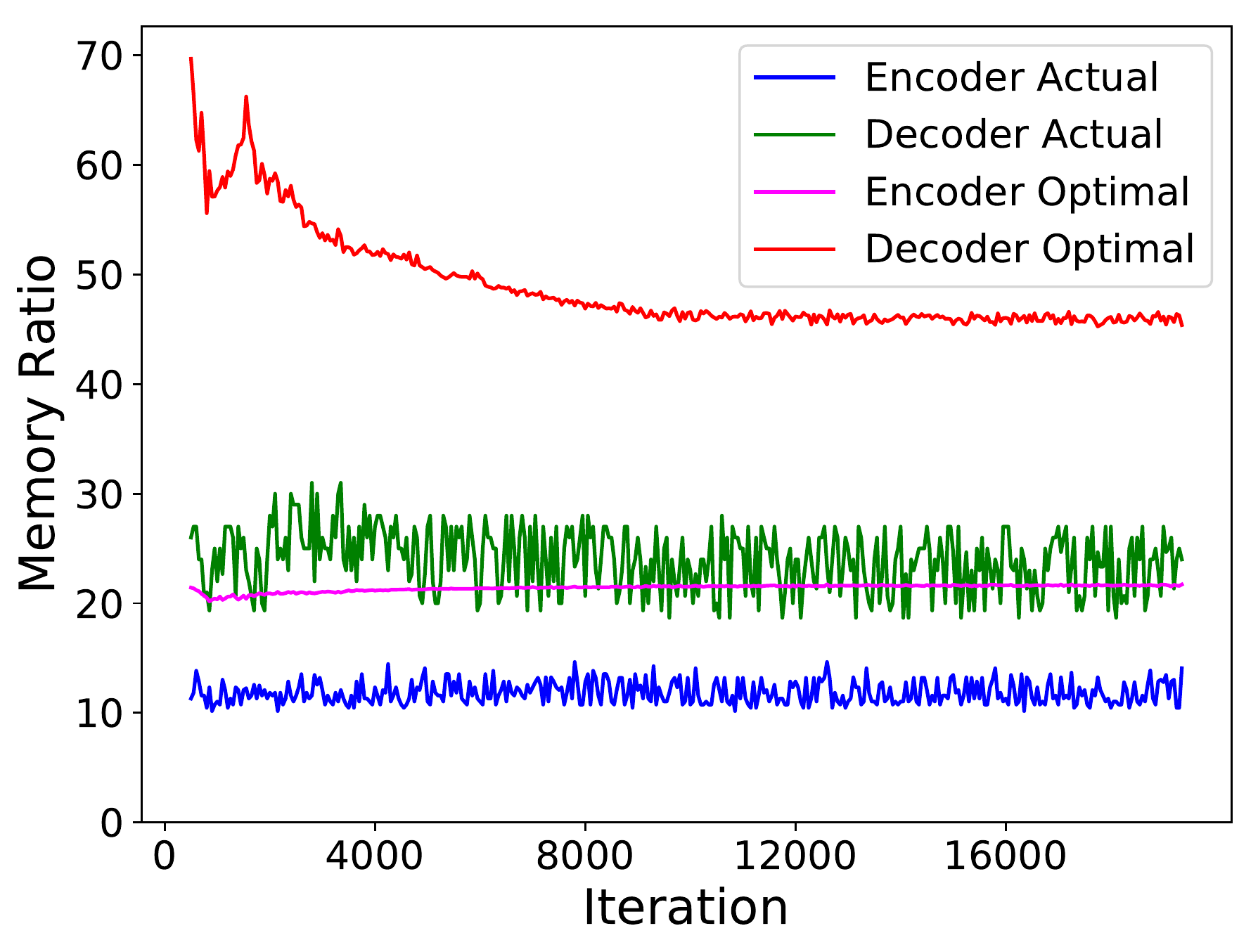}
    \includegraphics[width=0.32\linewidth]{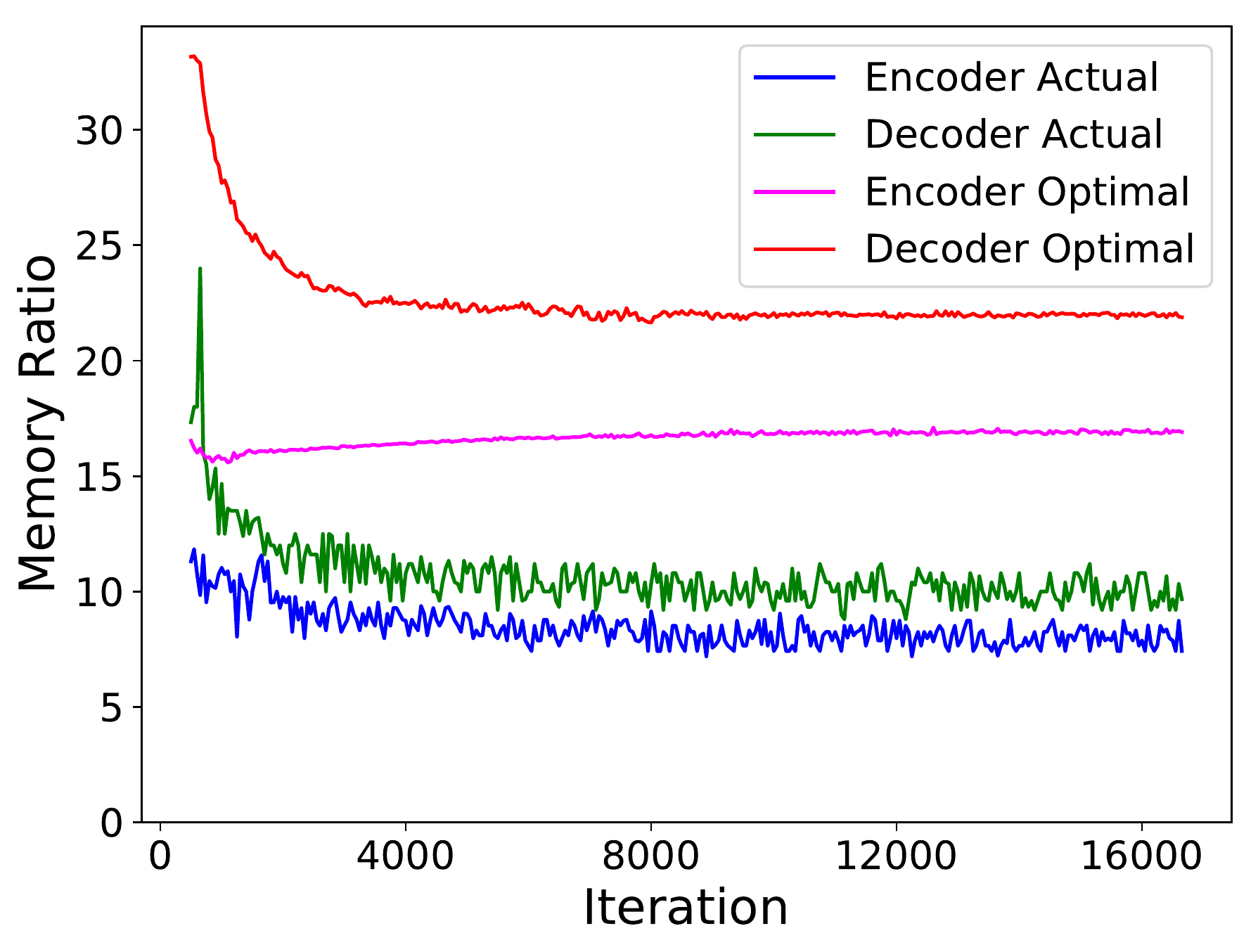}
    \includegraphics[width=0.32\linewidth]{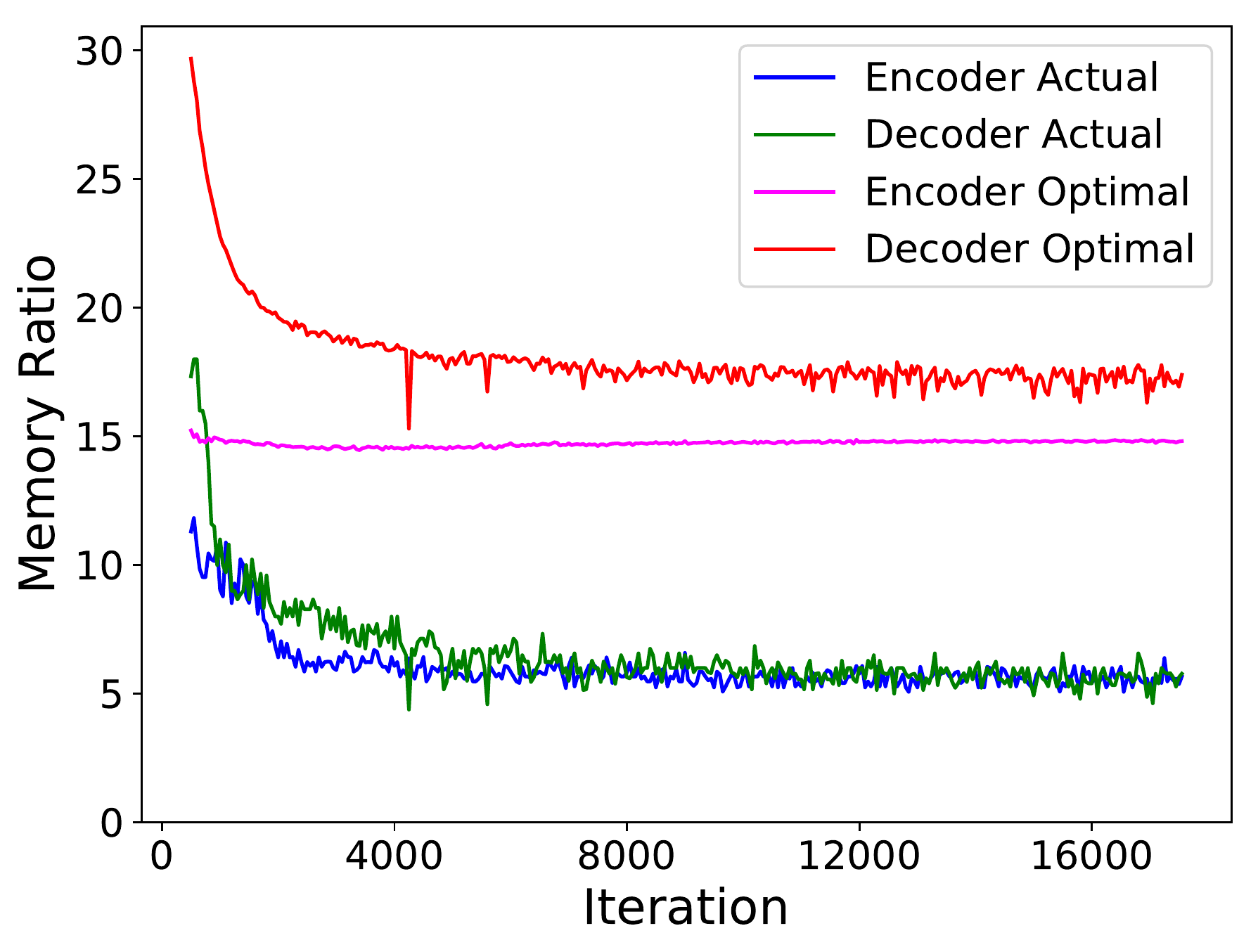}
    \label{fig:revgrumemorysavings}
    \caption{\textbf{RevLSTM 20H.} From left to right: 1 bit, 3 bits, and no limit on forgetting.}
\end{figure}

\begin{figure}[H]
    \centering
    \includegraphics[width=0.32\linewidth]{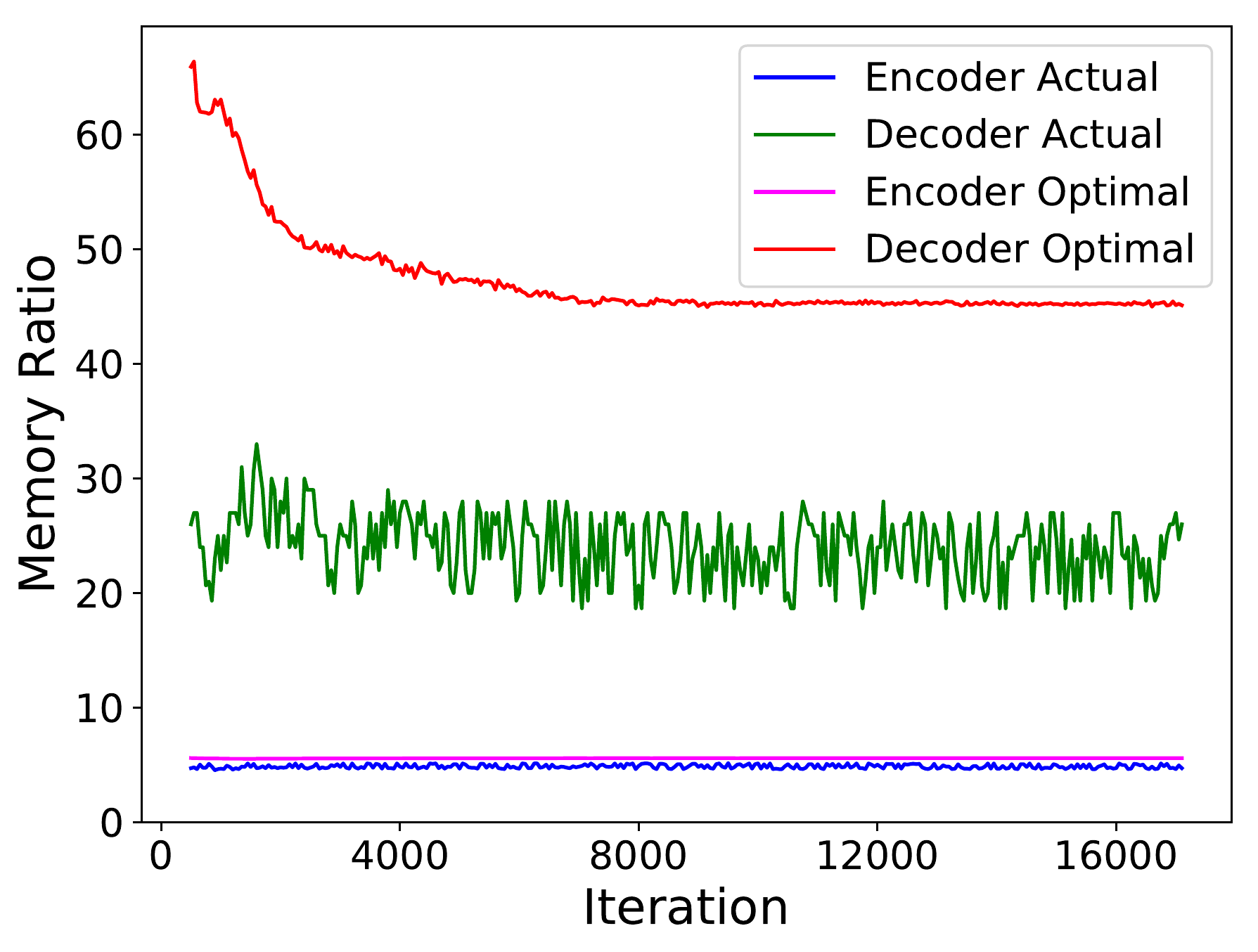}
    \includegraphics[width=0.32\linewidth]{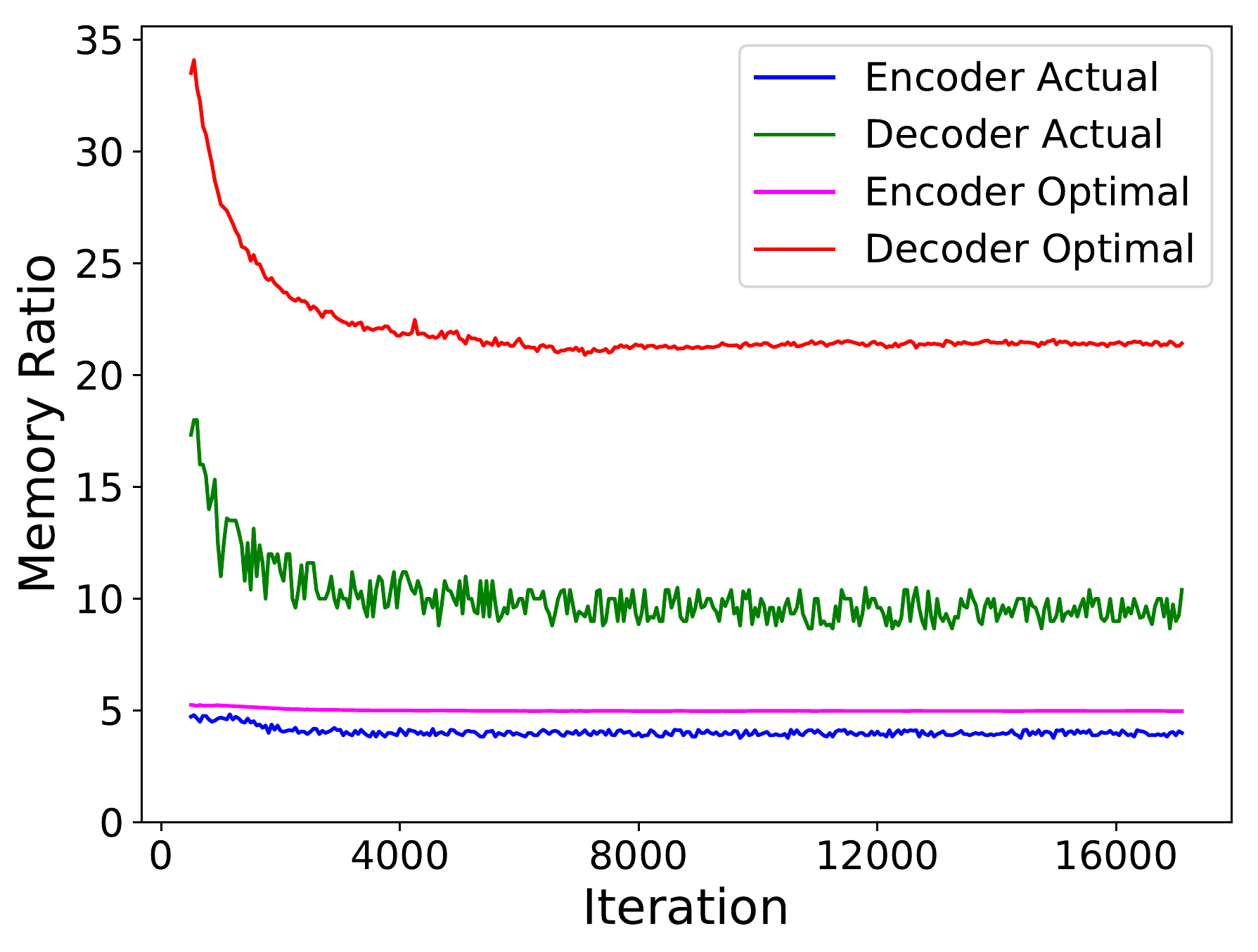}
    \includegraphics[width=0.32\linewidth]{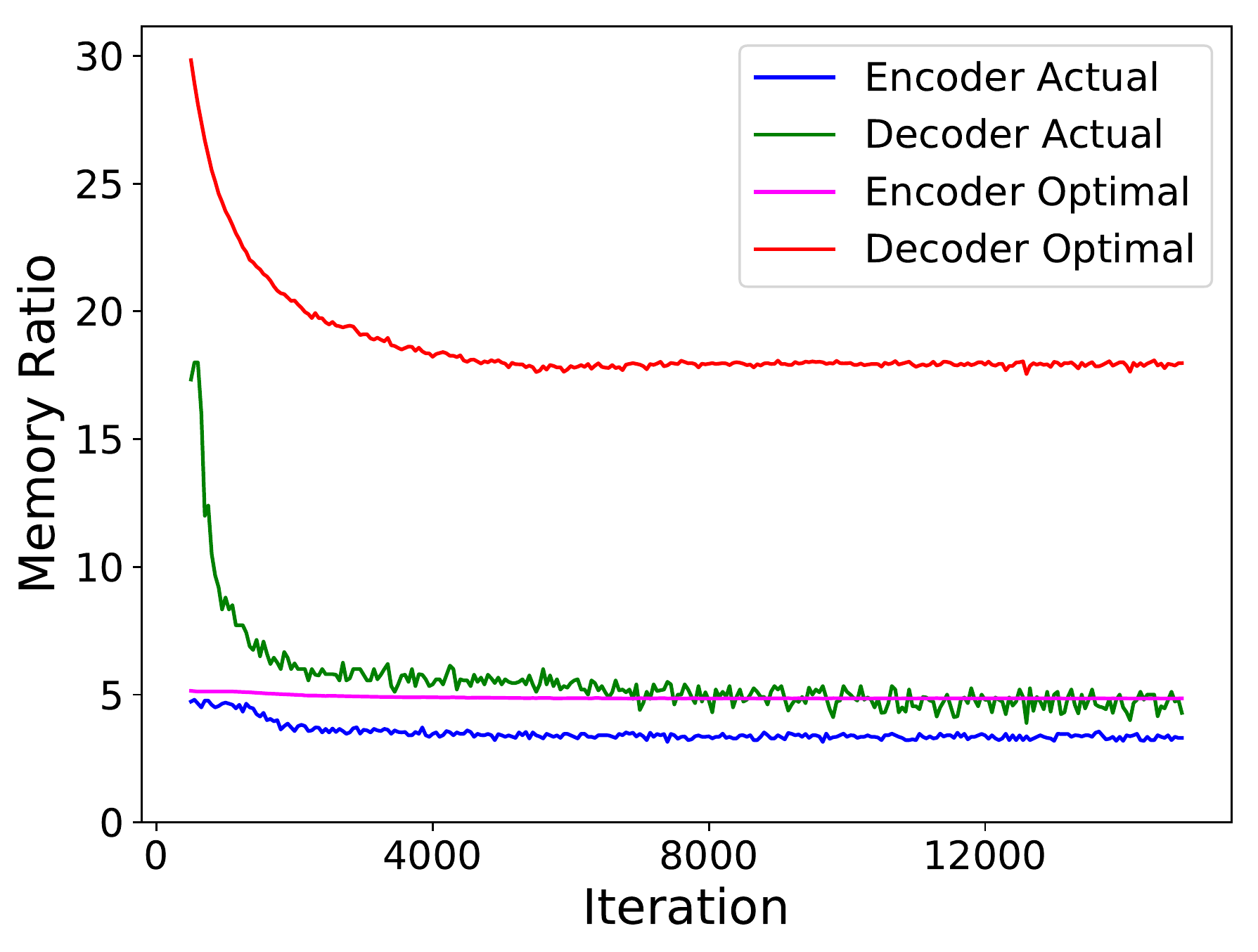}
    \label{fig:revgrumemorysavings}
    \caption{\textbf{RevLSTM 100H.} From left to right: 1 bit, 3 bits, and no limit on forgetting.}
\end{figure}

\begin{figure}[H]
    \centering
    \includegraphics[width=0.32\linewidth]{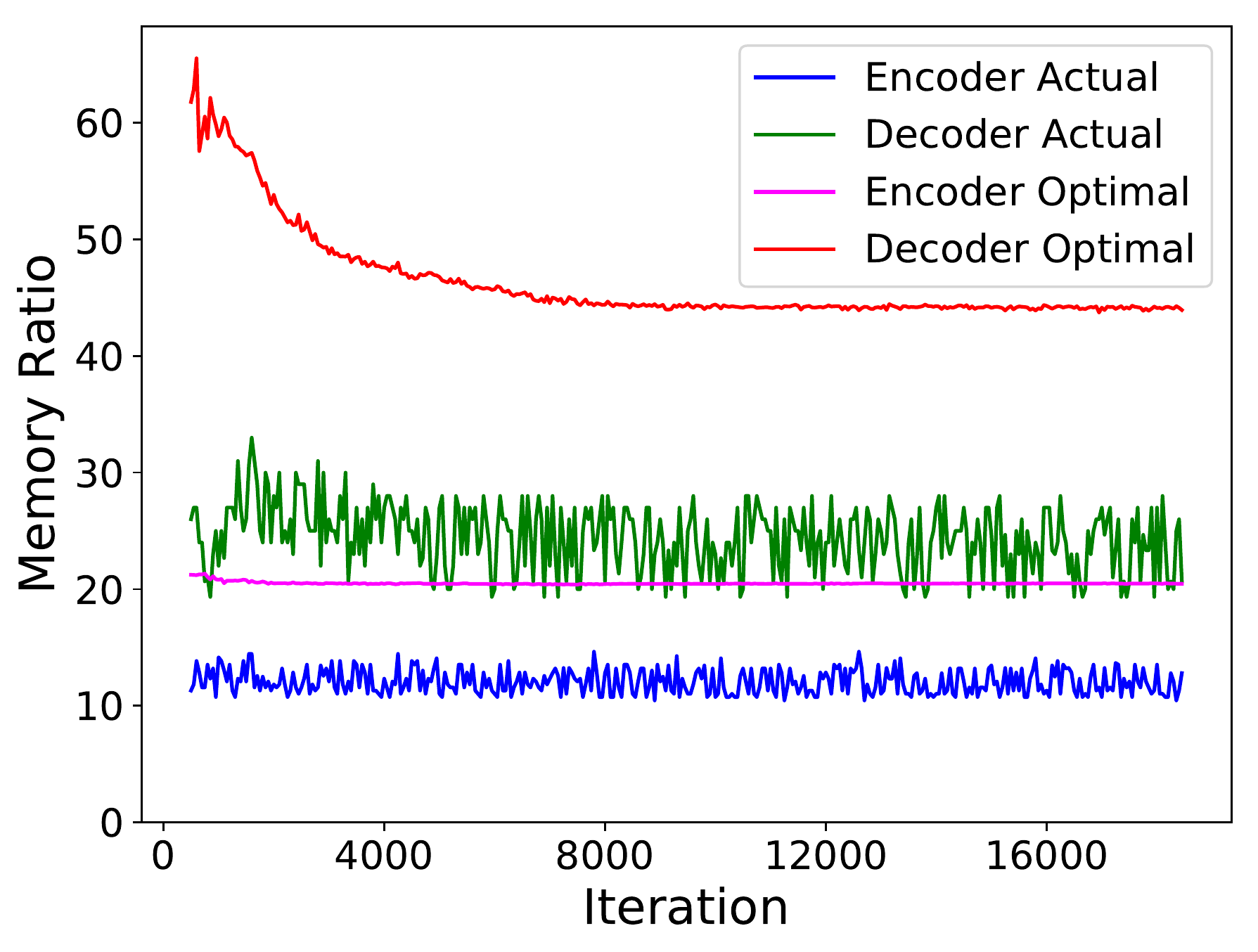}
    \includegraphics[width=0.32\linewidth]{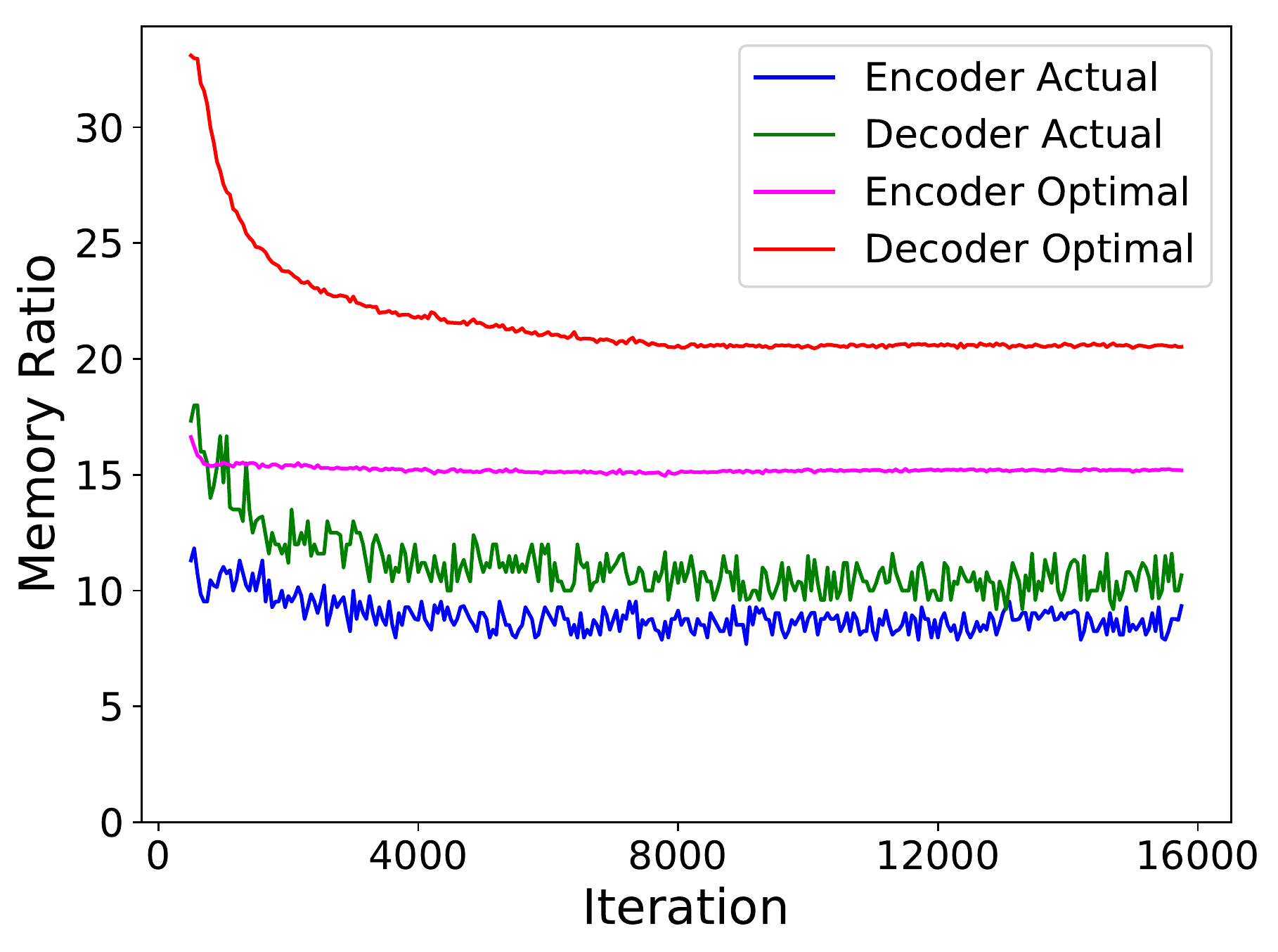}
    \includegraphics[width=0.32\linewidth]{paper_plots/nmt_figures/nmt_memory/RevLSTM_100H_nolim.pdf}
    \label{fig:revgrumemorysavings}
    \caption{\textbf{RevLSTM Emb+20H.} From left to right: 1 bit, 3 bits, and no limit on forgetting.}
\end{figure}

\subsubsection{RevGRU and RevLSTM on IWSLT-2016}
\label{app:iwslt-mem-plots}

Here, we plot the memory savings achieved by our two-layer models on IWSLT-2016, as well as the ideal memory savings, for both the encoder and decoder.

\begin{figure}[H]
    \centering
    \includegraphics[width=0.48\linewidth]{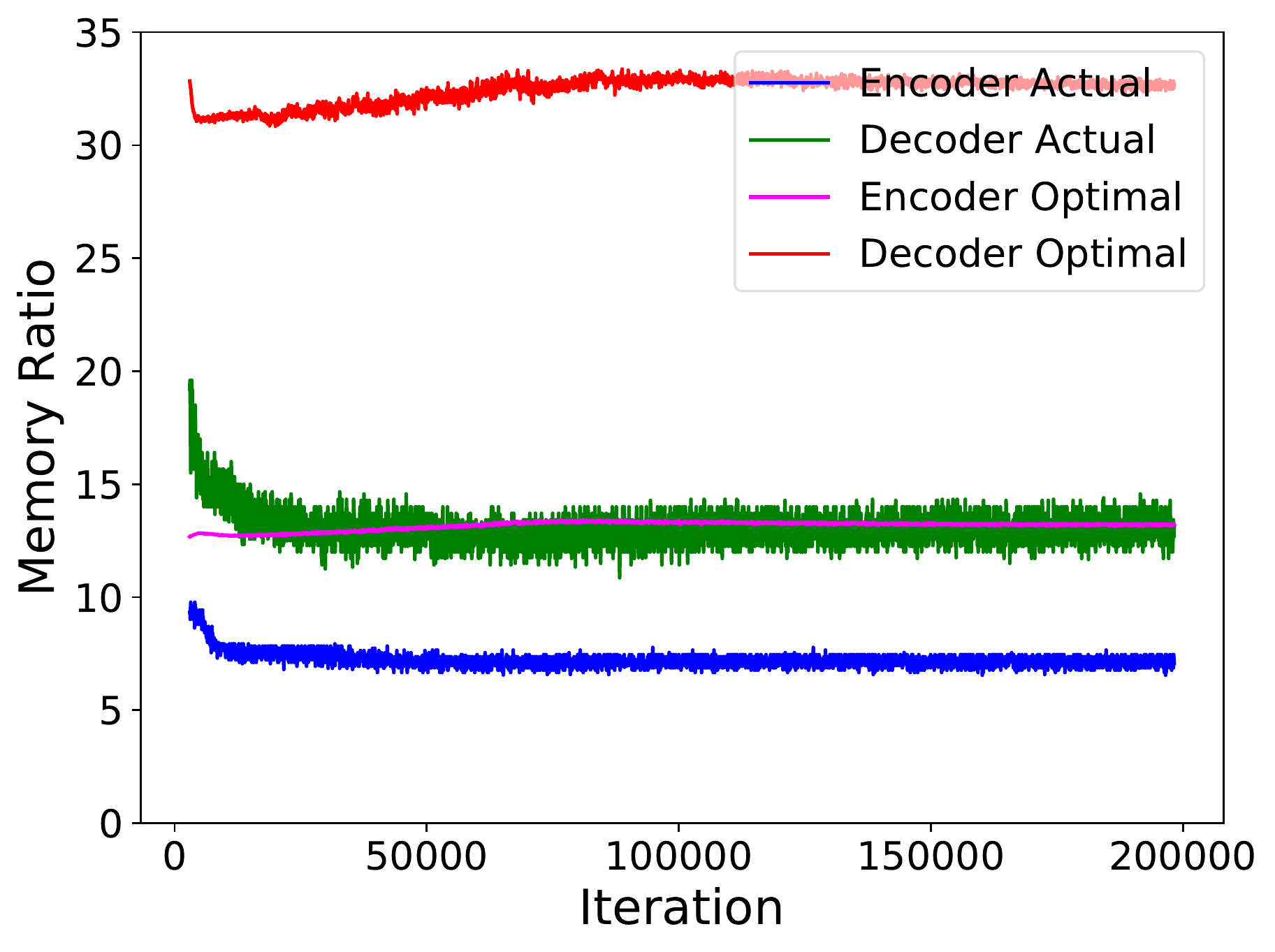}
    \includegraphics[width=0.47\linewidth]{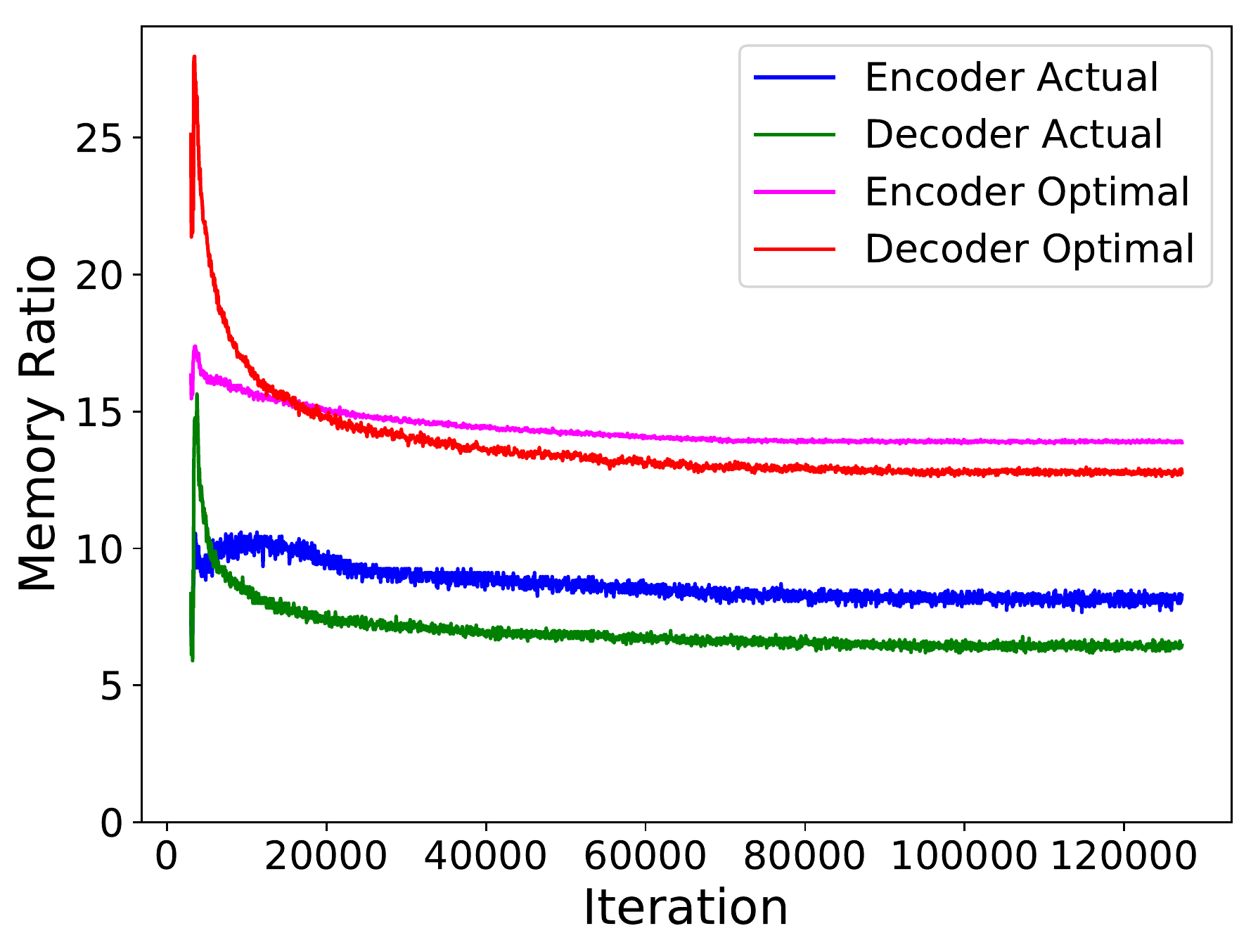}
    \caption{Memory savings on IWSLT. \textbf{Left:} RevGRU. \textbf{Right:} RevLSTM. Both models use attention over the concatenation of the word embeddings and a 60-dimensional slice of the hidden state.}
    \label{fig:iwslt-mem-plots}
\end{figure}

\section{Training/Validation Curves}
\label{app:train-valid-curves}

\subsection{Penn TreeBank}

\subsubsection*{1 layer RevGRU}
\begin{figure}[H]
  \caption*{Training/validation perplexity for a 1-layer RevGRU on Penn TreeBank with various restrictions on forgetting and a baseline GRU model. \textbf{Left:} Perplexity on the training set.  \textbf{Right:} Perplexity on the validation set.}
  \begin{center}
    \includegraphics[width=0.5\textwidth]{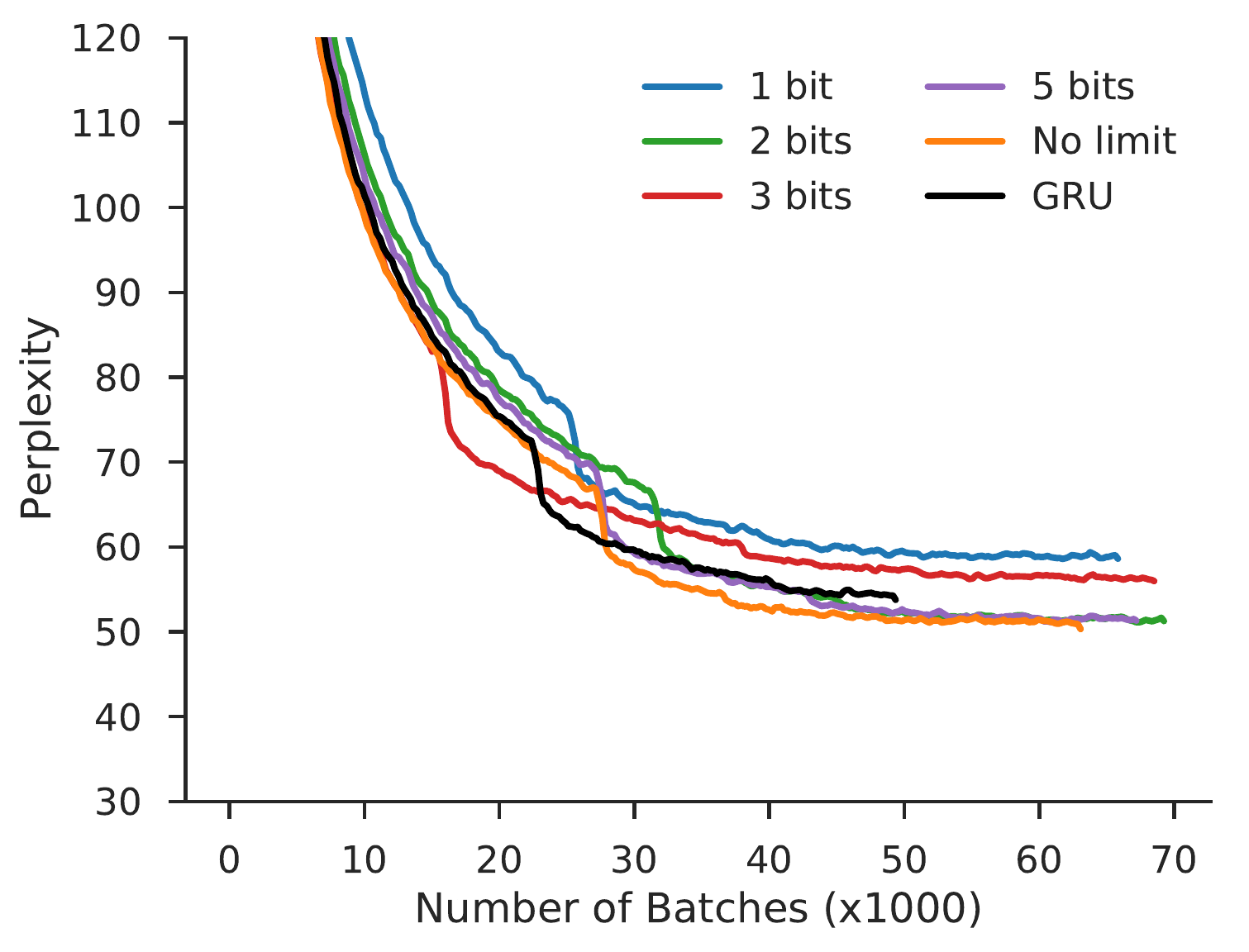}\includegraphics[width=0.5\textwidth]{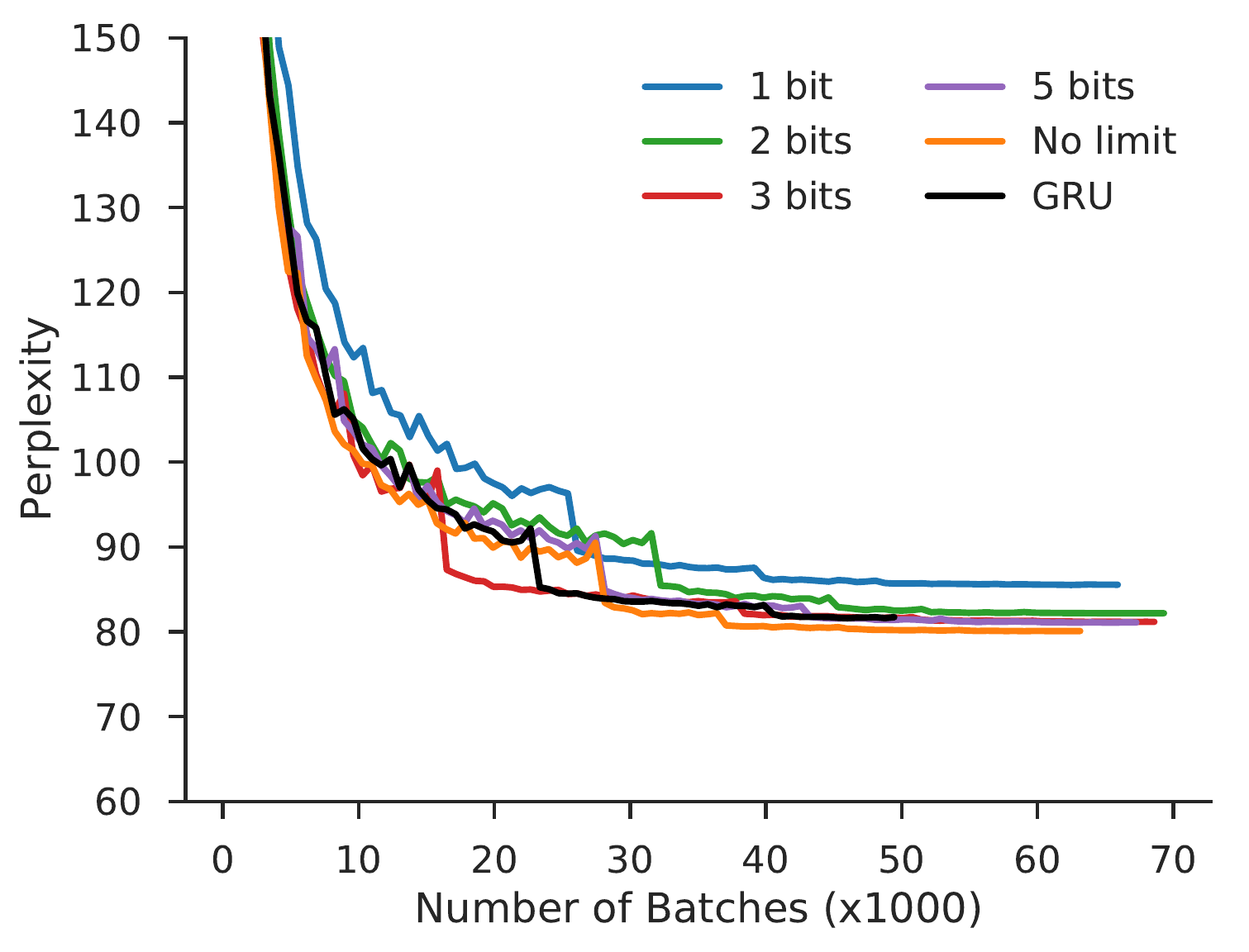}
  \end{center}
  \label{fig:single-dropout}
\end{figure}

\subsubsection*{2 layer RevGRU}
\begin{figure}[H]
  \caption*{Training/validation perplexity for a 2-layer RevGRU on Penn TreeBank with various restrictions on forgetting and a baseline GRU model. \textbf{Left:} Perplexity on the training set.  \textbf{Right:} Perplexity on the validation set.}
  \begin{center}
    \includegraphics[width=0.5\textwidth]{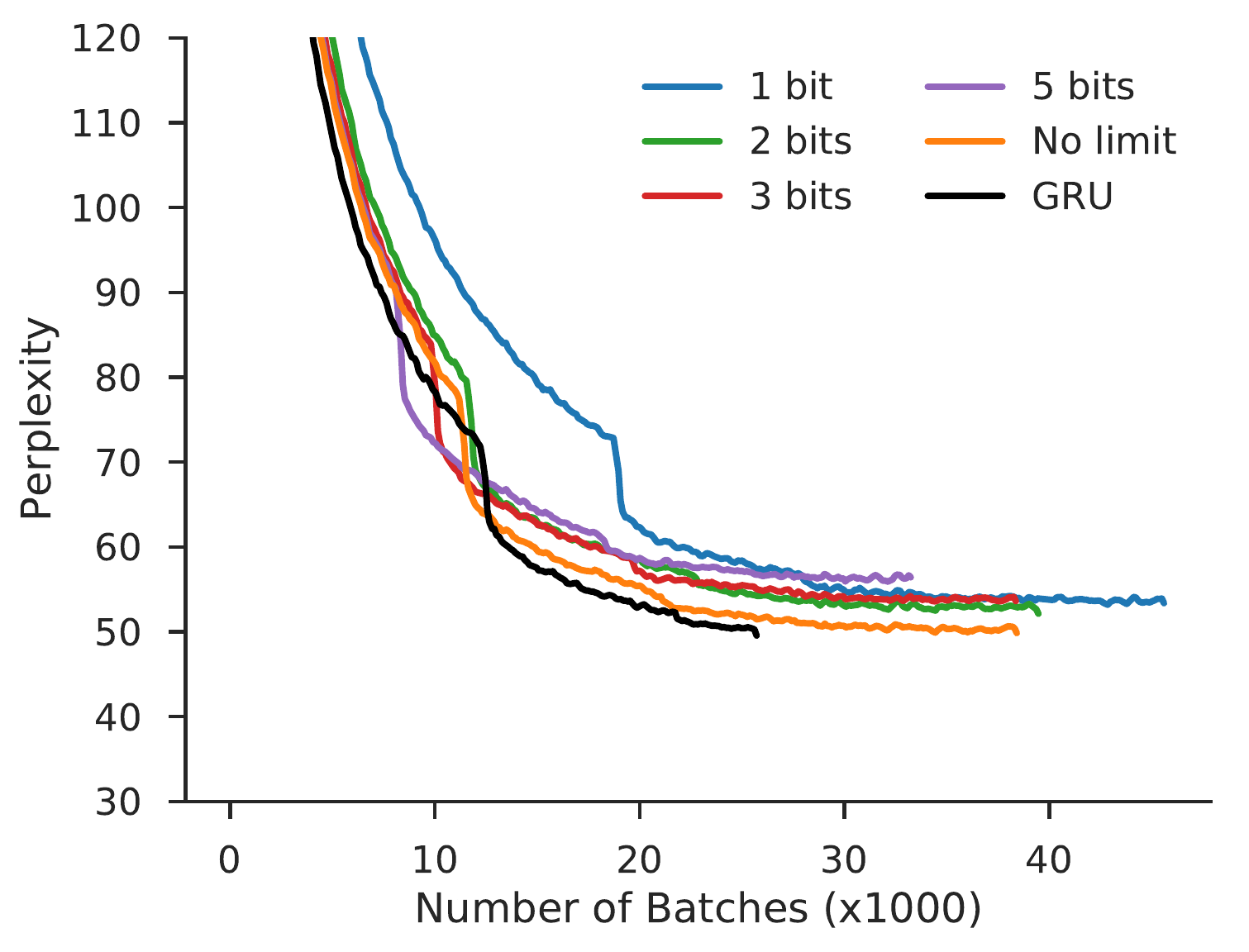}\includegraphics[width=0.5\textwidth]{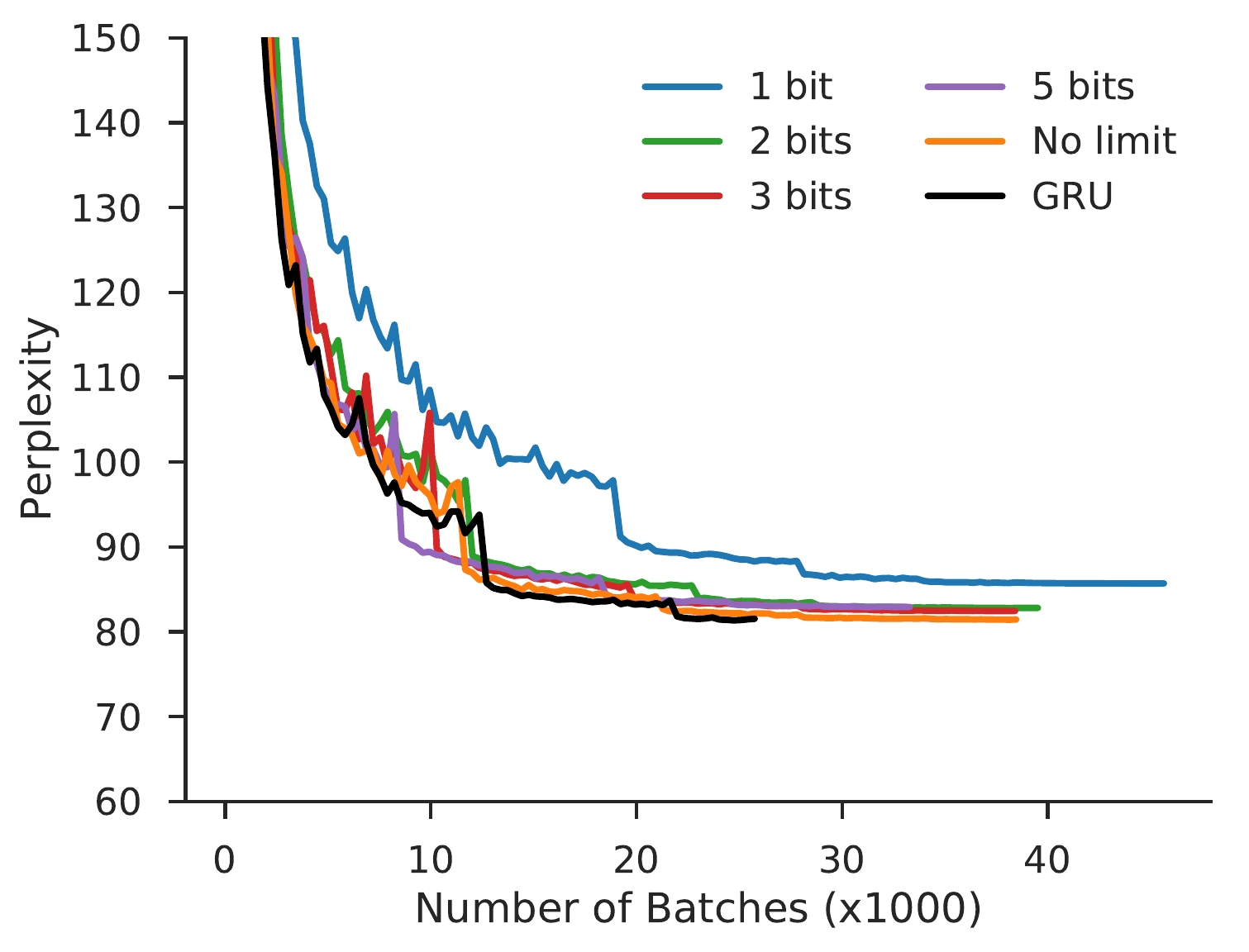}
  \end{center}
  \label{fig:single-dropout}
\end{figure}

\subsubsection*{1 layer RevLSTM}
\begin{figure}[H]
  \caption*{Training/validation perplexity for a 1-layer RevLSTM on Penn TreeBank with various restrictions on forgetting and a baseline LSTM model. \textbf{Left:} Perplexity on the training set.  \textbf{Right:} Perplexity on the validation set.}
  \begin{center}
    \includegraphics[width=0.5\textwidth]{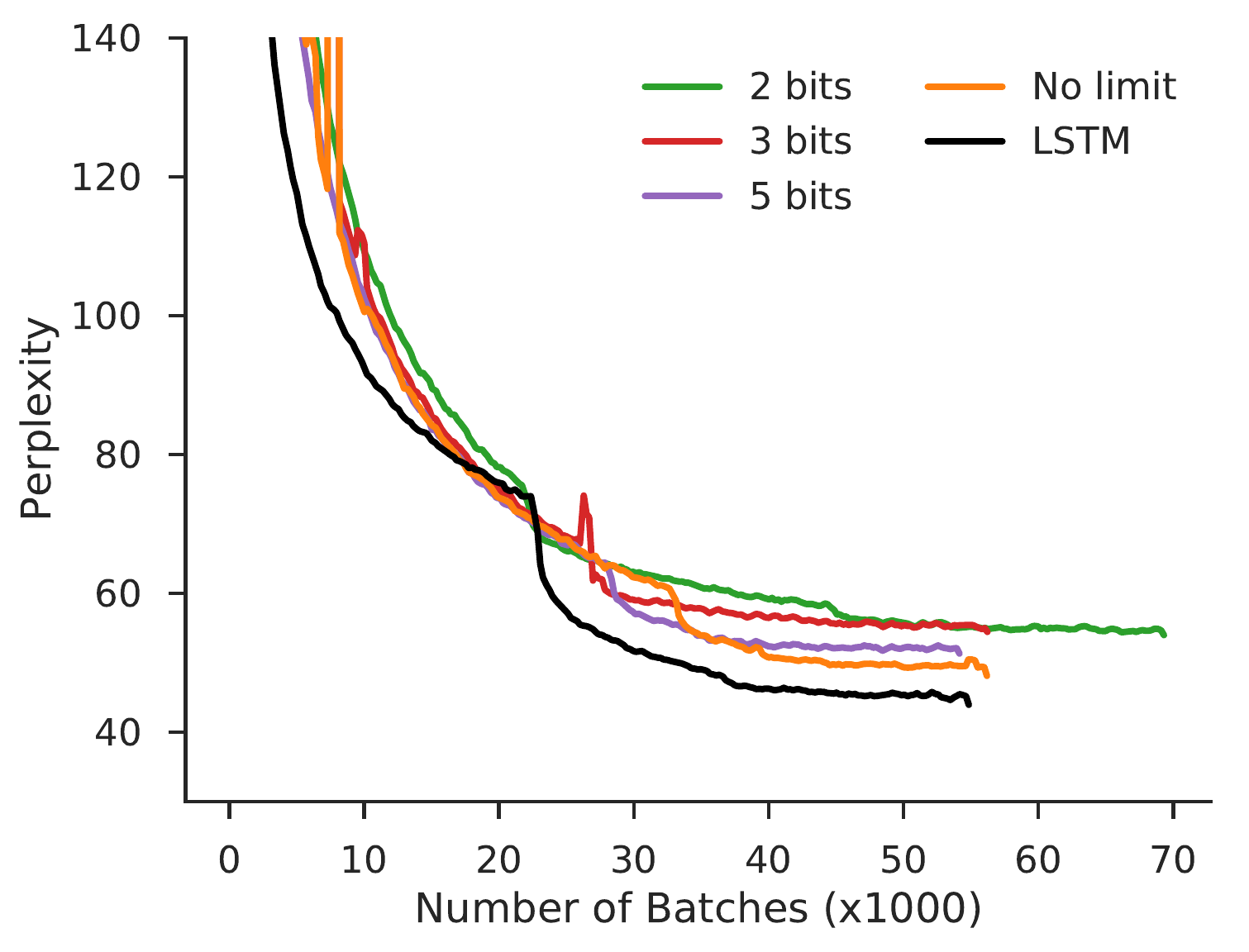}\includegraphics[width=0.5\textwidth]{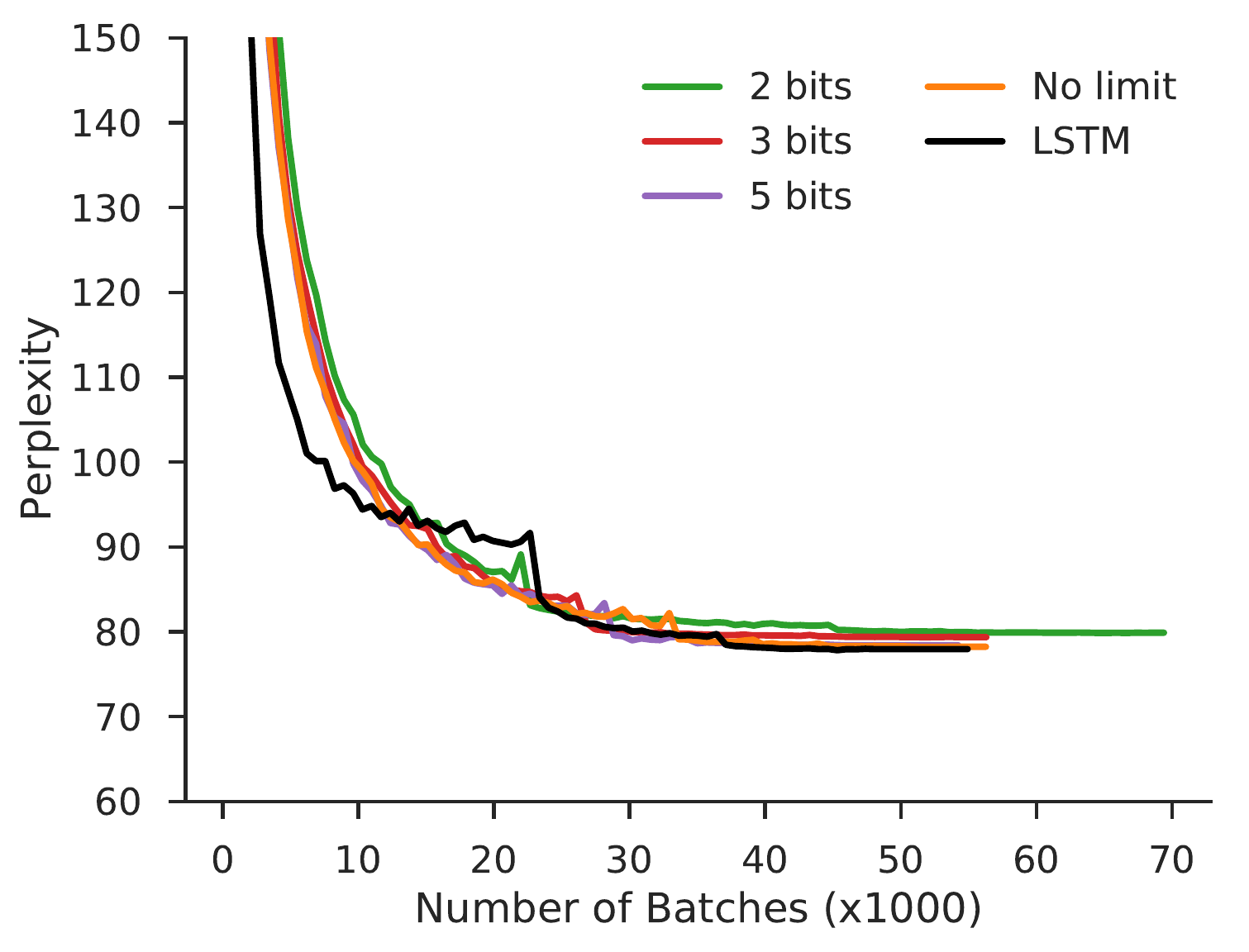}
  \end{center}
  \label{fig:single-dropout}
\end{figure}

\subsubsection*{2 layer RevLSTM}
\begin{figure}[H]
  \caption*{Training/validation perplexity for a 1-layer RevLSTM on Penn TreeBank with various restrictions on forgetting and a baseline LSTM model. \textbf{Left:} Perplexity on the training set.  \textbf{Right:} Perplexity on the validation set.}
  \begin{center}
    \includegraphics[width=0.5\textwidth]{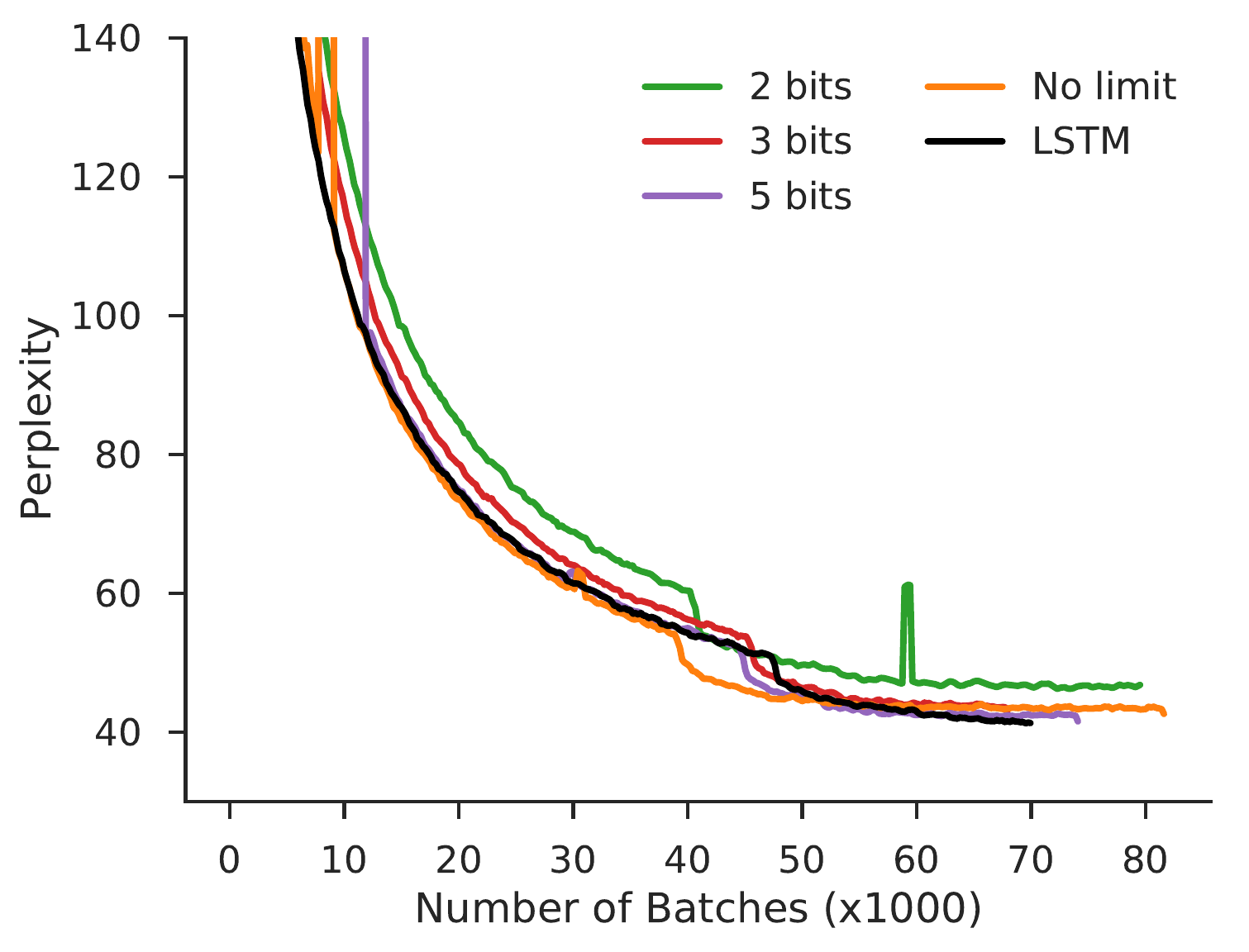}\includegraphics[width=0.5\textwidth]{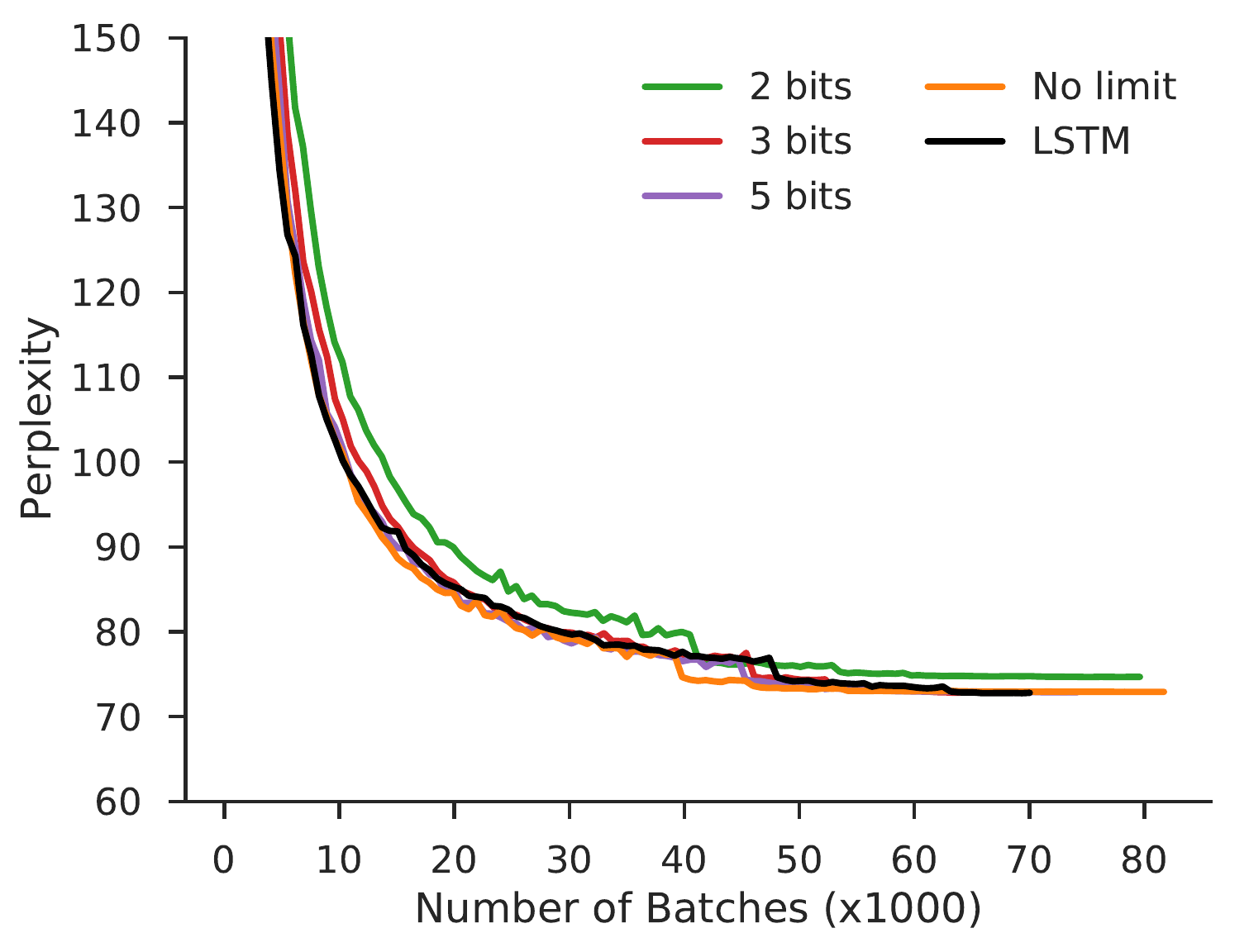}
  \end{center}
  \label{fig:single-dropout}
\end{figure}

\subsection{WikiText-2}

\subsubsection*{1 layer RevGRU}
\begin{figure}[H]
  \caption*{Training/validation perplexity for a 1-layer RevGRU on WikiText-2 with various restrictions on forgetting and a baseline GRU model. \textbf{Left:} Perplexity on the training set.  \textbf{Right:} Perplexity on the validation set.}
  \begin{center}
    \includegraphics[width=0.5\textwidth]{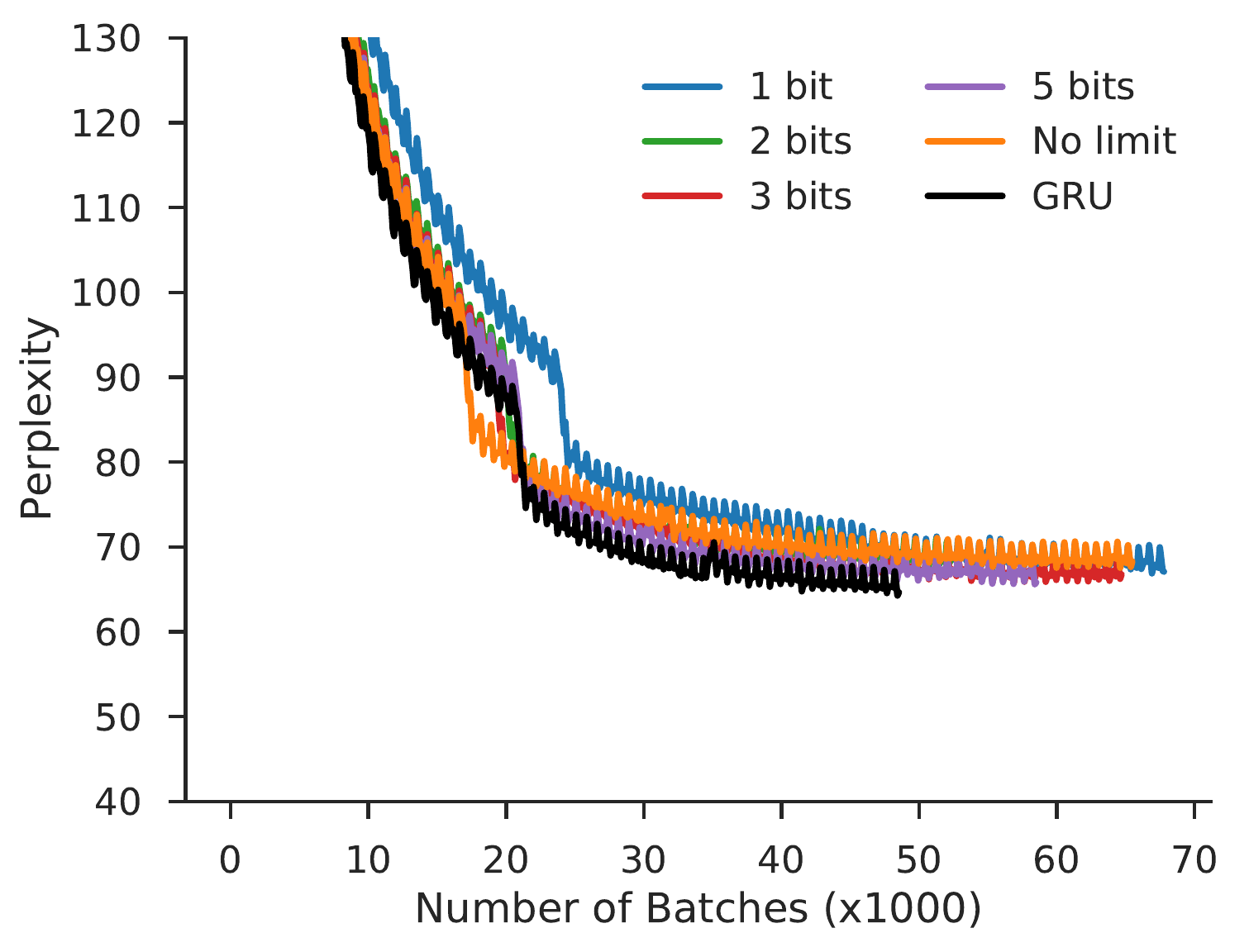}\includegraphics[width=0.5\textwidth]{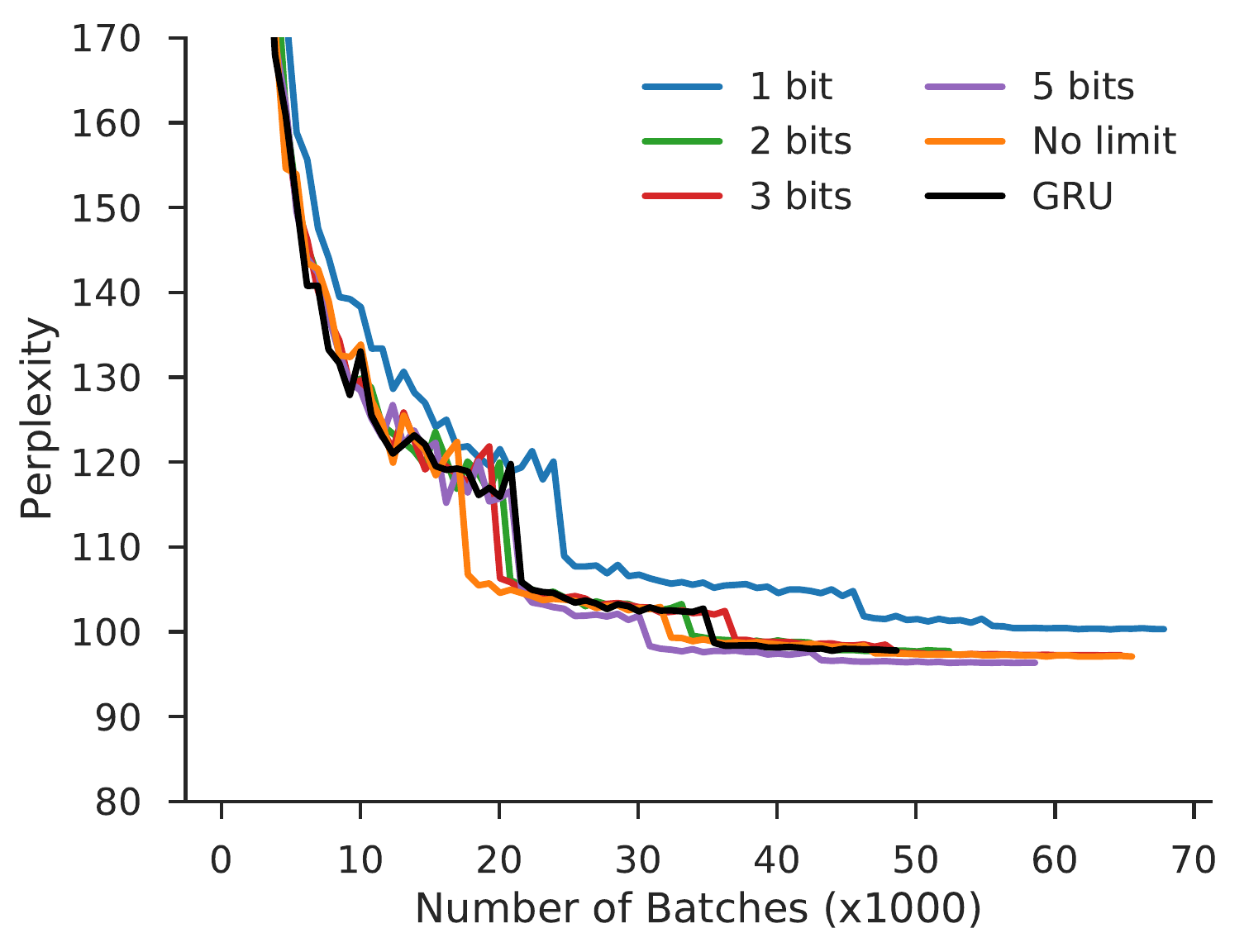}
  \end{center}
  \label{fig:single-dropout}
\end{figure}

\subsubsection*{2 layer RevGRU}
\begin{figure}[H]
  \caption*{Training/validation perplexity for a 2-layer RevGRU on WikiText-2 with various restrictions on forgetting and a baseline GRU model. \textbf{Left:} Perplexity on the training set.  \textbf{Right:} Perplexity on the validation set.}
  \begin{center}
    \includegraphics[width=0.5\textwidth]{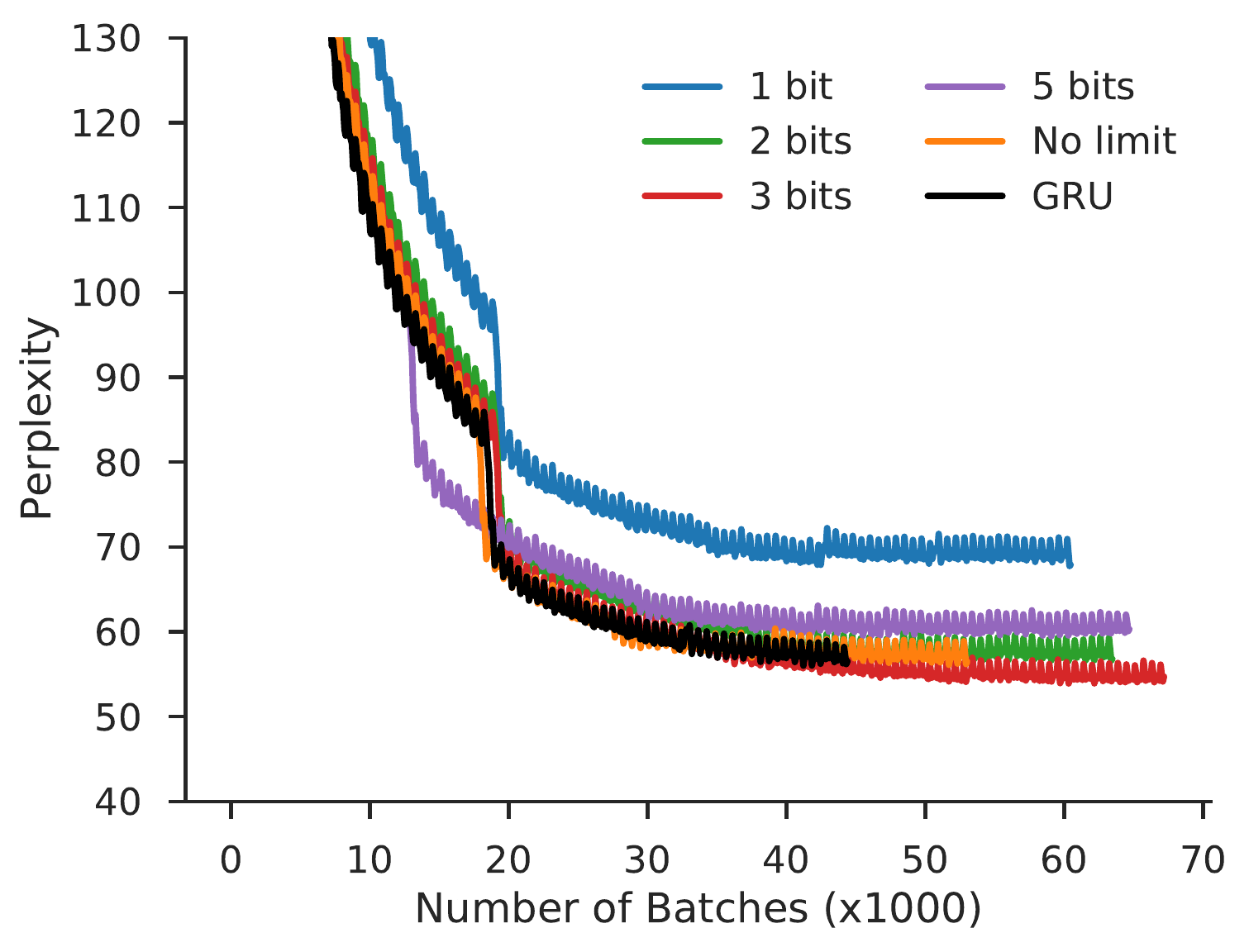}\includegraphics[width=0.5\textwidth]{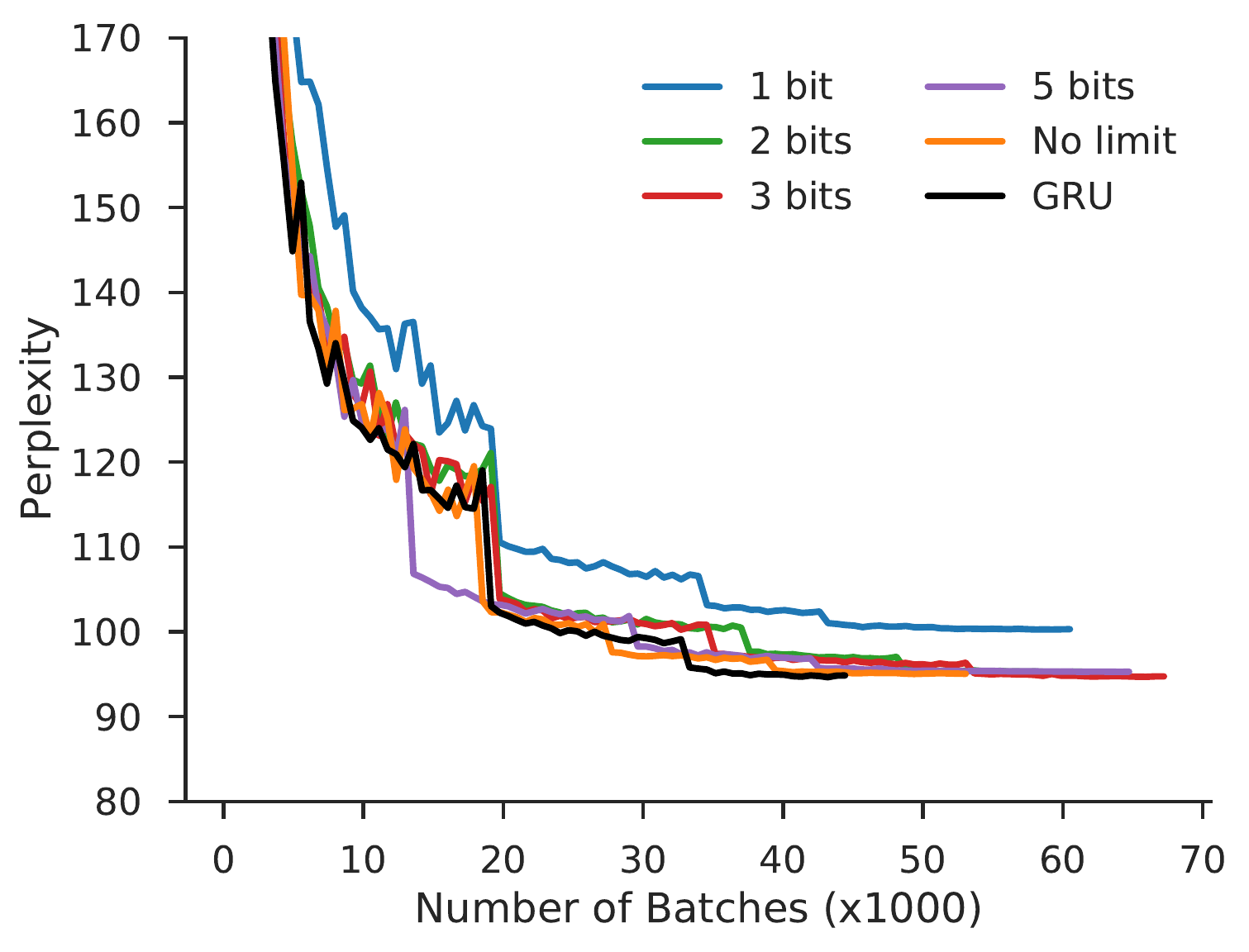}
  \end{center}
  \label{fig:single-dropout}
\end{figure}

\subsubsection*{1 layer RevLSTM}
\begin{figure}[H]
  \caption*{Training/validation perplexity for a 1-layer RevLSTM on WikiText-2 with various restrictions on forgetting and a baseline LSTM model. \textbf{Left:} Perplexity on the training set.  \textbf{Right:} Perplexity on the validation set.}
  \begin{center}
    \includegraphics[width=0.5\textwidth]{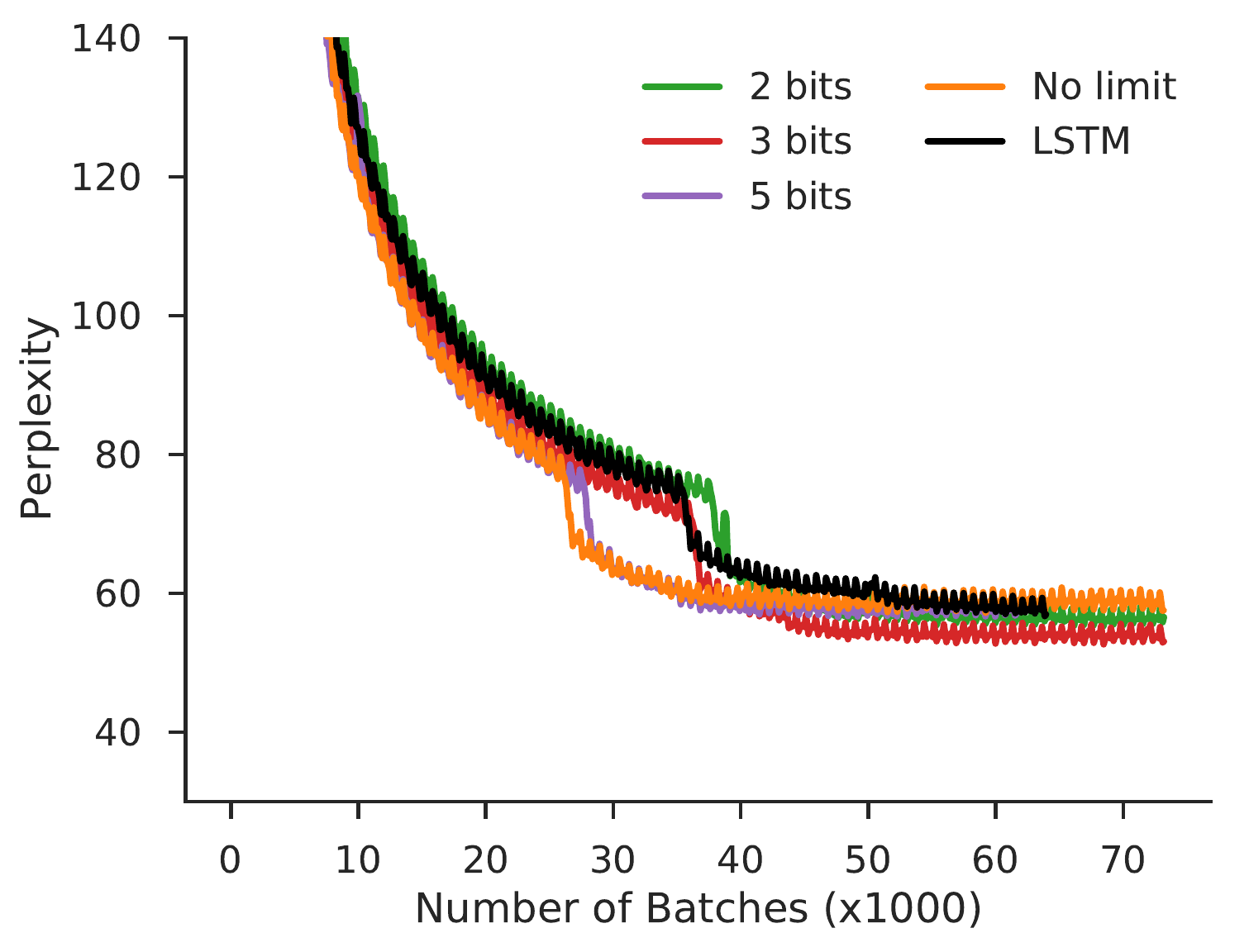}\includegraphics[width=0.5\textwidth]{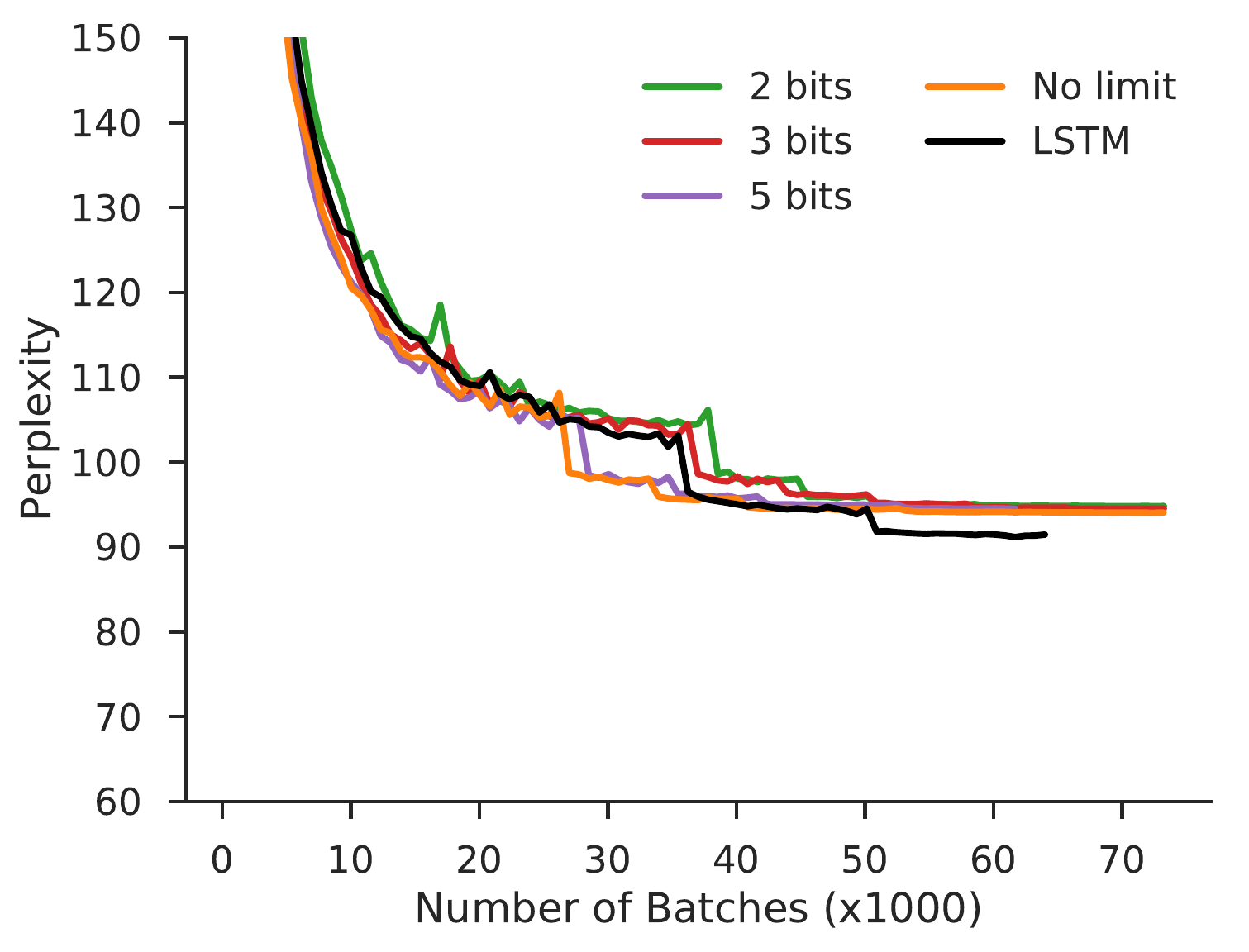}
  \end{center}
  \label{fig:single-dropout}
\end{figure}

\subsubsection*{2 layer RevLSTM}

\begin{figure}[H]
  \caption*{Training/validation perplexity for a 2-layer RevLSTM on WikiText-2 with various restrictions on forgetting and a baseline LSTM model. \textbf{Left:} Perplexity on the training set.  \textbf{Right:} Perplexity on the validation set.}
  \begin{center}
    \includegraphics[width=0.5\textwidth]{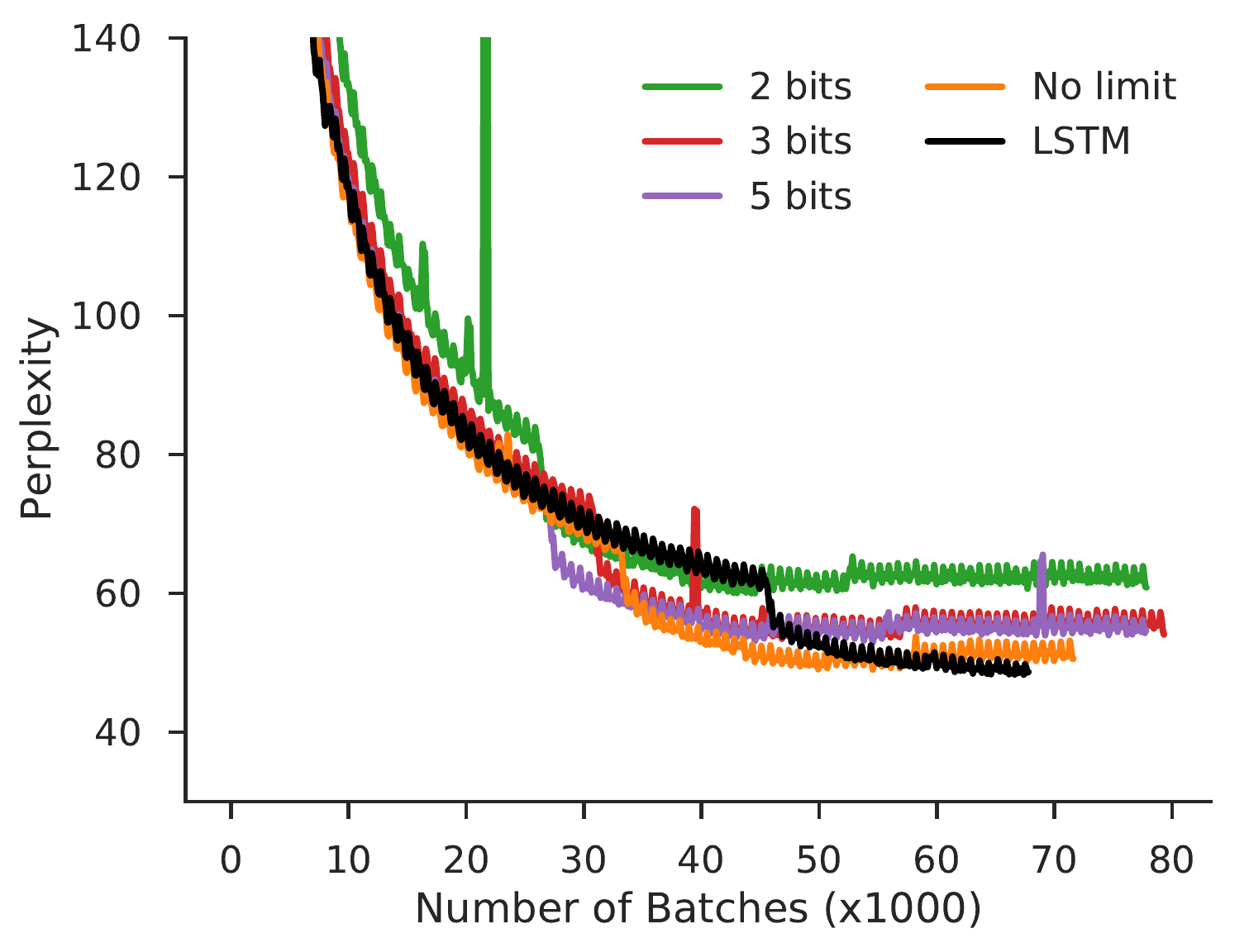}\includegraphics[width=0.5\textwidth]{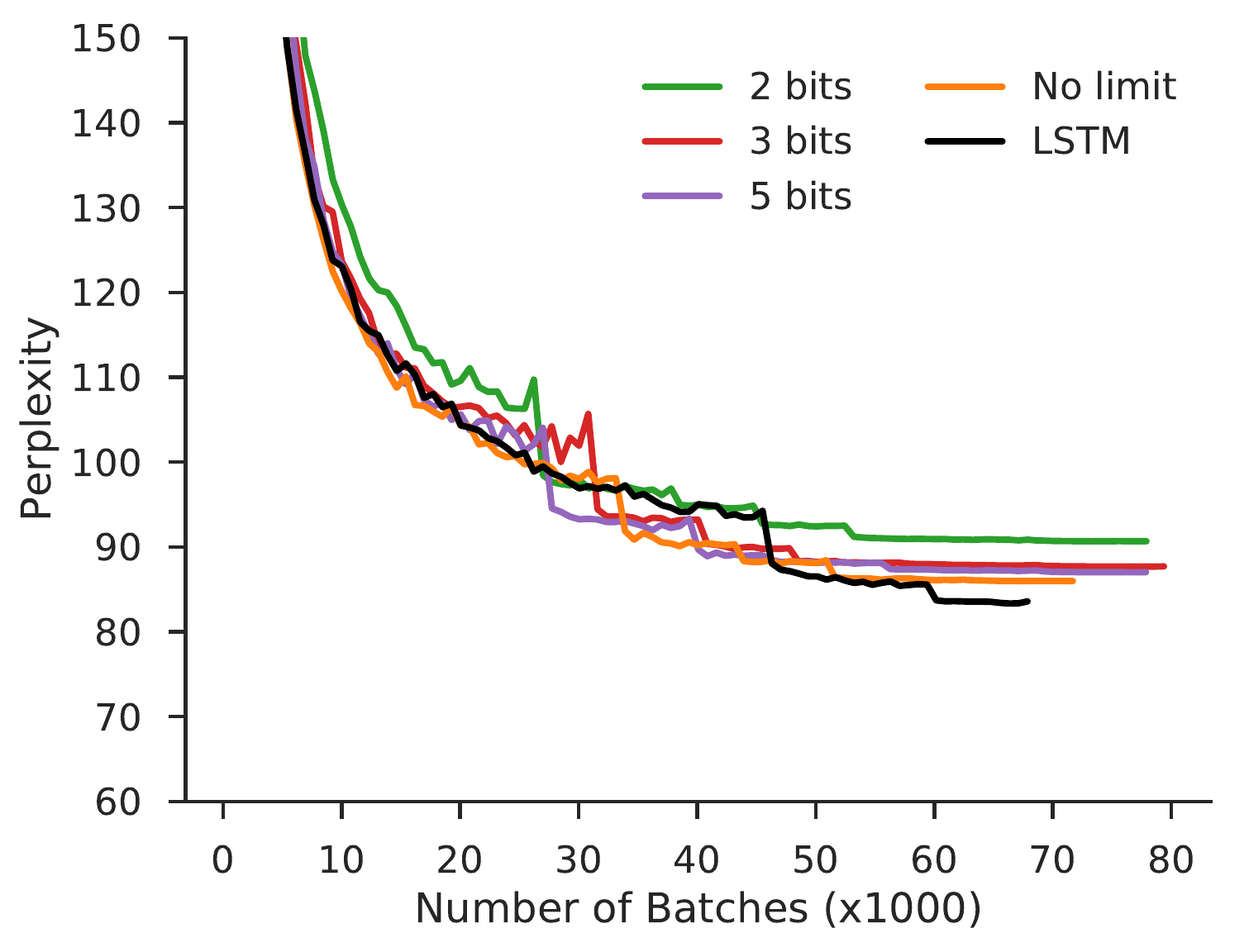}
  \end{center}
  \label{fig:single-dropout}
\end{figure}

\subsection{Multi30K NMT}
\label{app:train-valid-curves-multi30k}

In this section we show the training and validation curves for the RevLSTM and RevGRU NMT models with various types of attention (20H, 100H, 300H, Emb, and Emb+20H) and restrictions on forgetting (1, 2, 3, and 5 bits, and no limit on forgetting).
For Multi30K, both the encoder and decoder are single-layer, unidirectional RNNs with 300  hidden units.

\subsubsection{RevGRU}

\begin{figure}[H]
  \begin{center}
    \includegraphics[width=0.44\textwidth]{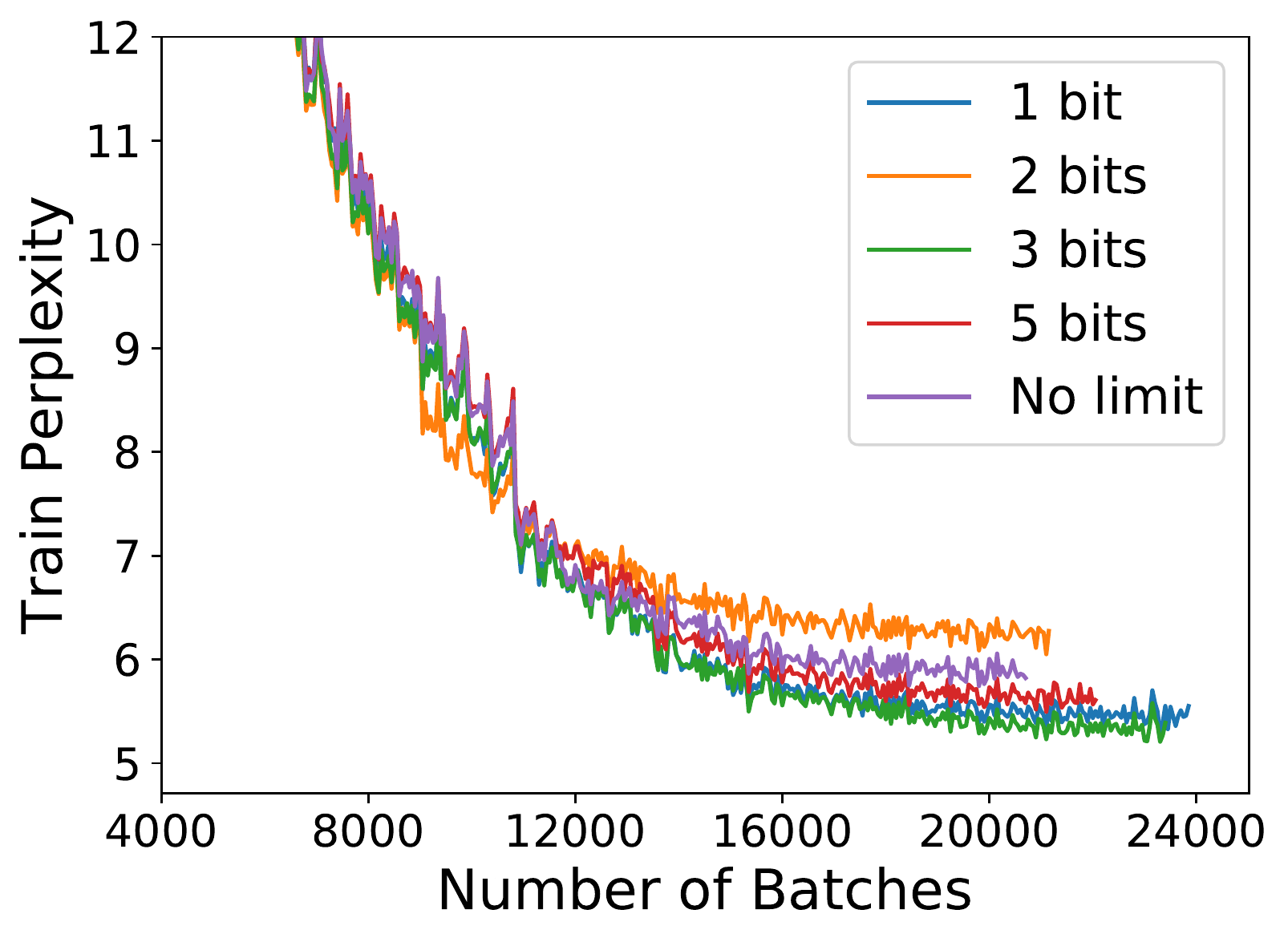}\includegraphics[width=0.45\textwidth]{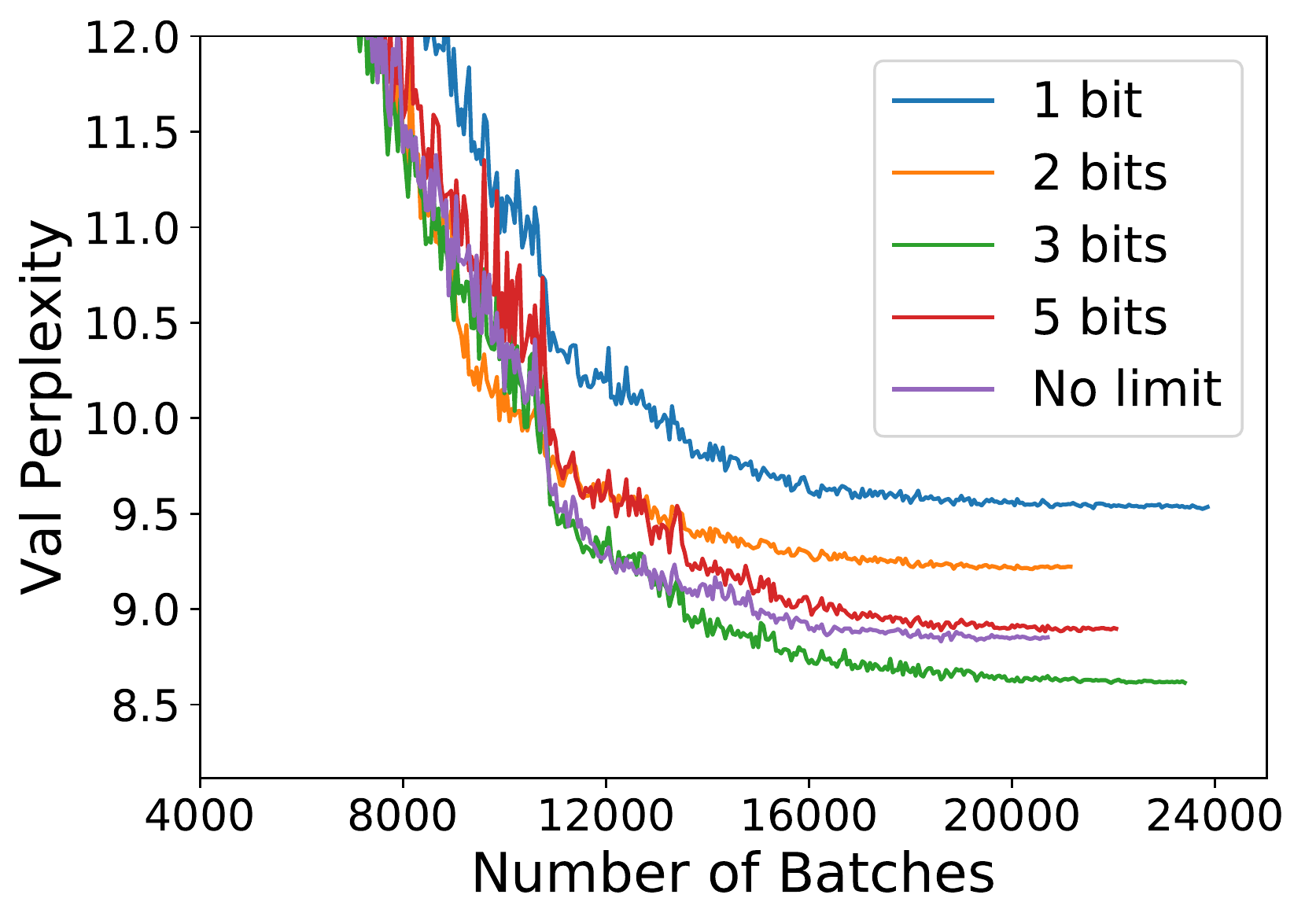}
  \end{center}
  \label{}
  \caption{\textbf{RevGRU 20H} (attention over a 20-dimensional slice of the hidden state).}
\end{figure}

\begin{figure}[H]
  \begin{center}
    \includegraphics[width=0.45\textwidth]{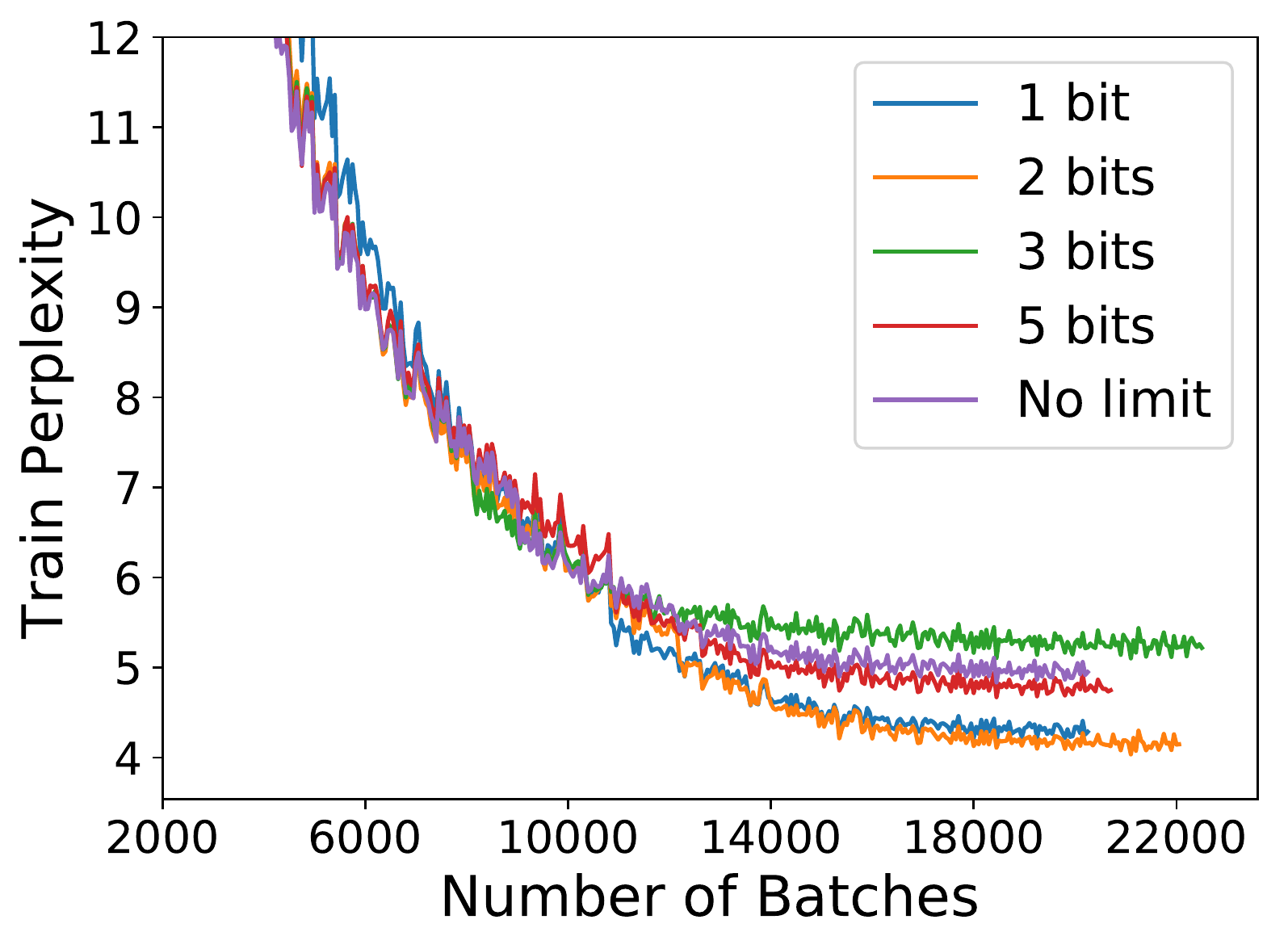}\includegraphics[width=0.45\textwidth]{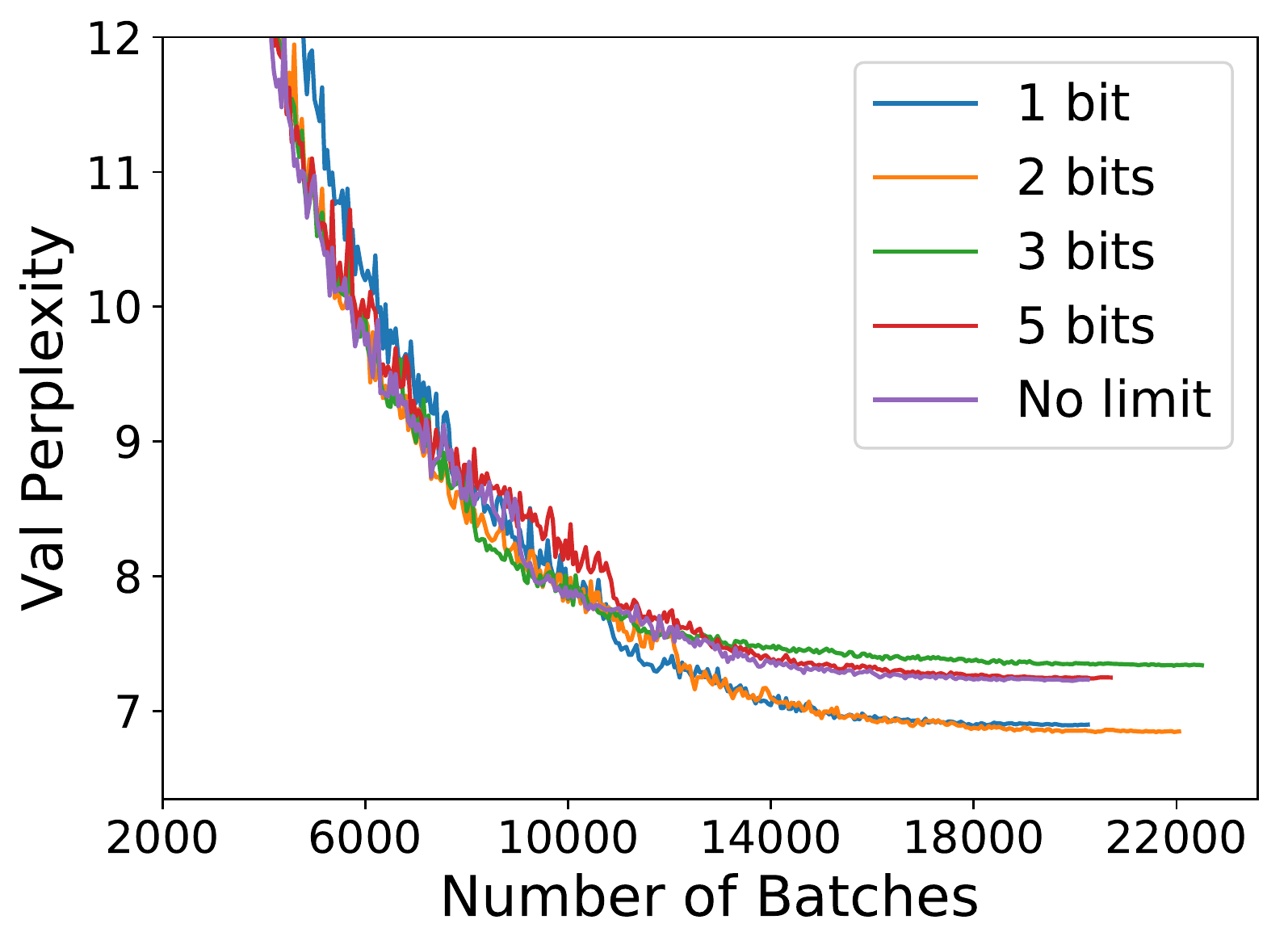}
  \end{center}
  \label{}
  \caption{\textbf{RevGRU 100H} (attention over a 100-dimensional slice of the hidden state).}
\end{figure}

\begin{figure}[H]
  \begin{center}
    \includegraphics[width=0.45\textwidth]{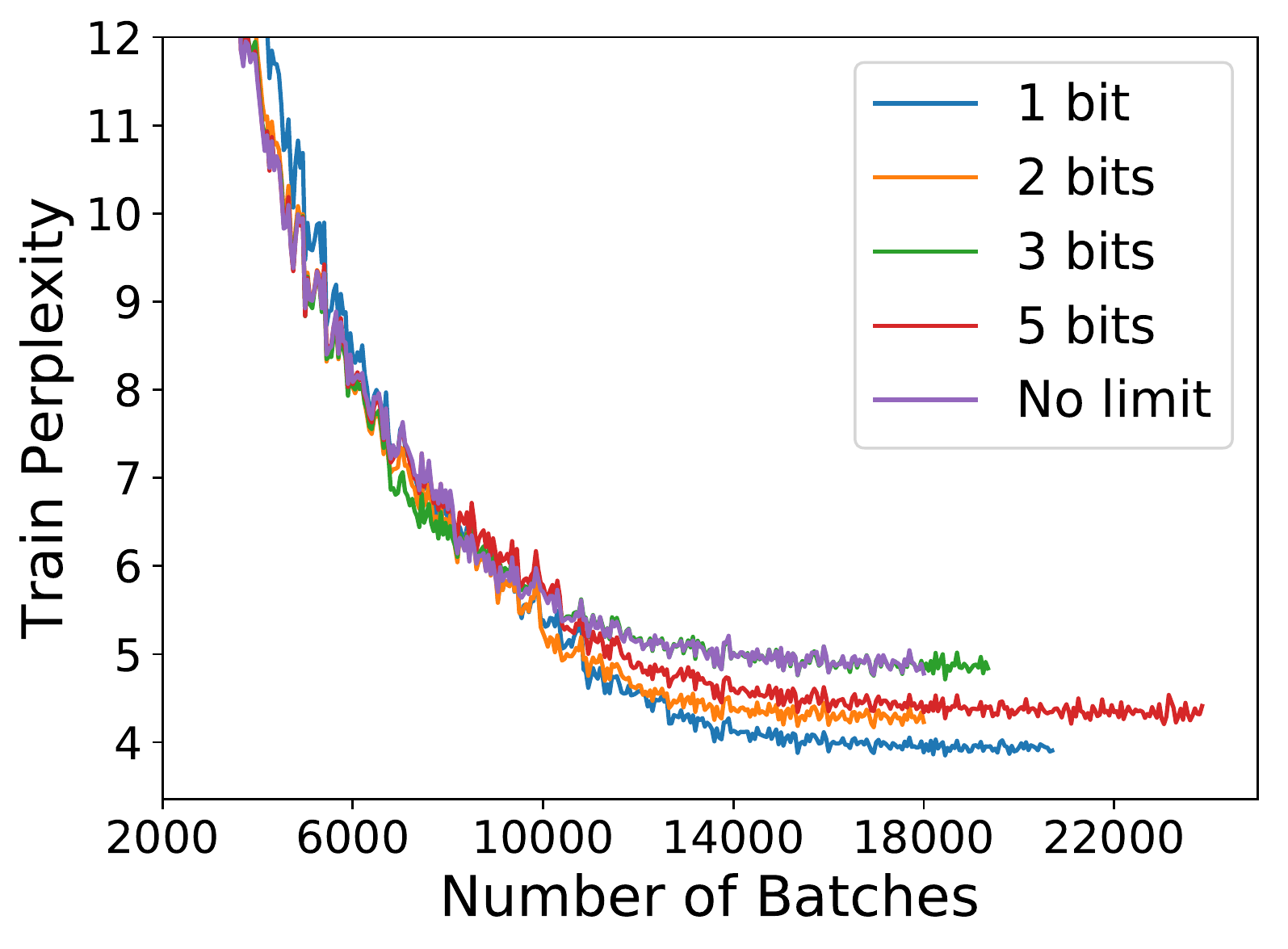}\includegraphics[width=0.45\textwidth]{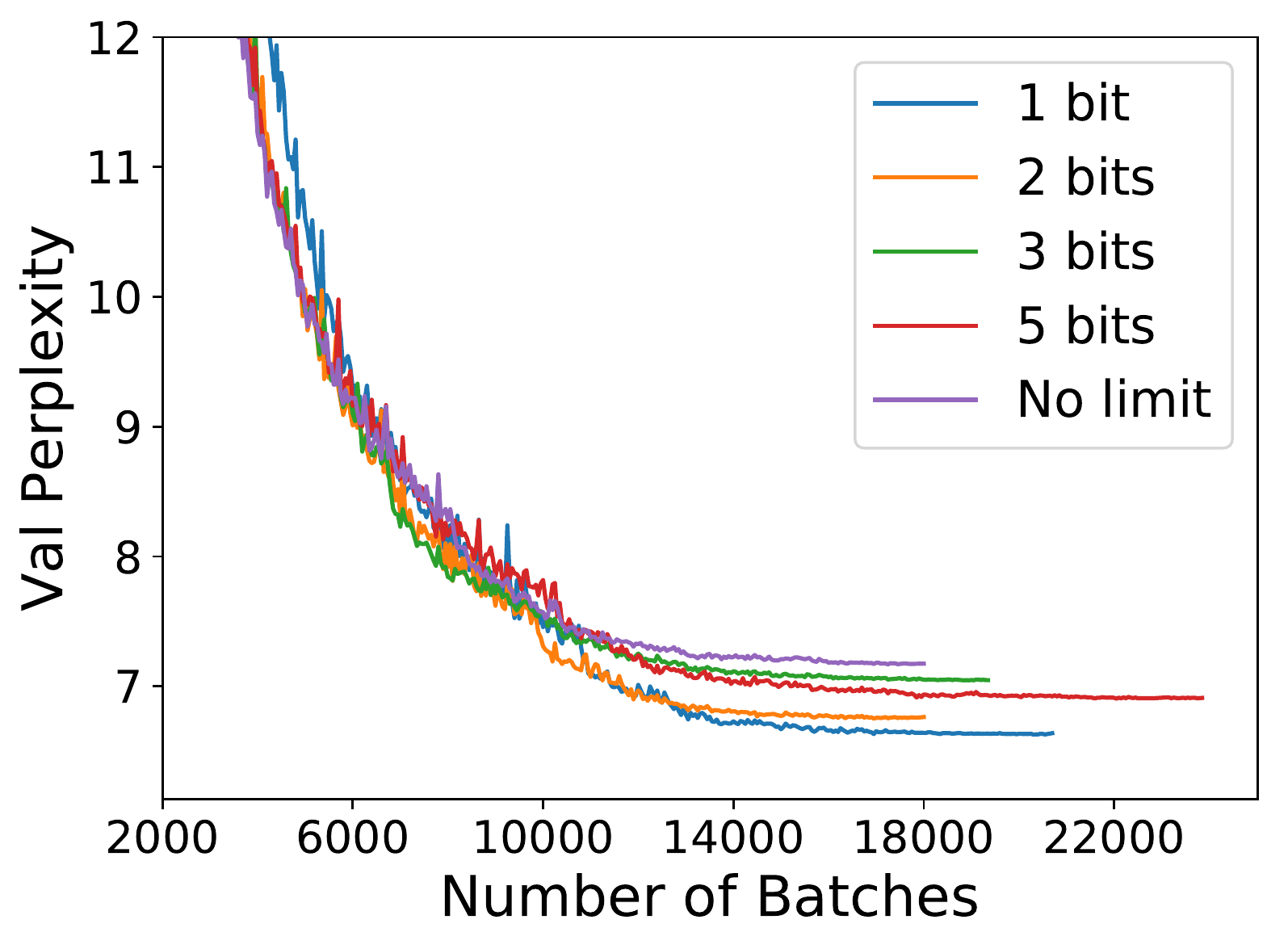}
  \end{center}
  \label{}
  \caption{\textbf{RevGRU 300H} (attention over the whole hidden state).}
\end{figure}

\begin{figure}[H]
  \begin{center}
    \includegraphics[width=0.44\textwidth]{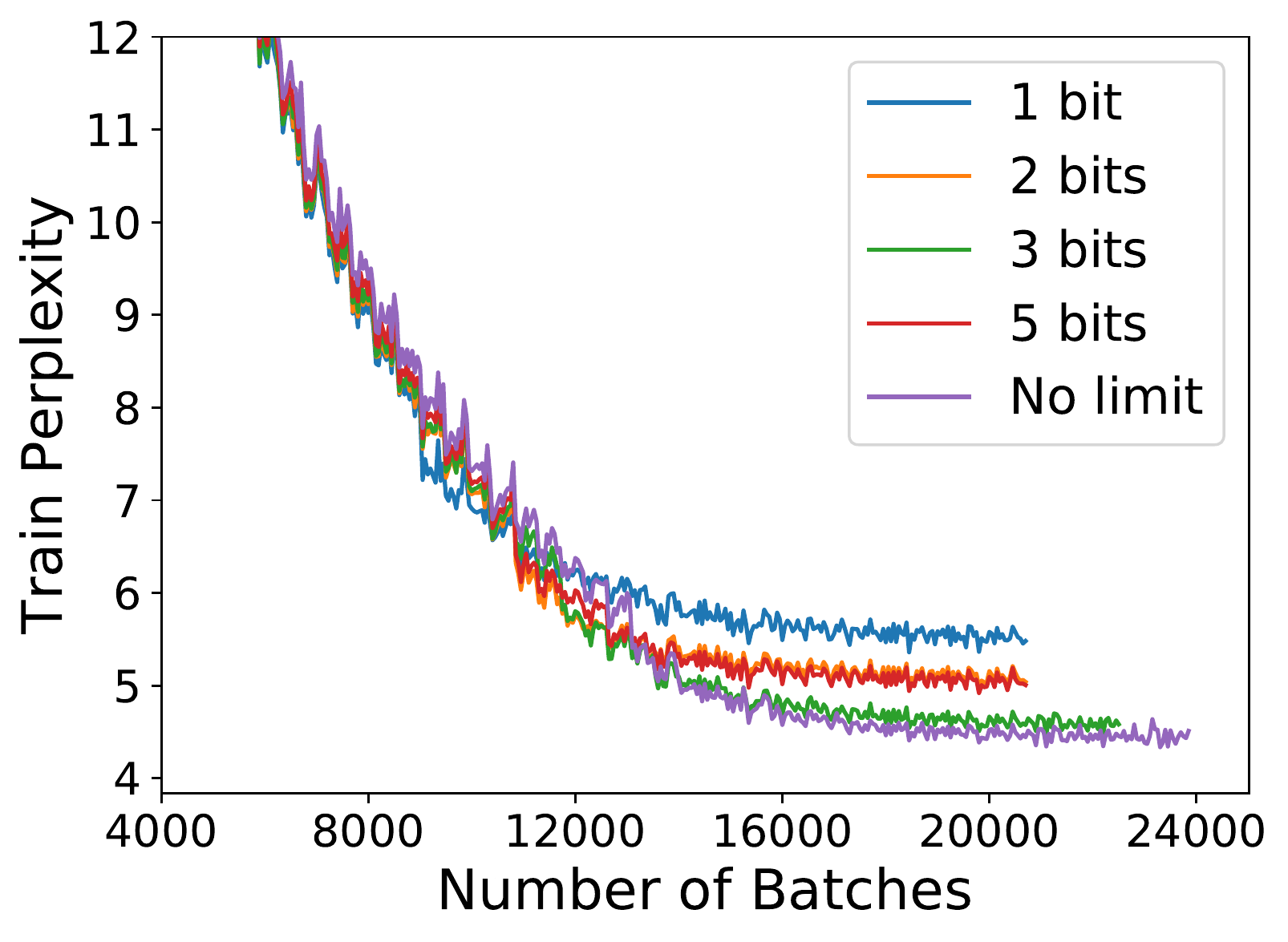}\includegraphics[width=0.45\textwidth]{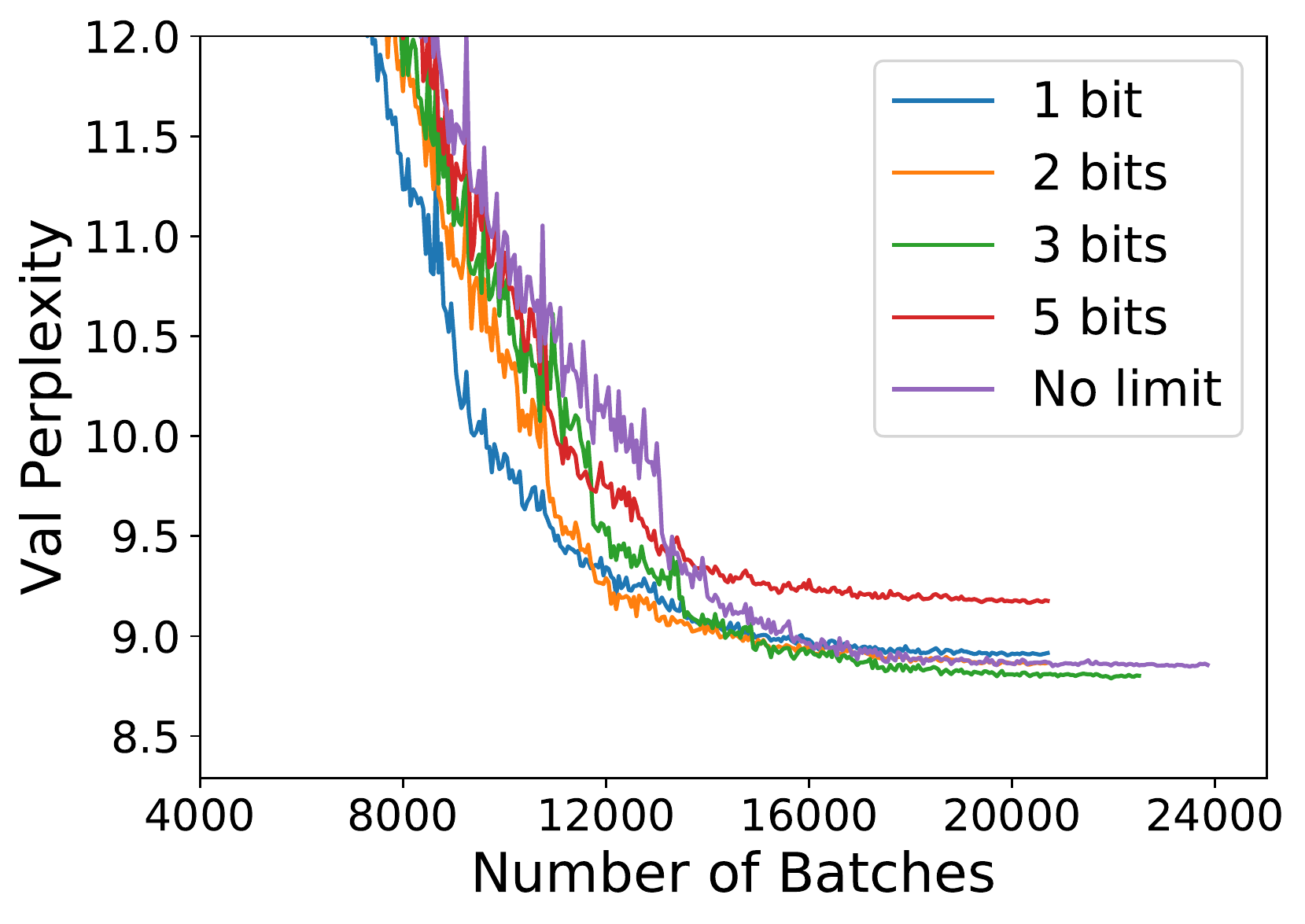}
  \end{center}
  \label{}
  \caption{\textbf{RevGRU Emb} (attention over the input word embeddings).}
\end{figure}

\begin{figure}[H]
  \begin{center}
    \includegraphics[width=0.45\textwidth]{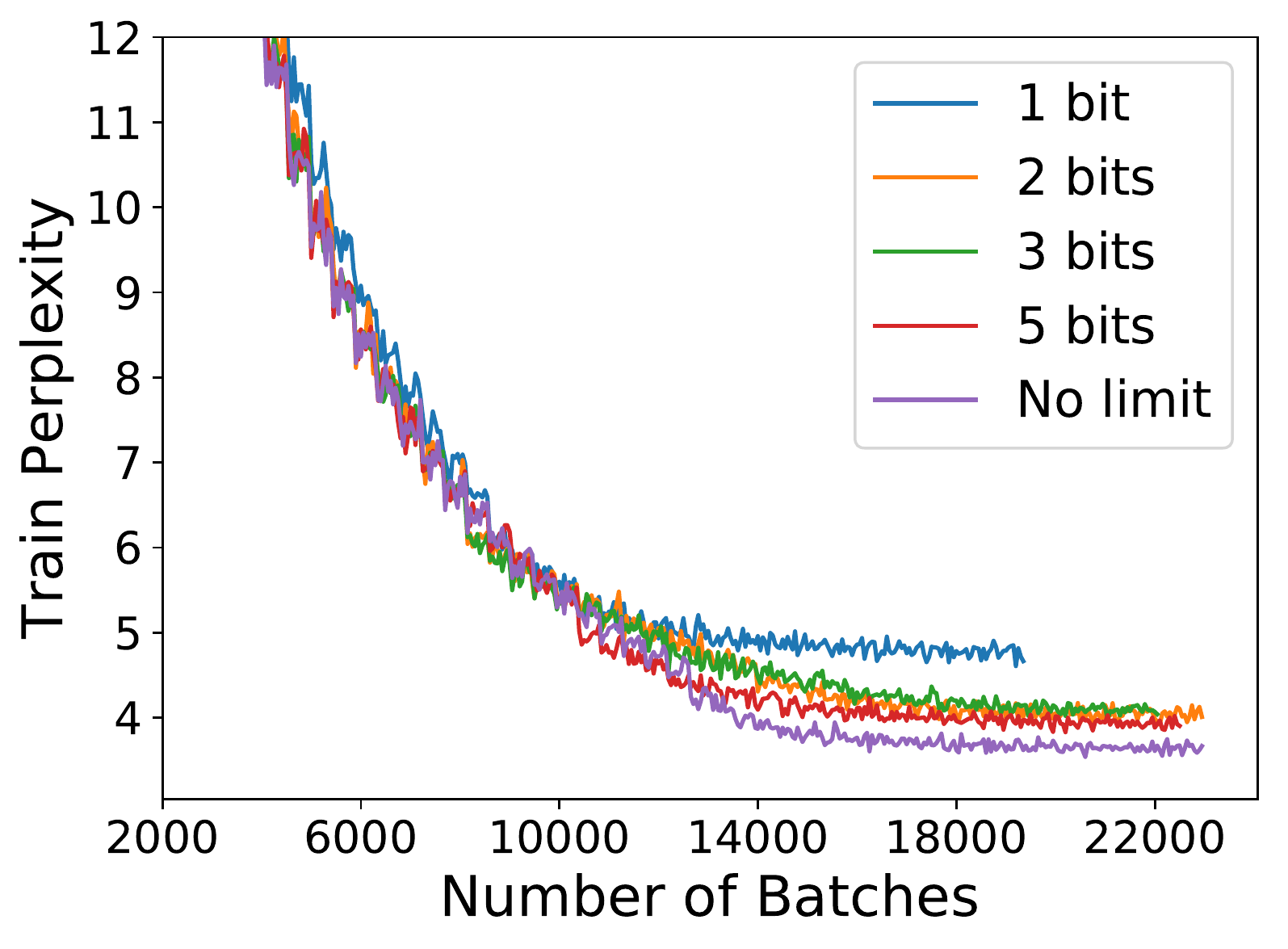}\includegraphics[width=0.45\textwidth]{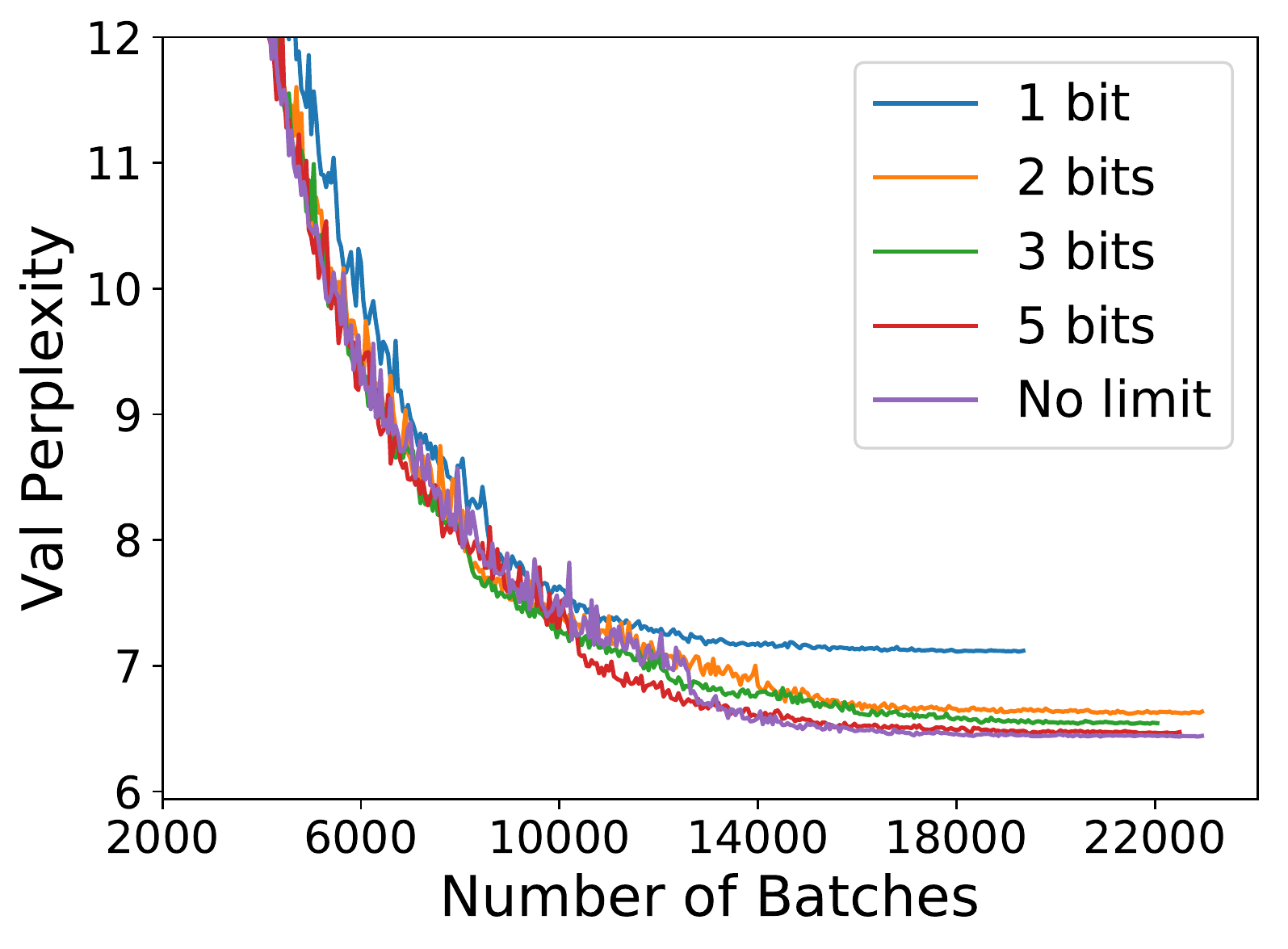}
  \end{center}
  \label{}
  \caption{\textbf{RevGRU Emb+20H} (attention over a concatenation of the word embeddings and a 20-dimensional slice of the hidden state).}
\end{figure}

\subsubsection{RevLSTM}

\begin{figure}[H]
  \begin{center}
    \includegraphics[width=0.45\textwidth]{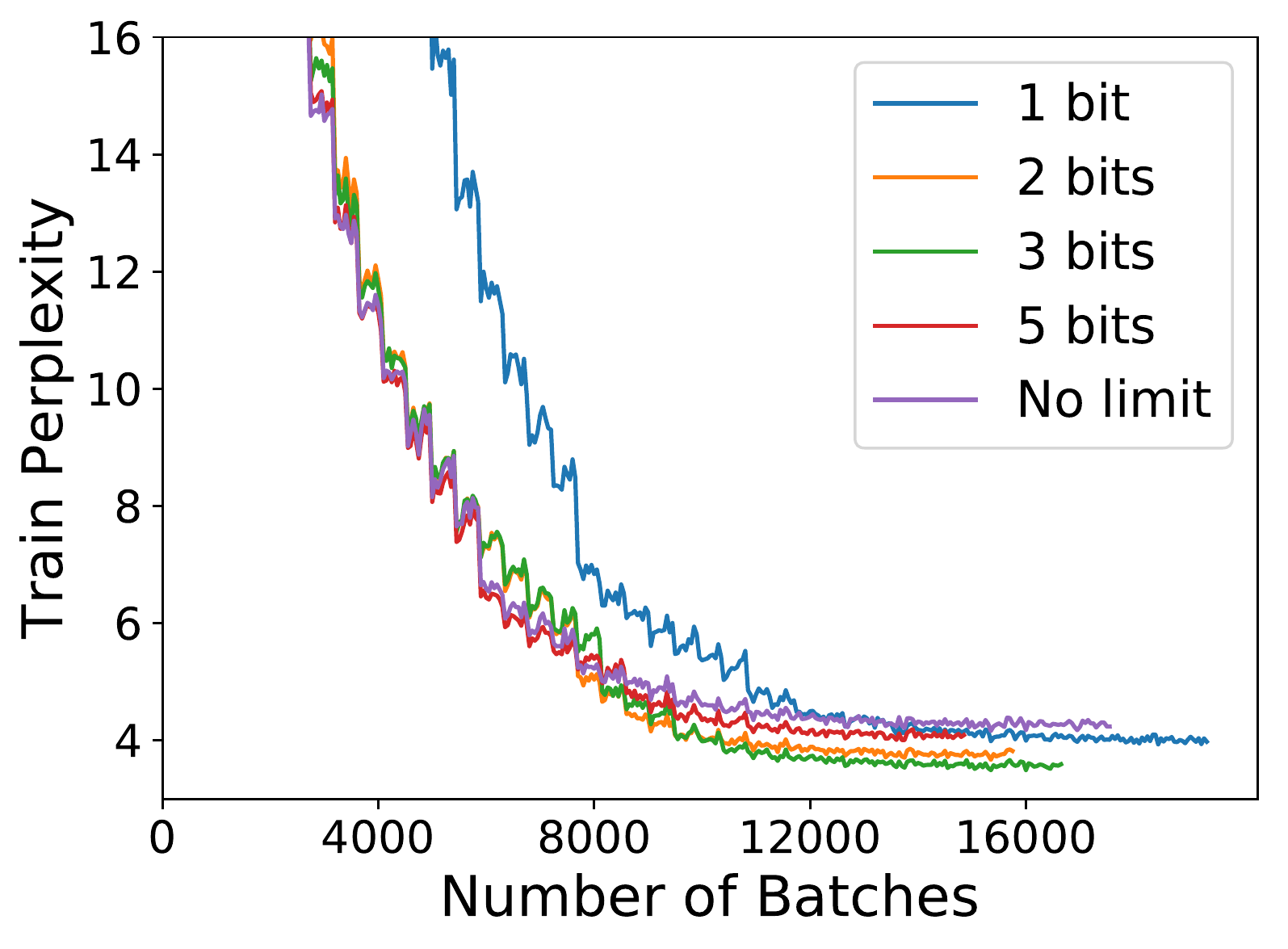}\includegraphics[width=0.45\textwidth]{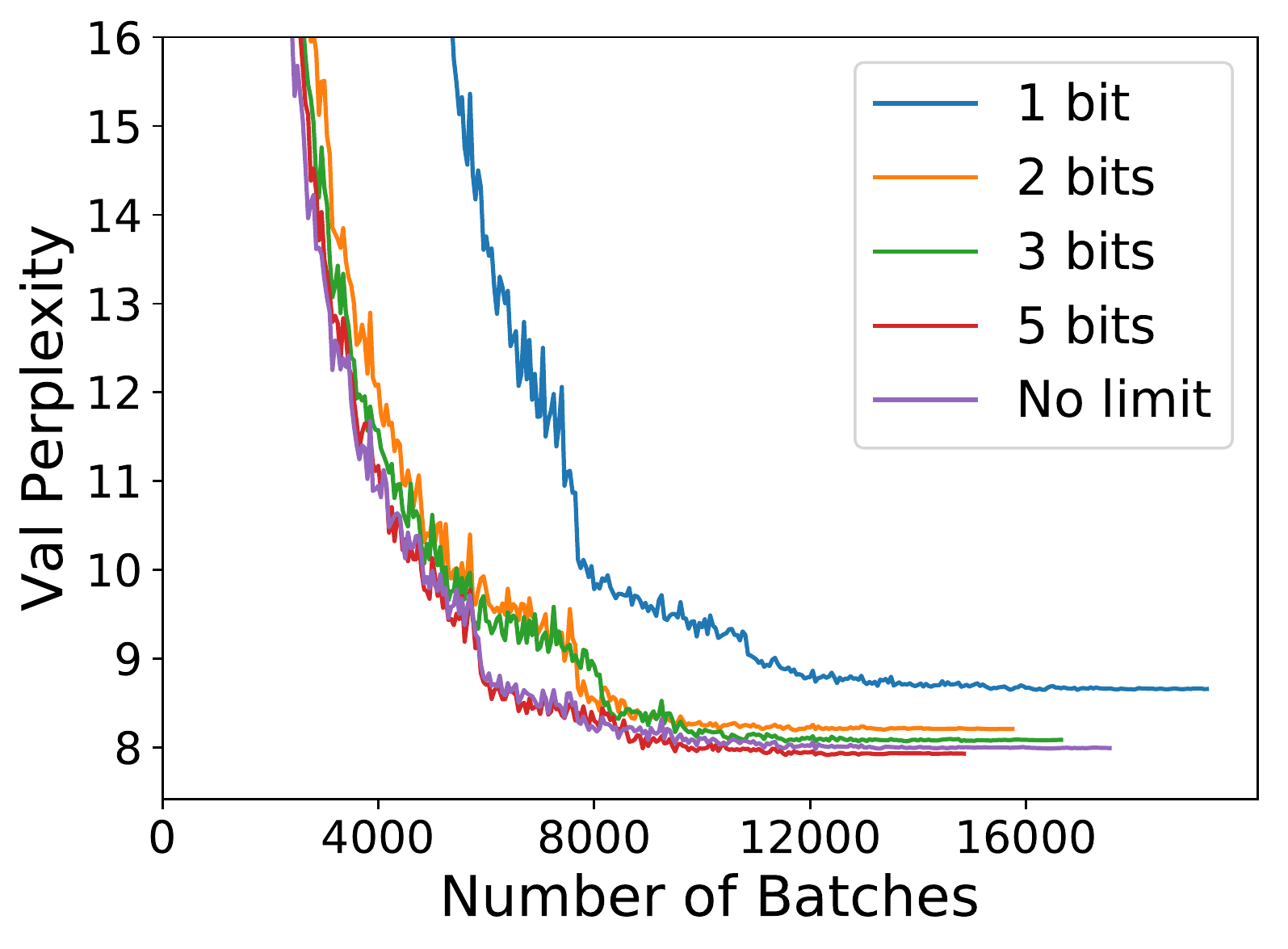}
  \end{center}
  \label{}
  \caption{\textbf{RevLSTM 20H} (attention over a 20-dimensional slice of the hidden state).}
\end{figure}

\begin{figure}[H]
  \begin{center}
    \includegraphics[width=0.45\textwidth]{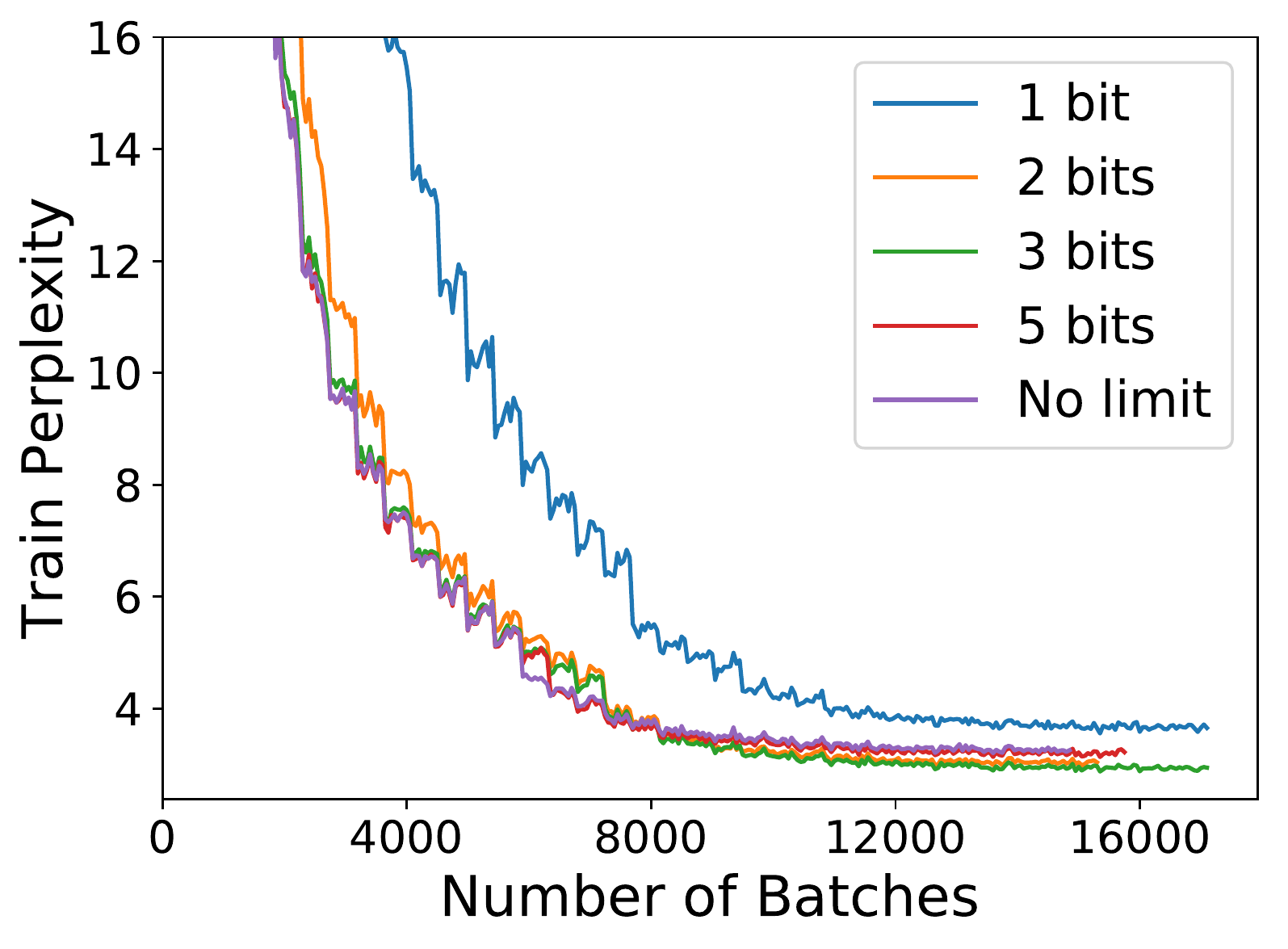}\includegraphics[width=0.45\textwidth]{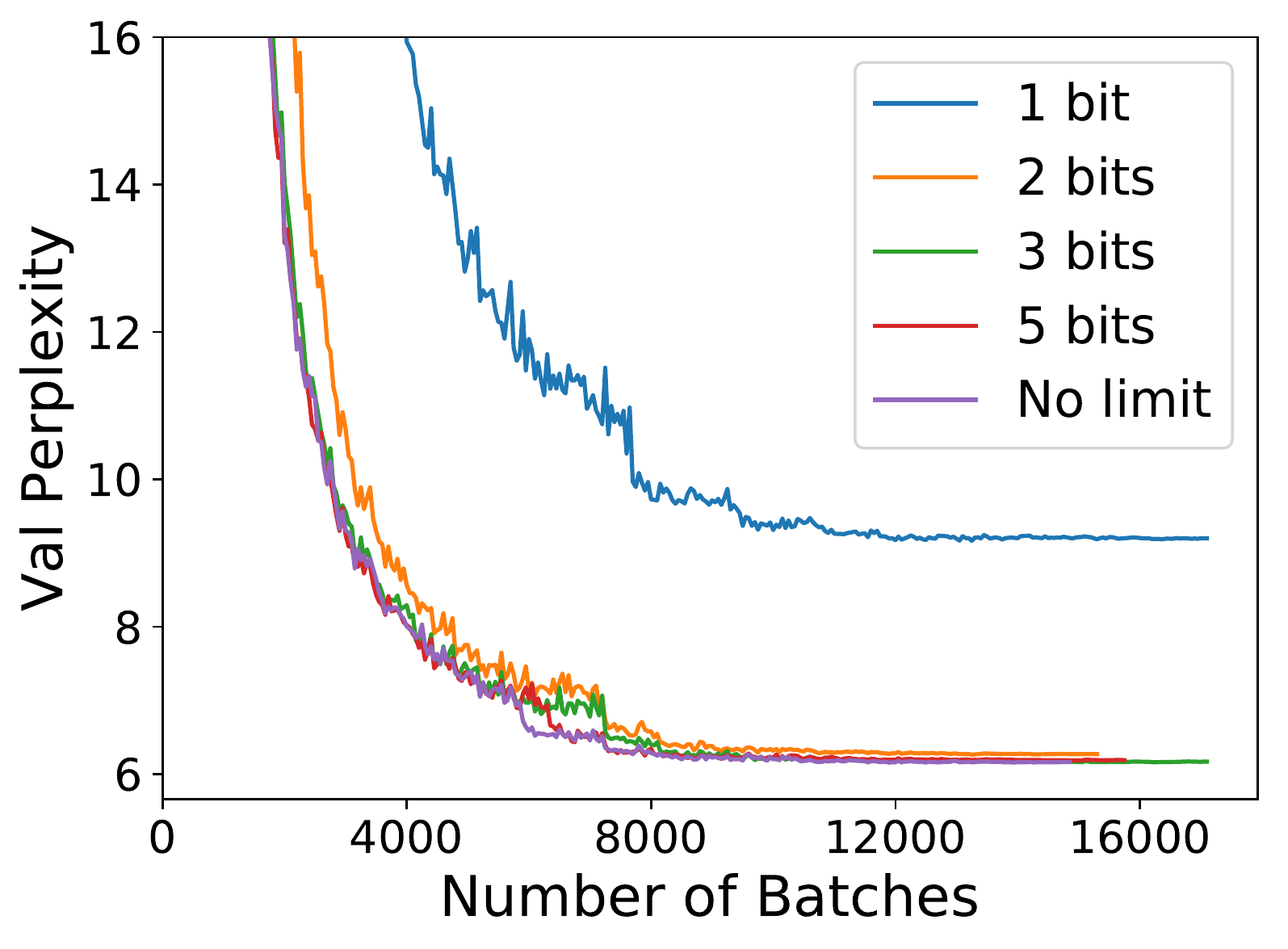}
  \end{center}
  \label{}
  \caption{\textbf{RevLSTM 100H} (attention over a 100-dimensional slice of the hidden state).}
\end{figure}

\begin{figure}[H]
  \begin{center}
    \includegraphics[width=0.45\textwidth]{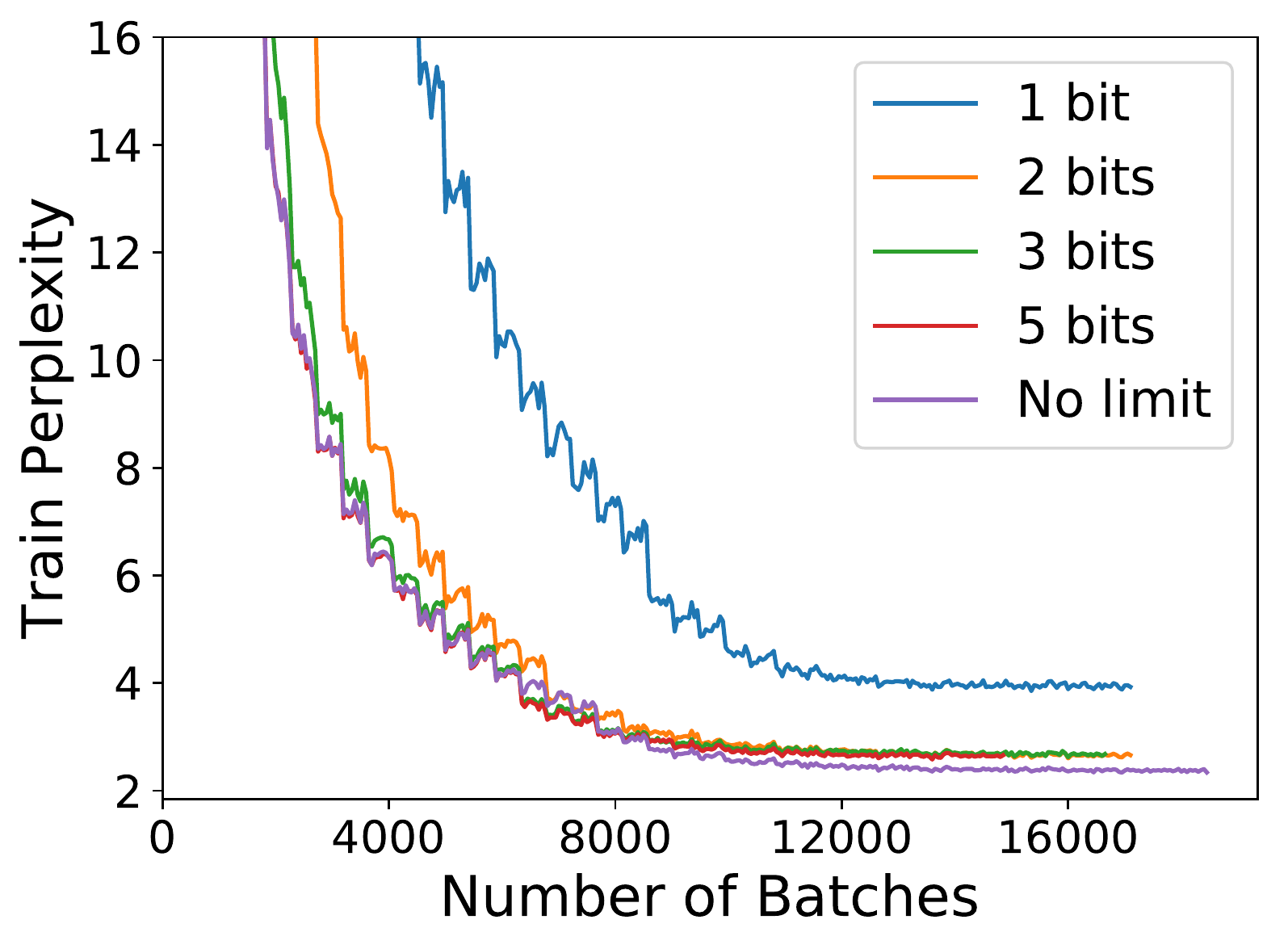}\includegraphics[width=0.45\textwidth]{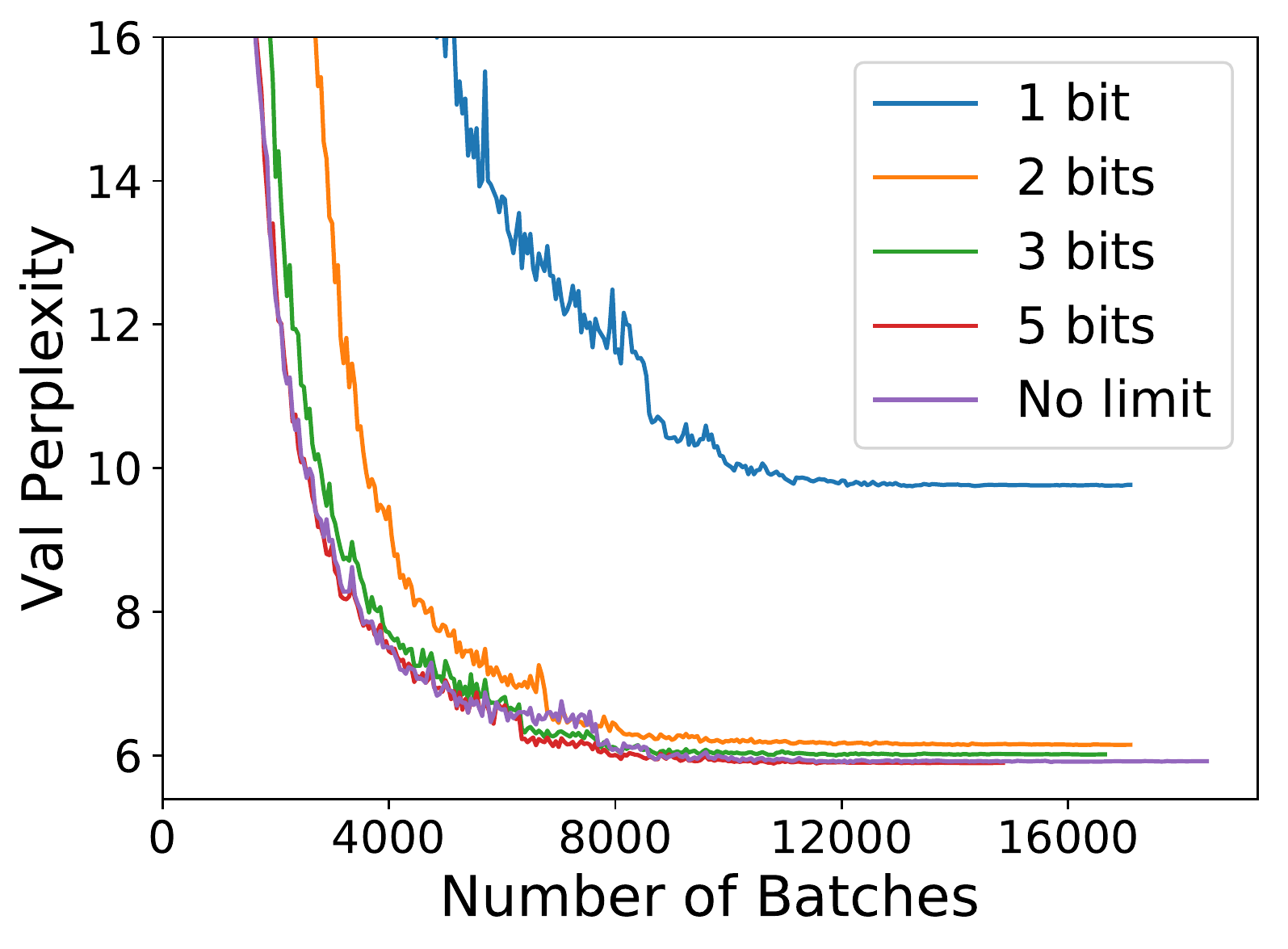}
  \end{center}
  \label{}
  \caption{\textbf{RevLSTM 300H} (attention over the whole hidden state).}
\end{figure}

\begin{figure}[H]
  \begin{center}
    \includegraphics[width=0.45\textwidth]{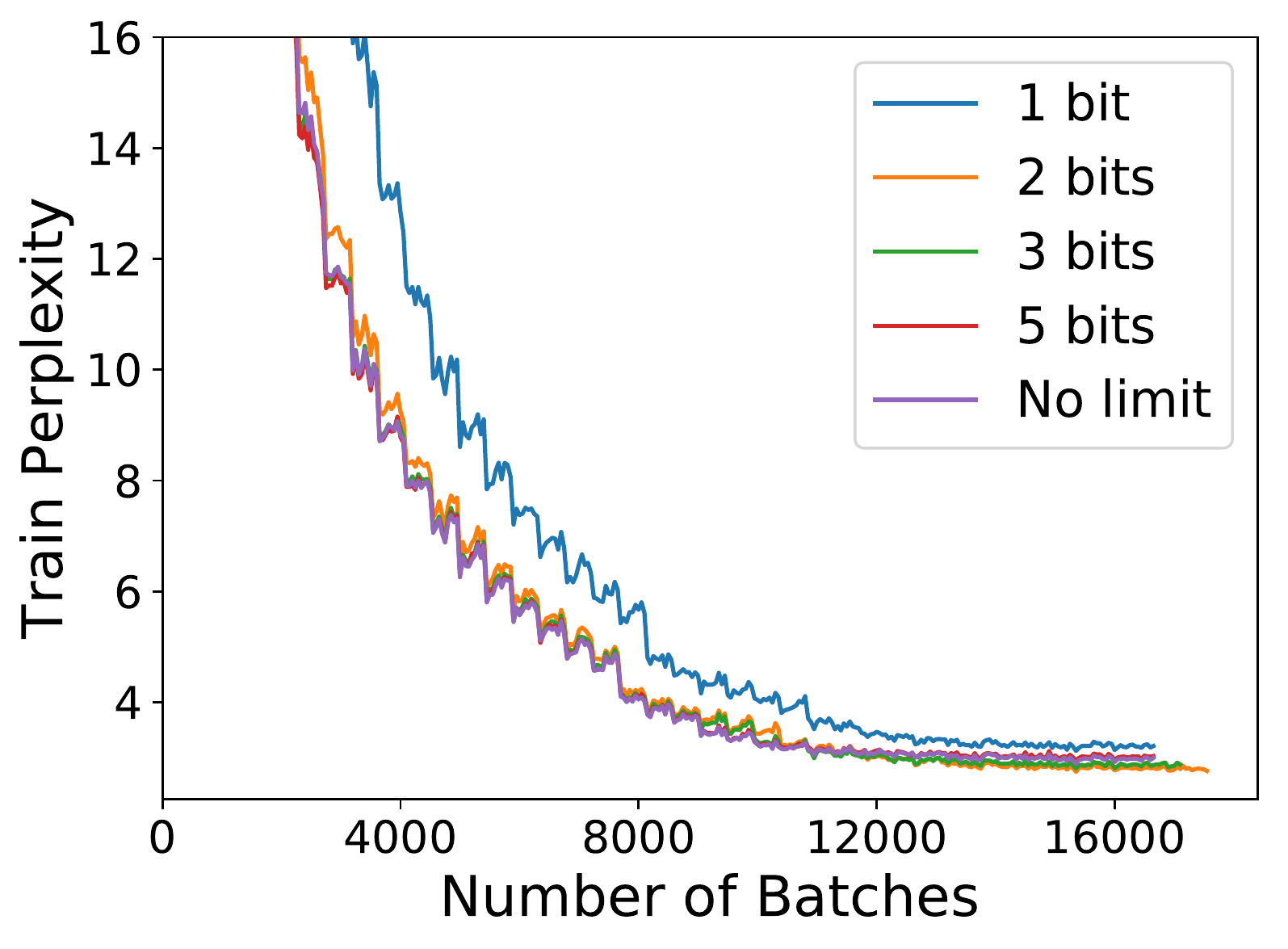}\includegraphics[width=0.45\textwidth]{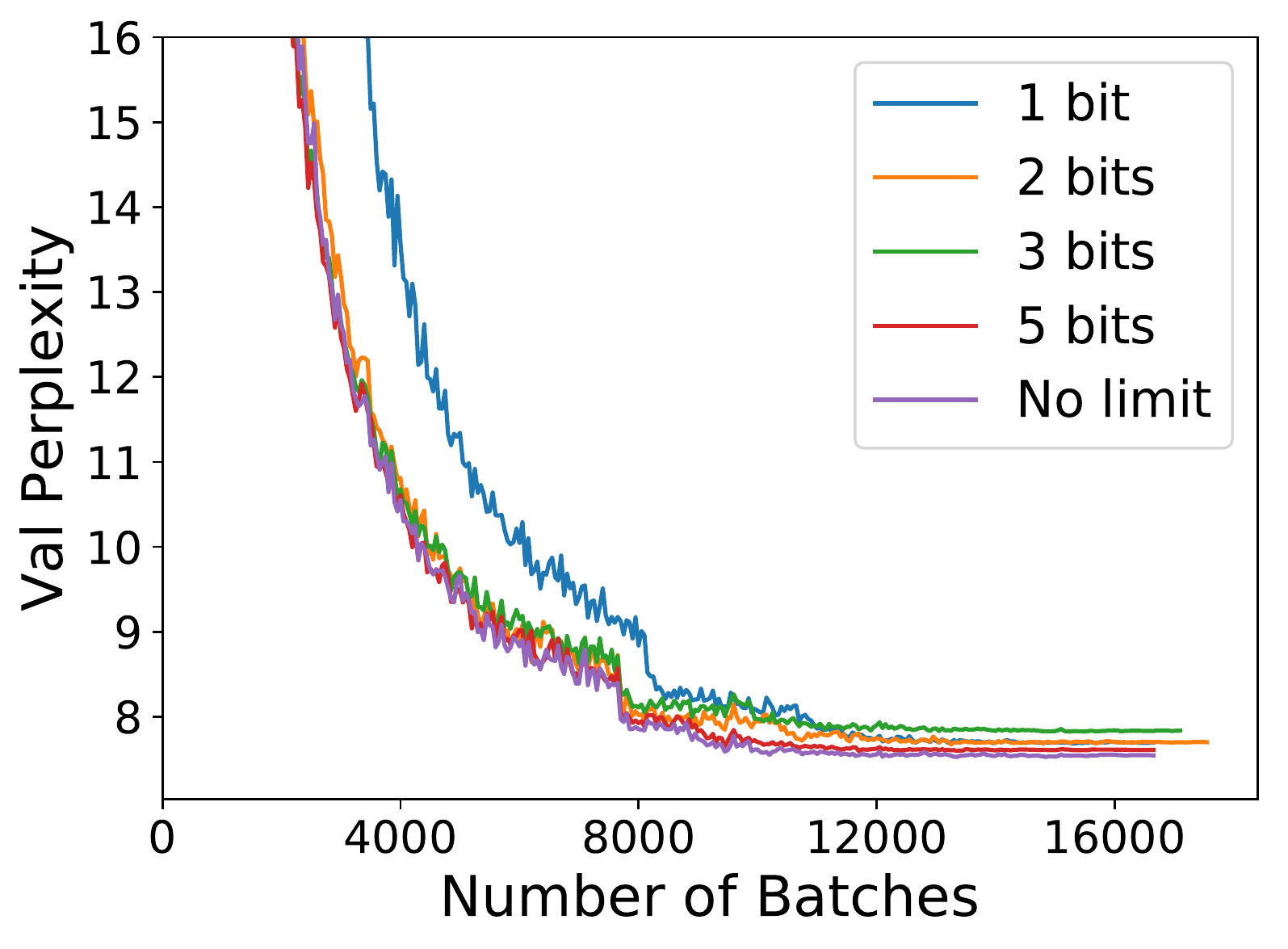}
  \end{center}
  \caption{\textbf{RevLSTM Emb} (attention over the input word embeddings).}
  \label{}
\end{figure}

\begin{figure}[H]
  \begin{center}
    \includegraphics[width=0.45\textwidth]{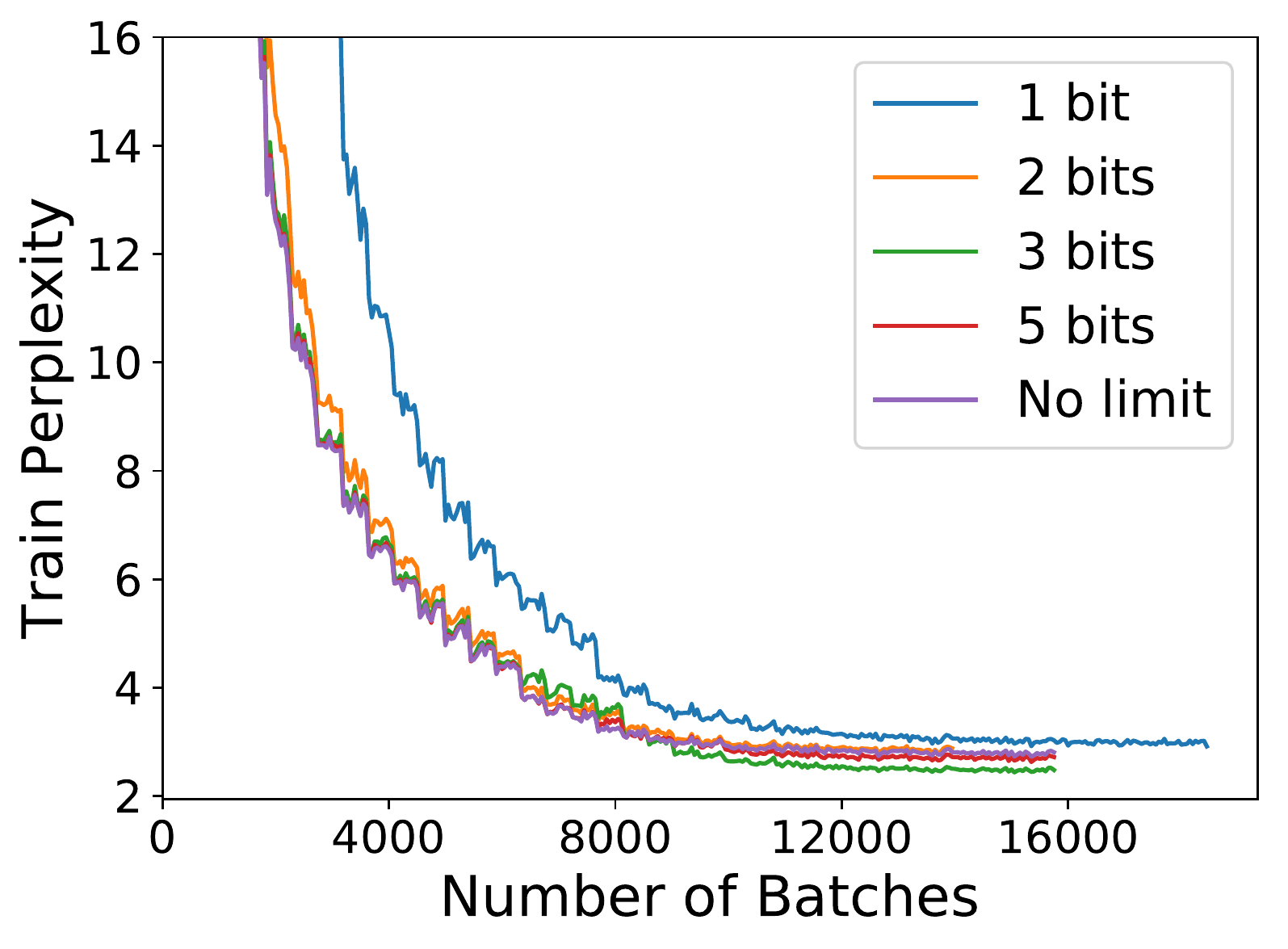}\includegraphics[width=0.45\textwidth]{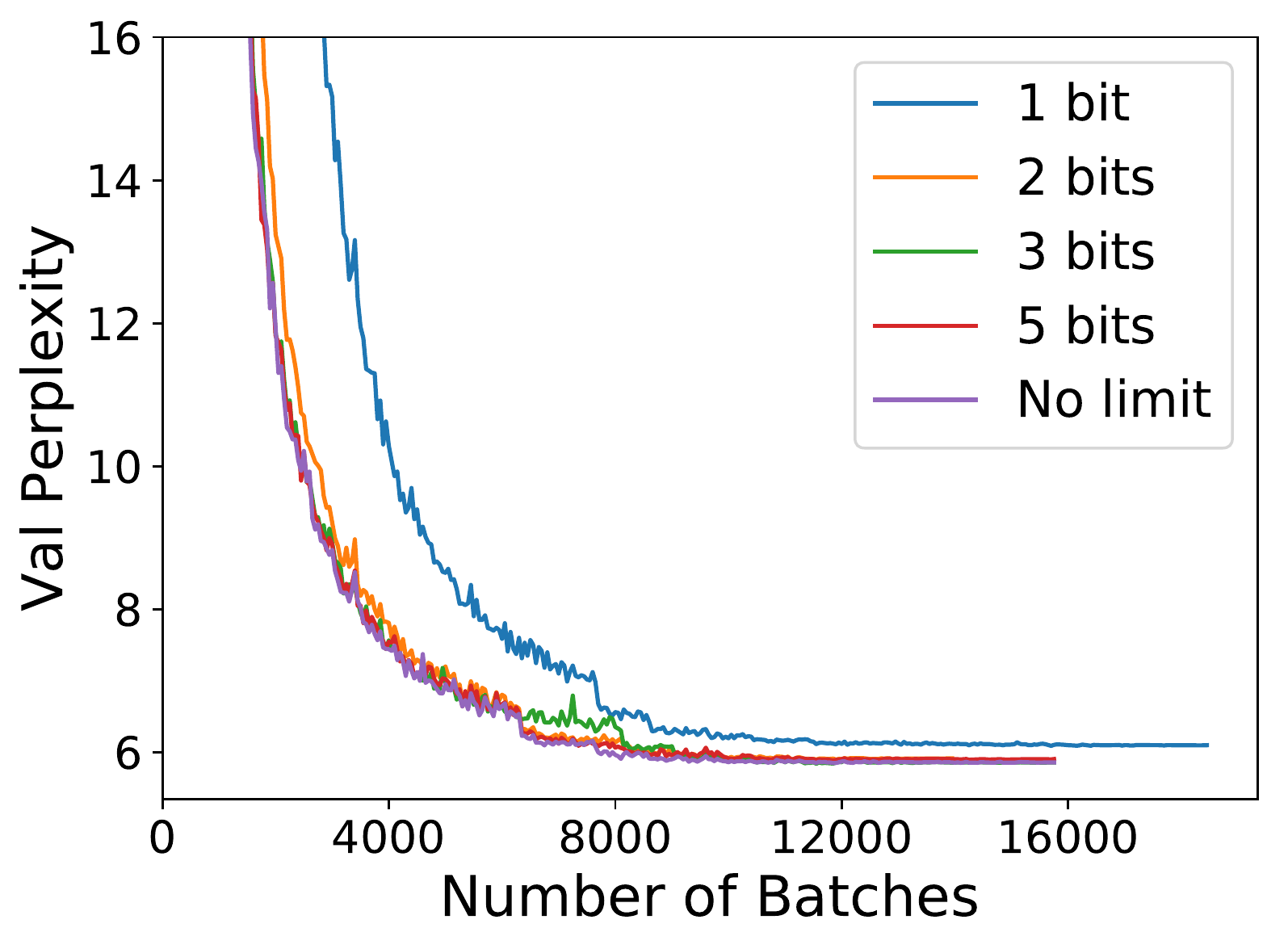}
  \end{center}
  \label{}
  \caption{\textbf{RevLSTM Emb+20H} (attention over a concatenation of the word embeddings and a 20-dimensional slice of the hidden state).}
\end{figure}

\subsection{IWSLT 2016}
\label{app:train-valid-curves-iwslt}

\begin{figure}[H]
  \begin{center}
    \includegraphics[width=0.4\textwidth]{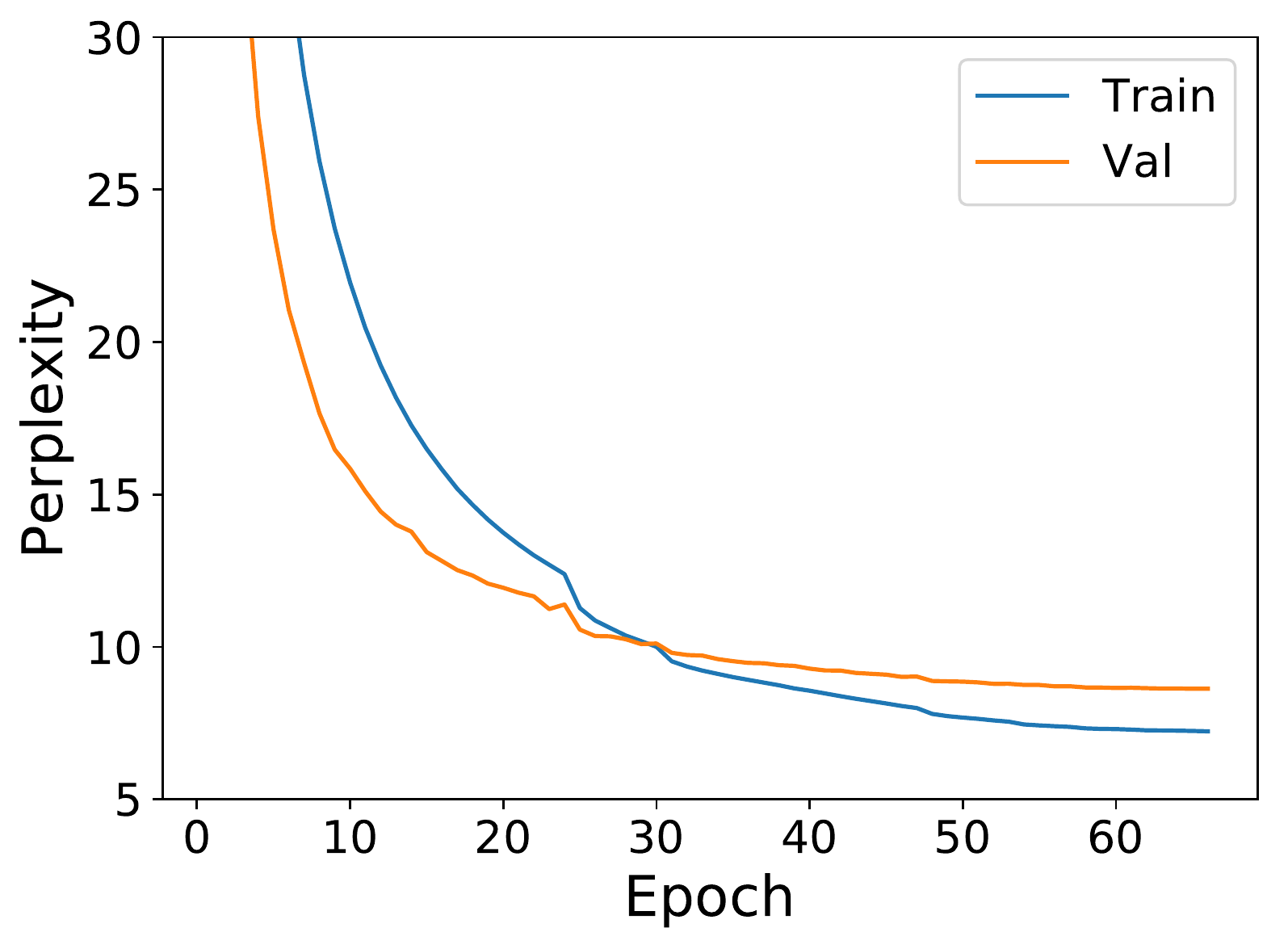}
    \hspace{4mm}
    \includegraphics[width=0.4\textwidth]{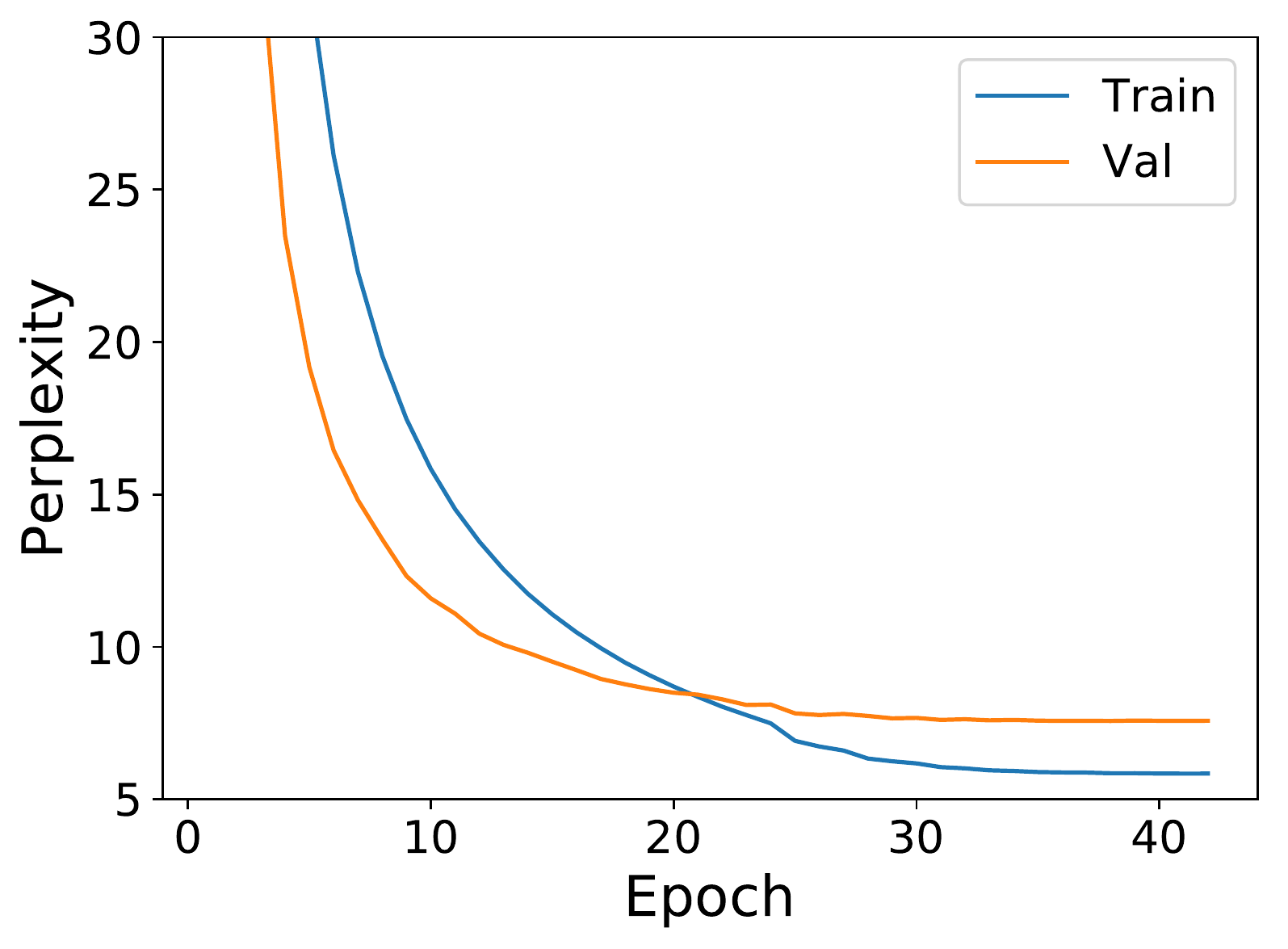}
  \end{center}
  \caption{Training/validation perplexity for a 2-layer, 600-hidden unit encoder-decoder architecture, with attention over a 60-dimensional slice of the hidden state, and 5 bit forgetting. \textbf{Left:} RevGRU.  \textbf{Right:} RevLSTM.}
  \label{}
\end{figure}


\end{document}